\documentclass[10pt]{article} 
\usepackage[preprint]{tmlr}

\usepackage{amsmath,amsfonts,bm}

\def\eqref#1{equation~\ref{#1}}

\def\1{\bm{1}}

\DeclareMathAlphabet{\mathsfit}{\encodingdefault}{\sfdefault}{m}{sl}
\SetMathAlphabet{\mathsfit}{bold}{\encodingdefault}{\sfdefault}{bx}{n}

\usepackage{hyperref}
\usepackage{url}

\usepackage{booktabs}
\usepackage{makecell}
\usepackage[table]{xcolor}

\definecolor{selfnle}{HTML}{EAF2FB}
\definecolor{humannle}{HTML}{EAF7EF}
\definecolor{llmnle}{HTML}{FFF1E6}

\usepackage{wrapfig}
\usepackage{xcolor}

\usepackage{needspace}

\usepackage{booktabs}
\usepackage{multirow} 
\usepackage{subcaption}
\usepackage{placeins}
\usepackage{adjustbox}
\usepackage{tcolorbox}
\tcbuselibrary{breakable}
\tcbuselibrary{skins}
\usepackage{xcolor} 
\usepackage{listings}  
\usepackage{tikz}  
\usepackage{fancyvrb}
\usepackage{float}
\usepackage{amssymb}    
\usepackage{pifont}     
\newcommand{\cmark}{\checkmark}
\newcommand{\xmark}{\ding{55}}

\title{When Do Explanations Help In-Context Learning? A Comparative Study of Natural Language Explanation Types and Faithfulness}

\author{\name Mahdi Dhaini\thanks{Corresponding author} \email mahdi.dhaini@lmu.de\\
      \addr MaiNLP lab, CIS, LMU Munich\\  Munich Center for Machine Learning (MCML)
      \AND
      \name Adam Dejl \email adam.dejl18@imperial.ac.uk\\
      \addr Department of Computing, Imperial College London
      \AND
      \name Juraj Vladika \email juraj.vladika@tum.de\\
      \addr Technical University of Munich
      \AND
       \name Volkan Özer \email volkan.oezer@tum.de\\
      \addr Technical University of Munich, Munich Center for Machine Learning (MCML)
      \AND
      \name Barbara Plank \email b.plank@lmu.de\\
      \addr MaiNLP lab, CIS, LMU Munich\\  Munich Center for Machine Learning (MCML)
      \AND
       \name Gjergji Kasneci \email gjergji.kasneci@tum.de\\
      \addr Technical University of Munich, Munich Center for Machine Learning (MCML)
      }

\begin{document}

\maketitle

\begin{abstract}
Natural language explanations (NLEs) are increasingly used as \emph{inputs}, for example, as few-shot rationales that influence model behavior in in-context learning (ICL). However, it remains unclear how different types of NLEs compare in their effects on downstream model performance in explanation-augmented prompting.
Therefore, we provide a comparative evaluation across six benchmarks and four instruction-tuned models, studying how NLE \emph{source} (human-written when available, self-generated explanations, generated by an external LLM) and NLE \emph{selection} (random vs faithfulness-based filtering) affect downstream utility of NLEs when used in ICL settings. Our extensive evaluation shows that, on classification-style benchmarks, adding NLEs to few-shot prompts often improves accuracy over few-shot prompting without explanations; among NLE sources, externally generated LLM-NLEs often provide strong downstream utility and remain competitive with human rationales where both are available, whereas self-NLEs are more sensitive to the selection strategy. On math reasoning, the effects are more model- and source-dependent.
We further show that faithfulness-based selection of self-NLEs yields small average gains overall, but can improve or reduce performance depending on the metric, task, and model. Different faithfulness metrics can disagree substantially, affecting which self-NLE examples are selected and their downstream predictive utility. Robustness tests with randomly swapped and out-of-distribution rationales indicate partial robustness, suggesting that semantic alignment contributes
to performance gains. Overall, our results provide insights for selecting and reporting explanations that influence model behavior in practical prompting pipelines.
\end{abstract}

\section{Introduction}
Natural language explanations (NLEs), also referred to as free-text rationales,
have emerged as a prominent approach to improving the interpretability of language model (LM) predictions \citep{nle_survey_2023, wiegreffe2021teach, bhan-2024-self-amplify, icann}. As natural-language justifications for model predictions, NLEs are widely used to communicate model reasoning \footnote{We do not intend anthropomorphizing; but treat explanations as one possible window into model ``reasoning,'' while 
ledging that they may provide a partial (or weak) reflection of the processes underlying predictions. Still, NLEs are widely used and therefore we study their impact in ICL settings.} to users. Beyond interpretability, however, NLEs may also serve a functional role: recent work suggests that providing explanations during inference can improve model performance \citep{icann}. \emph{Human-annotated rationales} are often considered a gold standard, but they are expensive, slow to obtain, and subject to annotation bias and inconsistency~\citep{yao-2023-human-explanations-helpful,hartmann-2022-survey-human-explanation-performance}. An alternative is to generate explanations automatically, either via self-explanations, in which the model justifies its own predictions, or by prompting an auxiliary LLM to generate rationales~\citep{mishra-etal-2024-characterizing-rationalizers,wang-etal-2025-cross-refine,wei-jie-etal-2024-interpretable-reasoning-nle}.

In parallel, \emph{in-context learning} (ICL) has emerged as one of the key LLM capabilities, enabling task adaptation via example-driven prompts without parameter updates~\citep{Liu-2023-prompting-llms-survey}. While ICL has shown strong performance on reasoning tasks, the impact of explanations within few-shot prompts remains underexplored. It is unclear whether different types of NLEs---human-annotated, self-generated, or LLM-generated---differ in their ability to improve model predictions when used as in-context exemplars. This distinction is practically important because these sources differ in cost and operational complexity: human rationales are expensive and can be scarce, external LLM rationales require an additional model in the loop, and self-explanations can be cheapest (depending on how they are generated) but may be less reliable, so we need evidence to guide the accuracy–cost trade-off and inform best-practice decisions in explanation-augmented prompting. Furthermore, little is known about the importance of explanation quality, or how models behave when exposed to irrelevant or misleading rationales.

NLEs are increasingly used by practitioners as part of diverse workflows: as additional context during inference to improve model performance and robustness \citep{NLE-ICL_robustness_2024, icann, bhan-2024-self-amplify}; as supervision to train models to generate explanations for their decisions \citep{camburu2018esnlinaturallanguageinference, rajani2019explain}; as synthetic rationales to bootstrap better reasoners \citep{star_2022}; and as human-facing justifications that communicate model or system decisions and can shape human decision-making \citep{XAI_justifications_2025, XAI_justification_2023, Alufaisan2021}. NLEs also appear in accountability and auditing practices, for example as documentation artifacts and evidence-like narratives supporting oversight \citep{accountability_2020, Mitchell_model_cards_2019}.
 
Despite this widespread use, we still \emph{lack systematic evidence} about how \emph{NLE source} (e.g., human, self-generated, or externally LLM-generated) and \emph{NLE quality} (particularly \emph{faithfulness}), affect the \emph{utility} of NLEs when they are used to influence model behavior, such as in explanation-augmented ICL. This gap is particularly consequential because NLEs can act as behavior-shaping inputs: they may improve performance when well-aligned, but can also mislead models or users when low-quality, mismatched, or unfaithful. Broadly, faithfulness refers to whether an explanation accurately reflects a model's decision-making process~\citep{jacovi-goldberg-2020-faithfulness}.

In this work, we investigate key design choices for selecting NLEs in explanation-augmented prompting. We operationalize this objective through the \emph{\textbf{predictive utility}} of NLEs in ICL, \emph{defined as the change in downstream task performance}, such as accuracy, when NLEs are included as few-shot rationales relative to few-shot prompting without explanations. Concretely, we (i) compare NLE sources to characterize their predictive utility across the evaluated tasks and models, (ii) examine how selecting self-explanations according to measured faithfulness influences their utility and how sensitive this selection is to the choice of faithfulness metric, and (iii) test robustness to rationale misalignment using randomly swapped and out-of-distribution rationales. Taken together, we treat NLEs as components of the prompt that can shape model behavior and examine NLE source, faithfulness-based selection, and semantic alignment as important design choices in explanation-augmented ICL.  

Our work tackles the following research questions: \\\\
\noindent\textbf{\textit{RQ1}}: How do different NLE sources compare in predictive utility when used as in-context rationales across tasks and models?\\
   \noindent\textbf{\textit{RQ2}}: How does the faithfulness of self-NLEs relate to their utility in ICL, and how sensitive is faithfulness-based selection to the choice of metric?\\
   \noindent \textbf{\textit{RQ3}}: How robust are NLE-augmented prompts to semantic misalignment (randomly swapped and out-of-distribution NLEs)?

Our main contributions are:

\begin{itemize}
\item We present a comprehensive comparative study of NLE types in ICL, where we evaluate human, self-, and LLM-generated NLEs as in-context rationales across six diverse reasoning benchmarks and four models to characterize when NLEs-augmented prompting helps. 
\item We introduce a modular evaluation framework that combines error-based sample selection, explanation quality scoring, and prompt construction to compare the impact of NLE source and selection strategy. To facilitate reproducibility and future work, we will release all code, explanation datasets, and prompts used in this study upon acceptance.
\item We show that externally LLM-generated NLEs typically yield the largest gains on classification-style benchmarks, while on math reasoning, the benefit is smaller and more model- and source-dependent.
\item We show that faithfulness-based selection can be beneficial but is metric- and dataset-dependent. We also find that NLE faithfulness metrics can disagree substantially and that this disagreement affects which self-NLEs are selected and how much they help in ICL.
\item We introduce misalignment stress tests and show that semantic alignment between NLEs and examples is important. 
\end{itemize}

\section{Background and Related Work}

\paragraph{Explainable datasets and Human-annotated Rationales}
The growing interest in Explainable NLP is evident from the increasing number of surveys on the topic \citep{danilevsky-xnlp-survey-2020,madsen-xnlp-survey-2023,Zhao-xnlp-survey-2024,Luo-2024-local-xnlp-survey}. This has also led to the introduction of several \textit{explainable datasets} that include explanations for each labeled sample across diverse NLP tasks \citep{mathew2021hatexplain,talmor2020cos}. \citet{wiegreffe2021teach} provide an extensive review of such datasets, many of which contain human-annotated explanations.
However, reliance on human annotators for textual explanations introduces several challenges. Collecting high-quality human-annotated explanations is generally time-consuming and resource-intensive and particularly not automatable \citep{hartmann-2022-survey-human-explanation-performance}. Moreover, recent research has raised specific concerns regarding human annotations: the reliability of such explanations may be compromised due to subjective and inconsistent judgments by annotators \citep{yao-2023-human-explanations-helpful}. In some cases, human-provided explanations may not improve---and can even impair---model performance \citep{yao-2023-human-explanations-helpful}. The diversity of explanation types introduces additional complications \citep{tan2021diversity} and may only improve performance for a constrained subset of LLMs with specific model sizes \citep{wei2022chain}.

\paragraph{LLM-generated NLEs}
Due to the limitations of human-annotated explanations, recent research has explored using LLMs to generate NLEs \cite{mishra-etal-2024-characterizing-rationalizers, wang-etal-2025-cross-refine}. Compared to traditional post-hoc feature attribution methods \citep{madsen-xnlp-survey-2023}, NLEs provide human-readable justifications, which can enhance transparency and user understanding.
Recent research has explored LLM-generated \textbf{self-explanations}, where the same LLM used for inference or evaluation generates a (self-)explanation for its prediction. These explanations can take the form of complete natural language text \citep{bhan-2024-self-amplify, madsen-2024-self-explanations-faithful} or salient tokens that resemble the output of post-hoc feature attribution methods \citep{bhan-2024-self-amplify}. Self-explanations are considered post-hoc explanations, as they are generated after the model has made its prediction.
However, the quality of self-explanations has been questioned in prior research, with some studies arguing that self-explanations are not always faithful to the reasoning processes of LMs \citep{madsen-2024-self-explanations-faithful}. Given the importance of assessing the quality of NLEs, including self-explanations, recent work has proposed various approaches to evaluate the faithfulness of NLEs \citep{atanasova-2023-faitfulness-nle, lext-2025-facct, madsen-2024-self-explanations-faithful}. 

\paragraph{Leveraging Explanation to Improve Performance of LMs}
One key aspect of leveraging explanations, particularly NLEs, including self-explanations, is their potential to improve the performance of LLMs, i.e., the performance gains resulting from incorporating these explanations.  
\citet{yao-2023-human-explanations-helpful} investigated how incorporating human-annotated rationales into the input can affect the performance of two pre-trained language models (PLMs), BART and T5. \citet{hartmann-2022-survey-human-explanation-performance} reviewed studies that utilize various types of human explanations, including NLEs, to enhance PLM performance. These studies primarily focus on human-annotated NLEs and on providing such explanations during inference to PLMs. 
In a related line of work, recent studies have explored the use of explanations in ICL to enhance the reasoning capabilities of LLMs. \citet{krishna-2023-amplify} introduced \textit{AMPLIFY}, a framework that leverages post-hoc feature attribution
methods applied to PLMs, used as proxy models for LLMs, to construct few-shot templates that serve as corrective signals, thereby improving LLM performance on reasoning tasks. However, their methodology is constrained by the use of computationally expensive post-hoc attribution techniques and reliance on proxy models (i.e., auxiliary PLMs) for constructing few-shot examples.
\citet{bhan-2024-self-amplify} proposed \textit{self-AMPLIFY}, which eliminates the need for auxiliary models by employing two forms of post-hoc self-explanations: top salient tokens and post-hoc chain-of-thought (CoT) rationales generated by small language models (SLMs). They investigated how these explanations affect the performance of SLMs on reasoning tasks. While self-AMPLIFY mitigates some of the computational burdens of AMPLIFY and offers an automated framework for rationale generation, it focuses exclusively on self-explanations without assessing their quality. Moreover, the study is limited to small language models and primarily compares performance against the original AMPLIFY framework.

In contrast to the previous literature, this paper provides the first comprehensive comparative study to investigate key design choices for selecting NLEs when they are used as \emph{behavior-shaping inputs} in explanation-augmented prompting. We evaluate the predictive utility of NLEs from three sources, self-generated, human-annotated (when available), and externally LLM-generated, when included as in-context rationales for both SLMs and LLMs. We further analyze how the \emph{quality} of self-explanations, measured via faithfulness, affects their utility, and we study how different selection strategies (random vs.\ faithfulness-based filtering) change which NLEs are included and how much they help. Our study focuses on NLEs due to their practical prevalence in deployed workflows and their accessibility as a steering signal for prompting across model scales.

\section{Our Approach - Comparative Study Setup} \label{sec:our-approach}

\begin{figure*}[ht!]
    \centering
    \includegraphics[width=0.99\linewidth]{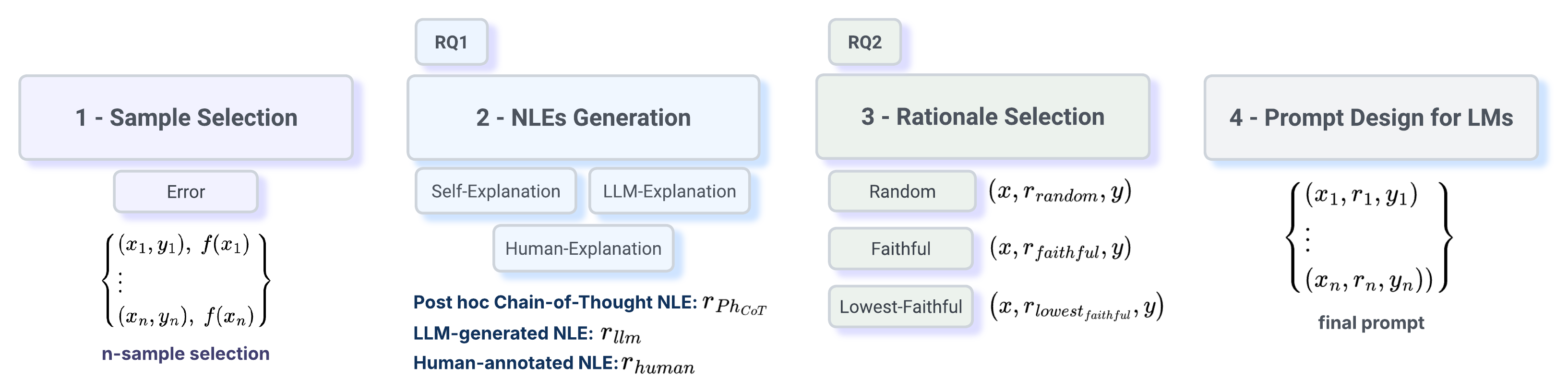}
    \caption{Overview of Experimental Setup.}
    \label{fig:experimental-setup-overview}
\end{figure*}

Figure \ref{fig:experimental-setup-overview} presents a summarized overview of our experimental framework (or comparative setup). It consists of 
four
main steps to 
investigate how NLE source, selection and alignment affect LM performance in ICL: (1) Few-shot sample selection, where we select $n$ few shot samples based on \textit{error} selection strategy.
(2) NLEs generation using either \emph{self-explanations}, 
\emph{LLM-explanations},
or \emph{human-annotated rationales}. (3) NLE selection, which resamples rationales and their respective $(x, y)$ pairs either \emph{randomly} or based on the \emph{highest-} or \emph{lowest-faithful} rationales. 
(4) Final prompt design: the final ICL prompt is constructed using the selected samples and their associated rationales.

\subsection{Few-shot Samples Selection}
In sample selection, we follow prior work that emphasizes choosing misclassified examples, since these samples can serve as corrective signals to help the language model avoid similar errors on the test set \citep{krishna-2023-amplify,bhan-2024-self-amplify}. We adopt the $n$-shot sampling strategy of \citet{bhan-2024-self-amplify}, in which \emph{error} samples for a given model~$f$ are those misclassified by~$f$ in a zero-shot setting (e.g., standard input-output prompting). In other words, the error samples set includes $(x,y)$ from the dataset if $f(x) \neq y$. 

\subsection{Natural Language Explanation Generation}
In this step, we consider three different strategies for obtaining NLEs: (1) Human-annotated explanations (human-NLEs), (2) Self-generated natural language explanations (self-NLEs), and (3) LLM-generated explanations (LLM-NLEs). By comparing multiple NLE \emph{sources}, we evaluate how NLE \emph{types} differ in their predictive utility on LLMs in ICL.

\paragraph{Human-NLEs}
We utilize \emph{human-NLEs} from explainable datasets, in which human annotators provide rationales for the correct label of each input. Given that human-annotated explanations have traditionally served as the gold standard for both evaluation and training of models to produce human-like rationales \citep{wiegreffe2021teach}, we assess their effectiveness in improving LLM reasoning performance. We only use human explanations when provided in the given dataset, and denote these as \(r_{\mathrm{human}}\).

\paragraph{Self-NLEs}
\emph{Self-NLEs} are generated by the same model \(f\) that is evaluated under the ICL prompt. We use the post-hoc chain-of-thought explanation method Ph-CoT introduced by \cite{bhan-2024-self-amplify}, where the model first makes a prediction and then produces a \emph{step-by-step} CoT-style explanation 
(e.g., ``generate a ⟨\(n_{steps}\)⟩-step explanation: \(step_1\), \(step_2\),...,\(step_n\)'')
for that prediction. 
However, for misclassified samples in the \emph{error} set, we do not use the original post-hoc explanations as-is. Instead, we regenerate self-explanations conditioned on the gold label. This is to ensure a \emph{fair} comparison between self-NLEs and \emph{human-} and \emph{LLM-generated} NLEs, which are also produced with the correct label, and to frame them as corrective rationales for the error-based prompting setup.
We also run an ablation using the original post-hoc self-explanations conditioned on the model's incorrect prediction, which shows that the gold-conditioned variant improves utility on average across the evaluated settings (Table~\ref{tab:selfexp-stepB-minus-stepA} in Appendix \ref{app:self-exp-step}). We denote these explanations as \(r_{\mathrm{Ph_{CoT}}}\).

\paragraph{LLM-NLEs}
We also leverage LLMs capabilities to generate NLEs in an annotation-style setup. This setup mimics human annotations; instead of relying on human annotators to write rationales for the correct label, we instruct an (external) explainer model \(f_{\mathrm{explainer}}\) to generate an explanation for a given input-output pair \((x,y)\) (e.g., ``generate a concise explanation'').
This differs from self-NLEs in the generation setup where LLM-NLEs are produced by an explainer model \(f_{\mathrm{explainer}}\) for \((x,y)\), independently of the evaluated model's prediction process, whereas self-NLEs are produced by the evaluated model \(f\) through the self-explanation/Ph-CoT pipeline. The explainer \(f_{\mathrm{explainer}}\) need not be the same model as \(f\), and when it is (e.g. \(f_{\mathrm{explainer}}\)=$f$), the condition remains distinct because the explanation is generated under an annotation-style prompt rather than through the Ph-CoT process. LLM-NLEs can therefore be \emph{generated once per dataset and reused across evaluation models}, avoiding the need to regenerate explanations separately for each evaluated model, which makes them practical for explanation-augmented ICL, since the explainer model is used only once for NLE generation. We denote these explanations as \(r_{\mathrm{llm}}\) and refer to them as \emph{LLM-NLEs} in this paper to distinguish them from \emph{self-NLEs}, even though both are generated by LLMs.

\subsection{NLE Selection}
\label{sec:nle-selection}

In this step, and inspired by RQ2, we designed three setups for selecting NLEs. Each of these setups for the selection of NLEs affects the final few-shot samples \((x,r,y)\) selected for the final prompt. We employ the following setups: 

\textbf{ (1) Random Selection} \textbf{(Setting 1)}: Similar to \cite{bhan-2024-self-amplify}, we select the few-shot examples $(x,r,y)$ randomly with no assessment of selected NLEs. 

\textbf{ (2) Most-Faithful Self-NLEs} \textbf{(Setting 2):} To study whether measured faithfulness affects the utility of self-NLEs, we rank self-NLEs based on their faithfulness. We compute faithfulness scores using two recent NLE faithfulness metrics, and refer to them as \(fm_1\) \citep{lext-2025-facct} and \(fm_2\) \citep{madsen-2024-self-explanations-faithful} (see \S \ref{sec:eval_faith} for details), and rank the self-NLEs separately under each metric, producing two selection variants: \emph{Faithful-\(fm_1\)} and \emph{Faithful-\(fm_2\)}.
We then select the top-$n$ most faithful NLEs along with their corresponding $(x,y)$ pairs for the few-shot template. For human and LLM-generated NLEs, we utilize the same $(x,y)$ pairs selected for the self-NLEs for fair comparison, but with their respective explanations. This setup tests whether highly faithful explanations outperform random ones in improving performance and also the influence of faithfulness metric choice.

\textbf{(3) Lowest-Faithful Self-NLEs (Setting 3):} This setup mirrors Setting~2 but selects the $n$ self-NLE examples with the lowest measured faithfulness under each metric, yielding \emph{Lowest-Faithful-$fm_1$} and \emph{Lowest-Faithful-$fm_2$} variants. This setting provides a contrastive condition for examining whether selecting examples with lower-faithfulness self-NLEs is associated with lower downstream predictive utility. Comparing the highest- and lowest-faithfulness conditions also allows us to assess whether the relationship between measured faithfulness and utility is consistent and monotonic across metrics, tasks, and models. 
Setting~3 is loosely inspired by informativeness-based active learning, which prioritizes uncertain or challenging instances for their potential learning value \citep{zhang-2022-active-learning-survey}. Analogously, we test whether examples with low self-NLE faithfulness provide corrective signals in ICL.

In addition to the three main selection settings, we include two rationale-misalignment settings (ablations) to address RQ3 and assess robustness to irrelevant or cross-domain rationales. 
In \textbf{Random Rationales} \textbf{(Setting 4)}, we replace each rationale with a randomly selected rationale from the same dataset (e.g., using rationale \(r_5\) from \((x_5,y_5)\) with pair \((x_2,y_2)\)). This tests whether gains depend on example-specific rationale alignment rather than only on the presence or format of rationales. 
In \textbf{OOD Rationales} \textbf{(Setting 5)}, we replace rationales with explanations drawn from a different dataset
(e.g., an \textsc{ecqa} example \((x_i,y_i)\) is paired with \(r_j^{*}\) from an e-\textsc{snli} example \((x_j^{*},y_j^{*})\)).
This creates a stronger cross-domain mismatch and allows us to evaluate robustness to out-of-distribution rationale noise.
We apply this setting only with LLM-NLEs due to computational constraints. Full construction details are provided in Appendix~\ref{app:rq-3-random-ood}. 
Our systematic comparison of the different setups explores how explanation type and selection criteria shape predictive utility in SLMs and LLMs.

\subsection{Final Prompt Design for LMs}
In the final step, we construct the complete prompt for the model in the ICL setting. The prompt consists of:  
(1) a \textit{preprompt}, which contains the instructions to generate an answer and an explanation. The preprompt differs between the self-NLE setup and the LLM- and human-NLE setups; and  
(2) the $n-$few-shot samples, i.e., the \((x, r, y)\) triplets selected as described in the previous steps. We provide examples of the different prompt structures used in Appendix~\ref{app:preprompts-final-prompts-examples}.

\section{Experimental Setup}

\subsection{Datasets}
We evaluate the performance of LLMs on six popular datasets. Among these, we consider two explainable datasets:   
(1) \textbf{ECQA}~\citep{aggarwal-etal-2021-explanations}, an extension of CommonsenseQA~\citep{talmor2020cos} that includes human-written rationales and requires models to justify their answers using structured commonsense knowledge; and (2) \textbf{e-SNLI}~\citep{camburu2018esnlinaturallanguageinference}, which contains premise–hypothesis pairs labeled as \textit{entailment}, \textit{neutral}, or \textit{contradiction}, depending on the logical relationship between the two. In addition, we include three tasks from the Big-Bench-Hard benchmark \citep{big-bench-hard}:  
(3) \textbf{SNARKS}, where the model must identify the sarcastic sentence from two nearly identical options; (4) \textbf{Causal Judgment} (\textsc{cj}), which involves reading a short story and determining how a typical person would answer a causal question about it; (5) \textbf{Boolean Expressions} (\textsc{Bool}), which tests the model's ability to evaluate the truth value of randomly generated Boolean expressions composed of constants and basic Boolean operators. Finally, we also evaluate on (6) \textbf{GSM8K} ~\citep{gsm8k_2021}, a mathematical reasoning dataset containing diverse grade school math word problems that require multi-step reasoning in natural language. 
Together, these benchmarks provide complementary coverage of commonsense, semantic inference, pragmatics, causal reasoning, symbolic logic, and complex multi-step math, enabling a controlled evaluation of how NLE source and selection (including faithfulness filtering and misalignment tests) affect predictive utility across diverse reasoning settings. Additional dataset details are provided in Appendix ~\ref{app:dataset-details}.

\subsection{Models}
For evaluation, we use instruction-tuned autoregressive LLMs of varying sizes. Specifically, we experiment with \texttt{GPT-4o-mini} \citep{openai2024gpt4ocard}, \texttt{Llama-3.1-8B}, and \texttt{Llama-3.3-70B} \cite{grattafiori2024llama3herdmodels}, and \texttt{Mistral-7B-Instruct-v0.3} \citep{mistral2023}, where \texttt{Llama-8B}, \texttt{Mistral-7B} are considered SLMs, while \texttt{Llama-70B} represents a significantly larger model. The intentional selection of these models of varying sizes enables us to compare and analyze the influence of model size on our results, particularly examining differences between SLMs (\texttt{Llama-8B}, \texttt{Mistral-7B}) and LLMs (\texttt{Llama-70B}) from same and different families. More specifically, we investigate how different experimental setups and configurations affect SLMs compared to LLMs. For GPT-based models, we use the OpenAI API, whereas for the Llama and Mistral models we rely on the Together AI APIs. To better simulate realistic prompt conditions, we set \(n\) to 6 and thus employ a 6-shot few-shot prompting setup across all models
consistent with prior work \citep{bhan-2024-self-amplify} (and also due to computational cost of our comprehensive pipeline). 
Through this diverse set of models and experimental settings, we comprehensively assess the scalability and robustness of our method across varying model capacities and architectural complexities.  
\\ \textbf{Models for Generating NLEs:} To get LLM-generated NLEs we employ two models, \texttt{GPT-4o-mini} and \texttt{o3-mini}. Using multiple explainer models allows us to diversify our selection and assess whether the choice of LLM affects the predictive utility of the generated LLM-NLEs.  
To generate self-NLEs, the same models used for evaluation (\texttt{4o-mini, \texttt{Llama-8B}, \texttt{Llama-70B}, and \texttt{Mistral-7B}}) are used to generate self-NLEs. 
We show examples of the different NLEs generated by each strategy and its variants for the datasets in 
Figures \ref{app:fig-explanations-examples-ecqa}-\ref{app:fig-explanations-examples-cj} in Appendix \ref{app:nle-examples-all-datasets}. 

\subsection{Evaluating Self-NLEs Faithfulness} \label{sec:eval_faith}
We use \emph{faithfulness} to refer to the extent to which a self-NLE reflects and is consistent with the model's answer-generating behavior, and we measure it quantitatively through two automated metrics $fm_1$ \& $fm_2$. 
$fm_1$ is the faithfulness metric in the LExT framework proposed by \citet{lext-2025-facct}, which combines three sub-metrics capturing different dimensions of faithfulness: (1) \textit{Question Answer Generation} (QAG), which assesses whether questions, generated from the explanation by the same model, can be answered using that explanation; (2) \textit{Counterfactual Stability}, which tests whether the model's prediction changes when the explanation is rewritten by the same model to support an alternative label; and (3) \textit{Contextual Faithfulness}, which removes important keywords identified by the model itself from the input context and tests whether the model recognizes that the remaining information is insufficient. Each sub-metric produces a score in $[0,1]$, and the three scores are averaged to obtain the final $fm_1$ score, also in $[0,1]$. 
For comparison with binary $fm_2$, we convert $fm_1$ into a relative binary label using a dataset-specific 75th percentile, treating the highest-scoring quartile as faithful and the remainder as unfaithful (see \S\ref{app:lext-metric-details} for more details).

For \emph{$fm_2$}, we adapt the self-consistency-based faithfulness framework proposed in \citet{madsen-2024-self-explanations-faithful} to self-NLEs. 
The framework tests whether a self-explanation is behaviorally consistent with the model's prediction under a counterfactual intervention.
We apply this metric to the self-NLE, which is minimally edited to support an alternative label, and the model is then asked, in a separate inference session, to answer the original question using the edited self-NLE as context. If the prediction changes to the intended alternative label, the self-NLE is marked faithful; otherwise, it is marked unfaithful. 
$fm_2$ provides a binary faithful/unfaithful score per self-NLE (see \S\ref{app:fm2-details} for more details).
We provide
examples of prompt templates in Appendix \ref{app:faithfulness-evaluation-prompts} and \ref{app:faithfulness-evaluation-prompts-fm2}.

\subsection{Baseline Methods}
To comprehensively evaluate our designed setups, we compare their performance against several baseline methods: Zero-Shot (ZS) \citep{srivastava2023imitationgamequantifyingextrapolating}, which uses standard input-output prompting; Few-Shot (FS), where the prompt includes only a few input-output samples \((x,y)\) without rationales; and zero-shot CoT \citep{zero-shot-cot}, where we prompt the LLM to elicit a step-by-step natural-language rationale followed by the final answer. 
For the FS baseline, we consider FS-Random (FS-R), following Setup 1 with randomly selected few-shot samples but excluding rationales. These baselines allow us to quantify the added value of including NLEs as in-context exemplars beyond standard prompting strategies. 

\section{Results and Analysis}

\subsection{Comparing NLE types and Setups (RQ1)}

\paragraph{\textbf{Comparison among different NLE types.}} 
Across datasets, we compare explanation-augmented prompting (self-NLEs, LLM-NLEs, and, where available, human-NLEs) against the baselines we specified. Table~\ref{tab:rq1-baseline-random-comp-five-runs} reports the main random-selection results averaged across models; Figure~\ref{fig:random_comparison_plot} and Table~\ref{tab:random-nle-comp} in Appendix~\ref{app:further_results} provide the corresponding visualization and model-specific results. 
Our primary goal is to compare the predictive utility of different NLE sources in an ICL setting. Overall, on the classification-style benchmarks (\textsc{ecqa}, e-\textsc{snli}, \textsc{snarks}, \textsc{boolean}, and \textsc{cj}), incorporating NLEs into the few-shot prompt often improves accuracy relative to FS-R, especially when using LLM- or human-generated explanations; self-NLEs can also help, but with more variability. Among NLE sources, LLM-NLEs (\texttt{GPT-4o-mini} or \texttt{o3-mini}) often yield the strongest performance and are consistently competitive across datasets. 

\begin{table*}[h!]
\caption{NLE Type comparison under random selection (RQ1). Values represent performance ($mean_{\text{standard deviation}}$) over five runs for baseline methods and random-selection explanation setup, averaged over the four evaluated models for each dataset. Bold indicates the best-performing method per dataset.}
\centering
\scriptsize
\setlength{\tabcolsep}{4pt}
\begin{tabular}{lccc cccc}
\toprule
& \multicolumn{3}{c}{Baselines} & \multicolumn{4}{c}{Random} \\
\cmidrule(lr){2-4} \cmidrule(lr){5-8}
Dataset & ZS & FS-R & CoT & Self-NLE & Human-NLE & LLM-4o & LLM-o3 \\
\midrule
\textsc{ecqa}    & $0.760_{0.011}$ & $0.791_{0.014}$ & $0.756_{0.015}$ & $0.753_{0.045}$ & $0.773_{0.029}$ & $\mathbf{0.799}_{0.004}$ & $0.780_{0.007}$ \\
e-\textsc{snli}  & $0.593_{0.008}$ & $0.724_{0.017}$ & $0.591_{0.012}$ & $0.711_{0.026}$ & $\mathbf{0.726}_{0.038}$ & $0.701_{0.009}$ & $0.608_{0.021}$ \\
\textsc{snarks}  & $0.566_{0.014}$ & $0.587_{0.021}$ & $0.752_{0.019}$ & $0.556_{0.014}$ & -- & $0.801_{0.009}$ & $\mathbf{0.814}_{0.016}$ \\
\textsc{boolean} & $0.516_{0.009}$ & $0.671_{0.011}$ & $0.806_{0.011}$ & $0.877_{0.033}$ & -- & $0.900_{0.003}$ & $\mathbf{0.916}_{0.011}$ \\
\textsc{cj}      & $0.587_{0.013}$ & $0.619_{0.018}$ & $0.648_{0.016}$ & $0.650_{0.023}$ & -- & $0.650_{0.015}$ & $\mathbf{0.654}_{0.008}$ \\
\textsc{gsm8k}   & $0.246_{0.005}$ & $0.724_{0.015}$ & $\mathbf{0.809}_{0.004}$ & $0.690_{0.007}$ & -- & $0.768_{0.008}$ & $0.765_{0.005}$ \\
\bottomrule
\end{tabular}
\label{tab:rq1-baseline-random-comp-five-runs}
\end{table*}

When human-NLEs are available, this pattern holds for \textsc{ecqa} but not for e-\textsc{snli}, where human-annotated NLEs are slightly stronger on average, suggesting that human rationales provide especially valuable corrective signals for this NLI task. In contrast, self-NLEs are more variable: while they frequently improve over ZS and can be comparable to or occasionally exceed FS-R, they typically lag behind LLM- (and, where available, human-) NLEs. On \textsc{gsm8k}, the same relative ordering among NLE types largely holds where LLM-NLEs are generally stronger than self-NLEs, while baseline performance is strongly driven by CoT, which is expected for multi-step math reasoning. Taken together, these results suggest that adding NLEs to few-shot prompts can provide benefits in many settings, with LLM-generated (and, where available, human) explanations being the most reliable sources. 

\FloatBarrier

\paragraph{Influence of the NLE-selection Setup (Setups 1 vs 2 vs 3).}
We now compare how different NLE selection setups, particularly moving from Setting 1 (Random) to Settings 2 (Most-Faithful) and 3 (Lowest-Faithful), impact the predictive utility of NLEs. We refer to Table \ref{tab:matrix} for a summary of the results and for more detailed results in Table \ref{tab:nle-source-selection-delta-by-dataset} and Figure \ref{fig:all_setups_comparison_plots} in Appendix \ref{app:full-results-across-setups}.  Overall, as shown in Table \ref{tab:matrix}, LLM- and human-NLEs (for e-\textsc{snli} and \textsc{ecqa}) often exhibit higher predictive utility than self-NLEs across Setups 1–3 (\textit{except} for faithful-$fm_1$/$fm_2$ on \textsc{gsm8k} and lowest-faithful $fm_2$ on \textsc{cj} where self-NLEs outperform LLM-NLEs). Thus, even when NLEs selection is driven by self-NLE faithfulness scores, LLM- and human-NLEs still show better utility in most settings. Moving across settings, LLM and human-NLEs also appear relatively stable in their predictive utility. Human-NLEs remain strong on \textsc{ecqa} and e-\textsc{snli} and are the best-performing source across all settings on e-\textsc{snli}. LLM-NLEs also maintain high absolute performance across most settings (except on \textsc{gsm8k} under Most-Faithful selection, where self-NLEs outperform LLM-NLE variants). Taken together, these results indicate that LLM- and human-NLEs often provide stronger and more stable downstream utility \emph{than} self-NLEs, even when explanations are selected using self-NLE faithfulness scores. In contrast, self-NLEs appear to benefit more selectively from faithfulness-based selection, which we discuss further in \S \ref{sec:rq2-results}. We refer the reader to  Figure \ref{fig:all-values-plots}, \ref{fig:all-values-plots2}, and Table \ref{tab:big-table-error-five-runs} in Appendix \ref{app:further_results} for \textit{per-model detailed results}.

\begin{table*}[!ht]
\centering
\scriptsize
\setlength{\tabcolsep}{5pt}
\caption{Best-performing NLE source under each selection setting (RQ1). Each cell reports the NLE source with the highest mean absolute accuracy, averaged over the four evaluation models, with the accuracy shown in parentheses. Cell colors indicate the NLE type: \colorbox{selfnle}{Self-NLE}, \colorbox{humannle}{Human-NLE}, and \colorbox{llmnle}{LLM-NLE}.}
\label{tab:matrix}
\begin{tabular}{lccccc}
\toprule
\textbf{Dataset}
& \textbf{Setting 1} 
& \multicolumn{2}{c}{\textbf{Setting 2 (Most-Faithful)}}
& \multicolumn{2}{c}{\textbf{Setting 3 (Lowest-Faithful)}} \\
\cmidrule(lr){2-2} \cmidrule(lr){3-4} \cmidrule(lr){5-6}
& \textbf{Random}
& \textbf{$fm_1$}
& \textbf{$fm_2$}
& \textbf{$fm_1$}
& \textbf{$fm_2$} \\
\midrule

\textsc{ecqa}
& \cellcolor{llmnle}\makecell{\textbf{LLM-4o}\\(0.799)}
& \cellcolor{humannle}\makecell{\textbf{Human-NLE}\\(0.774)}
& \cellcolor{humannle}\makecell{\textbf{Human-NLE}\\(0.783)}
& \cellcolor{llmnle}\makecell{\textbf{LLM-o3}\\(0.801)}
& \cellcolor{llmnle}\makecell{\textbf{LLM-o3}\\(0.806)} \\

e-\textsc{snli}
& \cellcolor{humannle}\makecell{\textbf{Human-NLE}\\(0.726)}
& \cellcolor{humannle}\makecell{\textbf{Human-NLE}\\(0.762)}
& \cellcolor{humannle}\makecell{\textbf{Human-NLE}\\(0.741)}
& \cellcolor{humannle}\makecell{\textbf{Human-NLE}\\(0.756)}
& \cellcolor{humannle}\makecell{\textbf{Human-NLE}\\(0.763)} \\

\textsc{snarks}
& \cellcolor{llmnle}\makecell{\textbf{LLM-o3}\\(0.814)}
& \cellcolor{llmnle}\makecell{\textbf{LLM-o3}\\(0.816)}
& \cellcolor{llmnle}\makecell{\textbf{LLM-o3}\\(0.818)}
& \cellcolor{llmnle}\makecell{\textbf{LLM-o3}\\(0.798)}
& \cellcolor{llmnle}\makecell{\textbf{LLM-o3}\\(0.794)} \\

\textsc{boolean}
& \cellcolor{llmnle}\makecell{\textbf{LLM-o3}\\(0.916)}
& \cellcolor{llmnle}\makecell{\textbf{LLM-4o}\\(0.926)}
& \cellcolor{llmnle}\makecell{\textbf{LLM-o3}\\(0.910)}
& \cellcolor{llmnle}\makecell{\textbf{LLM-o3}\\(0.911)}
& \cellcolor{llmnle}\makecell{\textbf{LLM-4o}\\(0.919)} \\

\textsc{cj}
& \cellcolor{llmnle}\makecell{\textbf{LLM-o3}\\(0.654)}
& \cellcolor{llmnle}\makecell{\textbf{LLM-o3}\\(0.634)}
& \cellcolor{llmnle}\makecell{\textbf{LLM-o3}\\(0.654)}
& \cellcolor{llmnle}\makecell{\textbf{LLM-o3}\\(0.647)}
& \cellcolor{selfnle}\makecell{\textbf{Self-NLE}\\(0.646)} \\

\textsc{gsm8k}
& \cellcolor{llmnle}\makecell{\textbf{LLM-4o}\\(0.768)}
& \cellcolor{selfnle}\makecell{\textbf{Self-NLE}\\(0.763)}
& \cellcolor{selfnle}\makecell{\textbf{Self-NLE}\\(0.686)}
& \cellcolor{llmnle}\makecell{\textbf{LLM-4o}\\(0.732)}
& \cellcolor{llmnle}\makecell{\textbf{LLM-4o}\\(0.766)} \\

\bottomrule
\end{tabular}
\end{table*}

\subsection{Relationship Between Measured Faithfulness and Self-NLE Predictive Utility (RQ2)} \label{sec:rq2-results}
In this experiment, we address RQ2 by measuring how the faithfulness of in-context \emph{self-NLEs} affects downstream performance and we refer to the results in Table~\ref{tab:self-exp-comp-settings-delta} (see Table~\ref{tab:stat-test-step-b-only} in Appendix~\ref{app:self-nle-across-setups} for per-model results). Overall, selecting \emph{high-faithful} self-NLEs yields small average improvements across datasets for $fm_1$ \& $fm_2$, but these aggregate gains mask variation across datasets, models, and metrics. On the classification-style benchmarks, selecting highly faithful self-NLEs can improve predictive utility (e.g., \textsc{ecqa} and e-\textsc{snli} on average), but is not uniformly beneficial across all datasets (e.g., \textsc{cj} shows slight degradations across selection strategies). On the other hand, \emph{lowest-faithfulness} selection is harmful on average (overall $-$0.035 for lowest-faithful-$fm_1$, $-$0.041 for lowest-faithful-$fm_2$), but not always: for some datasets (e.g., \textsc{snarks} and \textsc{ecqa} under lowest-faithful-$fm_2$), performance can still increase relative to random selection. On \textsc{gsm8k}, high-faithfulness selection exhibits strong
metric- and model-dependent effects. Selection according to $fm_1$ improves average performance by $0.073$, driven largely by the gain for \texttt{Llama-70B} ($\Delta=+0.350$). In contrast, $fm_2$-based selection produces almost no average change ($\Delta=-0.003$), because the substantial gain for \texttt{Llama-70B} ($\Delta=+0.371$) is offset by a large degradation for \texttt{Mistral-7B} ($\Delta=-0.405$). Selecting \emph{lowest-faithful} self-NLEs reduces performance for all evaluated models on \textsc{gsm8k}, particularly under $fm_2$. Taken together, these results show that selection based on measured self-NLE faithfulness can affect downstream predictive utility, but its effect depends on the faithfulness metric, task, and evaluated model. The choice of metric changes which NLEs are selected and is associated with different performance outcomes across settings.

\begin{table*}[h]
\caption{
Self-NLE performance under random and faithfulness-based selection (RQ2). 
\emph{Random Self-exp.} reports accuracy. 
Other columns report \(\Delta\) vs. random self-NLE selection
(($+$): improvement, ($-$): degradation). \textbf{Bold} marks the best strategy per dataset--model pair. ${\dagger}$ indicates that faithfulness-based self-NLE selection also outperforms the strongest non-NLE baseline for the same pair.
}
\centering
\scriptsize
\setlength\tabcolsep{5pt}
\begin{tabular}{l|l|c|c|c|c|c}
\toprule
\multirow{3}{*}{Dataset} & \multirow{3}{*}{Models} & Random & Faithful-fm1 & Faithful-fm2 & Lowest-Faithful-fm1 & Lowest-Faithful-fm2 \\
\cmidrule(lr){3-3} \cmidrule(lr){4-4} \cmidrule(lr){5-5} \cmidrule(lr){6-6} \cmidrule(lr){7-7}
& & Self-exp. & $\Delta$ vs Rand. & $\Delta$ vs Rand. & $\Delta$ vs Rand. & $\Delta$ vs Rand. \\
\midrule

\multirow{5}{*}{\textsc{ecqa}}
& \texttt{4o-mini}   & 0.828 & -0.004 & -0.008 & -0.004 & \textbf{+0.001$^{\dagger}$} \\
& \texttt{llama-70B}  & 0.745 & +0.037 & +0.068 & -0.086 & \textbf{+0.117$^{\dagger}$} \\
& \texttt{llama-8B}   & 0.780 & -0.021 & 0.000  & -0.550 & \textbf{+0.002} \\
& \texttt{mistral-7B} & 0.660 & +0.036 & +0.026 & +0.018 & \textbf{+0.064} \\
& Avg        & 0.753 & +0.012 & +0.022 & -0.155 & \textbf{+0.046} \\
\midrule

\multirow{5}{*}{e-\textsc{snli}}
& \texttt{4o-mini}   & 0.826 & -0.036 & -0.008 & \textbf{+0.020$^{\dagger}$} & +0.006 \\
& \texttt{llama-70B}  & 0.789 & \textbf{+0.027$^{\dagger}$} & -0.069 & -0.015 & -0.085 \\
& \texttt{llama-8B}   & 0.639 & -0.035 & -0.006 & \textbf{+0.073} & -0.067 \\
& \texttt{mistral-7B} & 0.591 & +0.051 & \textbf{+0.100} & +0.016 & 0.000 \\
& Avg        & 0.711 & +0.002 & +0.005 & \textbf{+0.024} & -0.036 \\
\midrule

\multirow{5}{*}{\textsc{snarks}}
& \texttt{4o-mini}   & 0.512 & +0.002 & +0.051 & +0.142 & \textbf{+0.263} \\
& \texttt{llama-70B}  & 0.816 & -0.022 & \textbf{+0.006} & +0.004 & -0.002 \\
& \texttt{llama-8B}   & 0.418 & +0.051 & +0.135 & -0.032 & \textbf{+0.171} \\
& \texttt{mistral-7B} & 0.477 & -0.046 & -0.036 & -0.083 & \textbf{+0.003} \\
& Avg        & 0.556 & -0.004 & +0.039 & +0.008 & \textbf{+0.109} \\
\midrule

\multirow{5}{*}{\textsc{boolean}}
& \texttt{4o-mini}   & 0.970 & -0.013 & -0.027 & -0.040 & -0.011 \\
& \texttt{llama-70B}  & 0.935 & +0.007 & \textbf{+0.041} & -0.126 & +0.037 \\
& \texttt{llama-8B}   & 0.872 & \textbf{+0.021$^{\dagger}$} & -0.020 & +0.001 & +0.001 \\
& \texttt{mistral-7B} & 0.729 & -0.046 & +0.016 & +0.037 & \textbf{+0.066$^{\dagger}$} \\
& Avg        & 0.877 & -0.008 & +0.002 & -0.032 & \textbf{+0.023$^{\dagger}$} \\
\midrule

\multirow{5}{*}{\textsc{cj}}
& \texttt{4o-mini}   & 0.695 & -0.010 & \textbf{+0.008$^{\dagger}$} & -0.023 & -0.012 \\
& \texttt{llama-70B}  & 0.650 & +0.055 & +0.028 & +0.032 & \textbf{+0.095$^{\dagger}$} \\
& \texttt{llama-8B}   & 0.640 & -0.062 & -0.041 & -0.010 & -0.085 \\
& \texttt{mistral-7B} & 0.614 & -0.052 & -0.046 & -0.025 & -0.012 \\
& Avg        & 0.650 & -0.017 & -0.013 & -0.007 & -0.004 \\
\midrule

\multirow{5}{*}{\textsc{gsm8k}}
& \texttt{4o-mini}   & 0.903 & -0.010 & \textbf{+0.013} & -0.027 & -0.526 \\
& \texttt{llama-70B}  & 0.552 & +0.350 & \textbf{+0.371} & -0.064 & -0.280 \\
& \texttt{llama-8B}   & 0.816 & \textbf{+0.009} & \textbf{+0.009} & -0.033 & -0.380 \\
& \texttt{mistral-7B} & 0.487 & -0.057 & -0.405 & -0.066 & -0.346 \\
& Avg        & 0.690 & \textbf{+0.073} & -0.003 & -0.048 & -0.383 \\
\midrule

All datasets & Avg
& 0.706 & \textbf{+0.010} & +0.009 & -0.035 & -0.041 \\
\bottomrule
\end{tabular}
\label{tab:self-exp-comp-settings-delta}
\end{table*}

\subsection{Robustness to Random and OOD Explanations (RQ3)} \label{app:rq3-results-random-ood} 
As an ablation study, in Settings~4 and~5 we stress-test how models respond to \emph{misaligned} few-shot explanations: (1) \textbf{Random rationales} (Setting 4), where rationales are swapped across in-domain \((x,y)\) pairs, and (2) \textbf{OOD rationales} (Setting 5), where rationales are drawn from mismatched datasets. Table~\ref{tab:random-adversarial-aaai} reports the average performance difference of these stress tests relative to the faithful-selection setup (Setting~2 with \(fm1\)), aggregated over five runs (with standard deviation values in Table~\ref{tab:random-adversarial-std} in Appendix~\ref{app:rq-3-random-ood}). Overall, OOD swaps consistently degrade performance, whereas in-domain random swaps have mixed effects across datasets and explanation sources. 
 
Overall in Setting 4, \textbf{randomly swapped rationales typically reduce performance} compared to faithful selection, with the magnitude depending on the dataset and explanation source. In some settings, the degradation is modest (and can even be slightly positive for self-explanations on \textsc{snarks}), 
suggesting partial robustness to in-domain rationale mismatch and that such in-domain rationales may still provide task-relevant cues.
However, the more pronounced drops observed on several datasets (e.g., e-\textsc{snli} and \textsc{boolean} for LLM-NLEs) indicate that \textbf{explanation-specific semantic alignment remains important} and that random swaps \emph{should not be treated as a harmless change}.

In the \textbf{OOD rationales} setup (Setting 5), performance \textbf{consistently degrades} relative to faithful explanations across datasets, but models retain non-trivial accuracy. This pattern supports a \emph{partial robustness} interpretation: models do not collapse under cross-domain rationales, yet they are consistently harmed by semantically mismatched explanations. Together, these stress tests 
indicate partial robustness and reinforce that \emph{semantic coherence} contributes to the gains from NLE-augmented ICL, motivating careful explanation selection. These behaviors are further illustrated by qualitative examples of final prompts and model predictions under Setups 4 and 5 (see Appendix~\ref{app:random-adv-outputs}). For instance, Figure~\ref{app:fig-adv-setting-example-ecqa-esnli-wrong} shows an incorrect prediction by \texttt{Llama-70B} on an \textsc{ecqa} sample with LLM-NLEs from e-\textsc{snli}. Similarly, Figure~\ref{app:prompt_structure_input_output_ll-explanation_4omini-bool-wrong} shows a failed \textsc{bool} prediction by \texttt{Llama-8B} with few-shot examples containing LLM rationales from \textsc{snarks}. Partial robustness examples, where the model still gets a correct output despite the inclusion of OOD NLEs, can still be seen e.g. in examples of Figures~\ref{app:prompt_structure_input_output_ll-explanation_4omini-bool-llama8b-snarks-correct} and \ref{app:fig-adv-setting-example-snarks-book-correct}. 

\begin{table}[!h]
\caption{
The \emph{difference} (\(\Delta\)) of average performance across five runs of models per dataset for Random and OOD settings \emph{vs} the faithful selection (setting 2 with \(fm_1\)) across datasets and explanation types (RQ3).
}
\centering
\scriptsize
\begin{tabular}{lccccc}
\toprule
\multirow{2}{*}{Dataset} 
  & \multicolumn{4}{c}{Random} 
  & \multirow{2}{*}{OOD} \\
\cmidrule(lr){2-5}
  & Self-exp. & Human & \texttt{4o-mini}& \texttt{o3-mini} & \\
\midrule
\textsc{ecqa}          & -0.062 & +0.002 & -0.124 & -0.002 & -0.081 \\
e-\textsc{snli}        & -0.053 & -0.106 & -0.162 & -0.128 & -0.113 \\
\textsc{snarks}        & +0.063 & -- & -0.033 & -0.061 & -0.101 \\
\textsc{bool}         & -0.098 & -- & -0.181 & -0.070 & -0.090 \\
\textsc{cj}            & -0.054 & -- & -0.086 & -0.074 & -0.029 \\
\textsc{gsm8k}         & -0.022 & -- & -0.023 & -0.048 & -0.035 \\
\midrule
All datasets  & -0.038 & -0.052 & -0.102 & -0.064 & -0.075 \\
\bottomrule
\end{tabular}
\label{tab:random-adversarial-aaai}
\end{table}

\section{Discussion}
\paragraph{\textbf{Which NLE to Use? Practical Guidance.}}
Taken together, our findings suggest the following practical guidance for selecting NLEs with high predictive utility: (1) adding NLEs to few-shot prompts often improves performance over ICL without explanations, although the gains are not universal and depend on the NLE source, selection strategy, and task. (2) external NLE sources, LLM-generated and, where available, human-written NLEs provide comparatively strong and stable predictive utility across the evaluated selection setups, whereas self-NLE utility is more sensitive to how explanations are selected. Because LLM- and human-NLEs also perform strongly under random selection on most evaluated classification-style benchmarks, they provide a practical starting point without requiring an additional faithfulness-scoring stage. (3) the \textsc{gsm8k} results suggest that faithfulness-selected self-NLEs can be particularly useful for math reasoning in some settings. This motivates further work on when faithfulness-based selection is beneficial and how to develop more reliable selection criteria across tasks and models.

\paragraph{\textbf{LLM-Generated vs. Human Explanations.}}
On the two datasets with human NLEs (\textsc{ecqa} and e-\textsc{snli}), LLM-generated explanations are broadly competitive with human-annotated rationales. On \textsc{ecqa}, LLM-generated explanations often match or outperform human explanations across several settings, whereas on e-\textsc{snli} human rationales remain slightly stronger on average and LLM-generated explanations are generally comparable rather than clearly superior. Since LLM-generated explanations can be produced automatically via prompting, they are substantially cheaper and faster than manual annotation, making them a \emph{practical alternative} for practitioners when human explanations are unavailable or expensive. At the same time, the e-\textsc{snli} results suggest that human-written rationales may still be advantageous for tasks requiring fine-grained semantic alignment.

\paragraph{\textbf{Disagreement Between NLE Faithfulness Metrics.}}
Our RQ2 findings highlight a critical issue for NLE faithfulness evaluation: \emph{faithfulness metrics can disagree substantially}. As shown in Table~\ref{tab:disagreement_analysis_faithfulness} (Appendix ~\ref{app:heatmaps-disagreement-ecqa}), the average disagreement rate is $\sim0.5$ across datasets and models: about half of NLEs receive conflicting faithful vs. unfaithful labels. The direction of disagreement is dataset-dependent yet consistent within each dataset across models, indicating that $fm_1$ and $fm_2$ are not interchangeable and likely capture different operational aspects of self-NLE faithfulness. Although some disagreement may be expected because the two metrics are grounded in different faithfulness checks, the magnitude and consistency of the disagreement show that metric choice matters in practice, especially when such metrics are used to evaluate or select faithful self-NLEs. This implies that \emph{metric choice is a substantive decision}, especially in settings that require faithful explanations or rely on NLEs for data augmentation and performance improvements~\citep{chen-etal-2025-rose, icann}. \emph{In practice, a single metric should not be treated as ground truth for self-explanation faithfulness;} disagreement should be interpreted as measurement uncertainty and, when possible, quantified and reported. This is particularly relevant for practitioners relying on evaluation toolboxes that implement only one faithfulness metric. Given the scarcity of metrics for measuring faithfulness in self-NLEs, this level of disagreement is concerning and warrants further investigation by the Explainable NLP community. More broadly, these findings motivate mechanistic analyses and more rigorous evaluations of faithfulness metrics for NLEs, and connect to broader challenges in explainable NLP where automated metrics are frequently used to assess explanation quality.  We present further results about this disagreement in Appendix~\ref{app:heatmaps-disagreement-ecqa}.

\paragraph{\textbf{Faithfulness-Based NLE Selection Can Help, but Not Always.}}
Selecting self-NLEs by measured faithfulness yields small average improvements over random selection, but the relationship between faithfulness and predictive utility of self-NLEs is neither uniform nor consistently monotonic. High-faithfulness selection improves downstream performance in some model--dataset combinations, while lowest-faithfulness selection is harmful on average but can still improve performance in some classification settings. Consistent with the metric-disagreement observation, $fm_1$ and $fm_2$ can lead to different (and sometimes opposite) outcomes, indicating that faithfulness is not a metric-agnostic proxy for downstream utility. This is particularly evident on \textsc{gsm8k} where high-faithfulness selection using $fm_1$ improves average performance, whereas using $fm_2$ produces almost no average change because its effects differ sharply across models, substantially benefiting \texttt{Llama-70B} while strongly degrading \texttt{Mistral-7B}. Lowest-faithfulness selection is also especially harmful on this task, particularly under $fm_2$. In particular, \texttt{Mistral-7B} results on \textsc{gsm8k} suggest that when a model’s self-explanations utility is low, $fm_2$-based filtering can amplify selection errors, and sharply reduce downstream utility, an effect which could be related to this particular model's low performance on this dataset but is still worth further investigation in future work.
Overall, faithfulness can provide a useful selection signal, but its effectiveness depends on the metric, task, and evaluated model. Therefore, a single faithfulness metric should not be assumed to provide a consistently reliable filtering criterion.

\paragraph{\textbf{Insights on Robustness and Misalignment.}}
Our random and OOD NLEs stress tests
provide additional insights into when explanation-augmented ICL is robust to misalignment where both randomly swapped in-domain rationales and out-of-domain rationales generally \emph{reduce} performance on average. At the same time, models retain reasonable performance under these perturbations, indicating partial robustness to rationale misalignment. OOD swaps are consistently harmful across datasets, whereas the effects of in-domain random swaps are more variable. This variability may partly arise because in-domain rationales may still provide task-relevant cues or reasoning patterns despite being paired with different examples.
Overall, these findings suggest that
\emph{example-specific semantic alignment remains important}; thus, NLE-augmented prompting benefits from careful rationale selection and stress-testing when used in ICL settings.
Practically, this suggests that using non-aligned NLEs (even when drawn from the same dataset) can reduce or negate the benefits of explanation-augmented prompting. 
When deploying NLE-augmented ICL, practitioners should therefore prefer explanations that are \emph{semantically aligned with each few-shot example} and treat random swapping as a \emph{stress test rather than a viable selection strategy}.

\paragraph{\textbf{NLEs vs. CoT Prompting.}}
CoT prompting is a widely used approach for eliciting step-by-step reasoning from LLMs and is a strong baseline in our experiments, particularly on multi-step math reasoning (e.g., \textsc{gsm8k}). At the same time, our results in Table \ref{tab:rq1-baseline-random-comp-five-runs} (and Tables \ref{tab:big-table-error-five-runs}, \ref{tab:avg-results-error-five-runs} in Appendix \ref{app:full-results-across-setups}) show that incorporating structured NLEs as few-shot exemplars can be highly effective: on several classification-style benchmarks, LLM- or human-generated NLEs mostly outperform CoT in a number of model–dataset settings, while also consistently improving over standard few-shot prompting. Beyond accuracy, NLEs offer practical advantages: they provide reusable reasoning patterns that can be precomputed (e.g., via an external LLM), enabling modular prompt design and decoupling explanation construction from inference-time generation. Overall, these findings suggest that NLE-augmented prompting and CoT are complementary strategies whose relative effectiveness is task-dependent: CoT is particularly effective for step-by-step math reasoning, whereas well-aligned NLE exemplars can serve as a flexible and competitive alternative on classification-style reasoning tasks, especially under constrained or latency-sensitive settings.

\paragraph{\textbf{Impact of Model Size on Explanation Sensitivity.}}
As shown in Table~\ref{tab:big-table-error-five-runs} (and Figure~\ref{fig:llama_comparison} in Appendix \ref{app:size-sensitivity}), the relative accuracy gains from explanation-augmented prompting are larger for smaller models. \texttt{Llama-8B} and \texttt{Mistral-7B} (except on \textsc{gsm8k}) show more pronounced performance gains when given high-quality explanations, especially under faithful selection regimes. This suggests that external reasoning support compensates for limited internal abstraction capabilities. In contrast, \texttt{Llama-70B} performs more robustly across all setups, with less variance between strategies, indicating stronger intrinsic reasoning. These findings align with prior observations that larger models internalize more abstract reasoning patterns and benefit less from external scaffolding. Nonetheless, both (smaller) models consistently improve with well-selected explanations, reinforcing the utility of prompt-based guidance at all scales. We also provide descriptive statistics on NLE length and an exploratory discussion of the relative cost (\S\ref{app:nle-length}) and predictive utility of the two LLM-based NLE generators (\S\ref{app:llm-nle-cost}) in the appendix.

\section{Conclusion}
This work investigates design choices for selecting NLEs in explanation-augmented prompting and presents a comparative evaluation of natural language explanations used as few-shot rationales in ICL, comparing self-generated NLEs, NLEs generated by external LLMs, and, where available, human-written NLEs across six benchmarks and four instruction-tuned models. Our findings provide the following answers to the research questions. First, on the evaluated classification-style benchmarks, adding NLEs often improves accuracy relative to few-shot prompting without explanations. NLEs generated by an external LLM and, where available, human-written NLEs more consistently yield higher predictive utility than self-NLEs across selection settings. On \textsc{gsm8k}, however, the utility of NLEs varies more substantially with the model, NLE source, and selection strategy. Second, faithfulness-based selection of self-NLEs yields small average gains, but its effects vary across metrics, tasks, and models, sometimes improving and sometimes reducing downstream performance. Metric choice can lead to substantially different selections and, thus, different performance outcomes. Third, randomly swapped and out-of-distribution rationales generally reduce performance, although accuracy does not collapse. This indicates partial robustness to misaligned rationales while also demonstrating the importance of semantic alignment. Overall, our findings show that NLE source, NLE-selection strategy, and faithfulness metric are important design choices in explanation-augmented ICL and should be reported explicitly. We discuss the \textbf{limitations} of our study in Appendix~\ref{app:limitations}.

\subsubsection*{Acknowledgments}
We acknowledge the support for MD and BP through the ERC Consolidator Grant DIALECT 101043235. This research was also partially supported by the German Federal Ministry of Education and Research (BMBF) grant 01IS23069 Software Campus 3.0 (TU München).

\bibliography{main}
\bibliographystyle{tmlr}

\appendix

\section{Limitations} \label{app:limitations}
\paragraph{Empirical and Computational Scope.}
Our work is a comparative study of how NLE type and selection affect ICL utility, rather than a frontier-model benchmark or an attempt to optimize the strongest possible prompting pipeline. Due to the \emph{computational cost} of our pipeline and our limited \emph{budget}, we fix several implementation choices, including the number of evaluation models, runs per configuration, and few-shot exemplars. Specifically, we evaluate four instruction-tuned models from multiple families and scales across six benchmarks, using a fixed 6-shot setup to isolate explanation effects under a controlled prompt budget, with ZS, FS, and CoT as baselines. These choices make the experimental comparison tractable and consistent across conditions, but may limit statistical power and generalizability to other model families, larger prompt budgets, or different prompting strategies. For the same reason, our OOD rationale setting is limited to LLM-NLEs, since extending it to all NLE types would substantially expand the experimental grid across models, datasets, NLE types, selection strategies, and runs. \emph{Future work} should relax these constraints by evaluating additional model families, increasing the number of runs, varying the number of few-shot exemplars, exploring additional prompting and hyperparameter settings, and extending the OOD analysis to other NLE types.
For our CoT baseline, we evaluate zero-shot CoT rather than more computationally intensive variants such as Auto-CoT \citep{zhang2023-auto-CoT}, which require an additional pipeline for selecting examples and generating reasoning demonstrations. Extending the comparison to these variants across all models, datasets, and runs would substantially increase computational cost and is left for future work.

\paragraph{Human-explanations coverage.} Our comparison of human vs. LLM explanations is limited because only \textsc{ecqa} and e-\textsc{snli} provide human rationales, which is a reflection of the broader scarcity of reasoning datasets with such annotations. We therefore treat these results as dataset-specific and avoid universal claims. Broader validation across additional datasets, domains, and multiple annotators is needed and planned for future work. 

\paragraph{Faithfulness of NLEs.} Our faithfulness analysis relies on automated metrics from prior work. Given the scarcity of NLE faithfulness metrics in NLP~\citep{madsen-2024-self-explanations-faithful,lext-2025-facct,atanasova-2023-faitfulness-nle}, we evaluate two recent metrics: $fm_1$~\citep{lext-2025-facct} and $fm_2$~\citep{madsen-2024-self-explanations-faithful}. While these metrics evaluate whether self-NLEs are grounded, answer-supporting, and behaviorally consistent under intervention-based checks, they do not directly reveal the \emph{model's hidden causal reasoning process}. Thus, our results should be interpreted as evidence about metric-based self-NLE faithfulness rather than definitive internal faithfulness. Moreover, $fm_1$ is a practical heuristic for ranking self-NLEs, but it is not definitive, for example due to the equal weighting of its submetrics~\citep{lext-2025-facct}.  Similarly, $fm_2$ provides a useful counterfactual self-consistency check, but it is binary and depends on the model's ability to generate and interpret a valid counterfactual edit of the self-NLE. 
Our notion of faithfulness concerns self-explanations produced by the task model, not proxy or external LLM explanations; thus, common LLM-as-judge approaches for NLG quality are inappropriate here because they primarily assess surface-level explanation quality rather than whether the self-explanation reflects the task model's behavior. Future work should develop and validate metrics tailored to self-NLE faithfulness, and further study predictive utility as a standalone criterion and in relation to faithfulness. 

\paragraph{Evaluation-Subset Variation Across Selection Settings.}
Across selection settings, because each selection strategy may select a different set of few-shot
examples, the excluded instances, and consequently the remaining
evaluation subsets, may differ very slightly across settings. Therefore, very
small cross-setting performance differences may partly reflect
variation in the evaluated instances so we interpret these cautiously.

\section{Implementation Details} \label{app:implementation-details} 
Unless otherwise specified, model calls use the default API $temperature$. We set a random seed to make the random selection of few-shot examples and NLEs reproducible. For all setups, models, and datasets used in the evaluation, we fix the number of few-shot samples to $n = 6$. In Setup 1, we set the number of steps for Ph-CoT to 3, following \citet{bhan-2024-self-amplify}. We report the average over 5 runs for all setups. For the ablation studies in Setups 4–5, due to the high computational cost, we employ \(fm_1\) for evaluating NLEs faithfulness. 

\paragraph{Samples Selection Strategy.} In our experiments, we also evaluated the \textit{success setting}—using correctly classified samples \cite{bhan-2024-self-amplify}—but error-sampling performed better, as expected since models learn more from previously misclassified examples; we therefore adopt error-sampling in our experiments.

\section{Additional Experimental Details}

\subsection{Datasets} \label{app:dataset-details}
We conduct experiments on six datasets with varying sizes, as shown in Table~\ref{tab:dataset_sizes}.
We use each subset as a common pool for selecting in-context samples and evaluating the model. After selecting the $n$ few-shot examples, we exclude them from evaluation and evaluate on the remaining instances in the dataset. This maximizes data utilization while ensuring that no evaluated instance appears in the prompt.

\begin{table}[!h]
    \caption{Datasets and their sizes used in experiments}
    \centering
    \small
    \begin{tabular}{lc}
     \toprule
     Dataset & Size \\
     \midrule
     \textsc{ecqa}       & 310 \\
     e-\textsc{snli}     & 168 \\
     \textsc{snarks}     & 181 \\
     \textsc{boolean}    & 250 \\
     \textsc{Causal Judgment} & 190 \\
     \textsc{gsm8k} & 250 \\
     \bottomrule
    \end{tabular}
    \label{tab:dataset_sizes}
\end{table}

\FloatBarrier

\subsection{OOD NLEs Selection} \label{app:adversarial-selection}
In Setup 5 in NLE selection, the OOD NLEs are selected from other datasets as follows:
\begin{itemize}
\item For \textsc{ecqa}: LLM-NLEs from e-\textsc{snli} (generated by \texttt{4o-mini}).
\item For e-\textsc{snli}: LLM-NLEs from \textsc{ecqa} (\texttt{4o-mini}). 
\item For \textsc{snarks}: LLM-NLEs from \textsc{boolean} (\texttt{4o-mini}).
\item For \textsc{boolean}: LLM-NLEs from \textsc{snarks} (\texttt{o3-mini}).
\item For \textsc{cj}: LLM-NLEs from \textsc{snarks} (\texttt{o3-mini}).
\item For \textsc{gsm8k}: LLM-NLEs from \textsc{boolean} (\texttt{4o-mini}).
\end{itemize}

\section{Random and OOD Settings} \label{app:rq-3-random-ood}

In addition to Setups 1, 2, 3, and inspired by the following question: "How robust are LLMs to noisy or OOD explanations included in ICL prompts?", we introduce two additional settings in the form of ablation studies to test the robustness of LLMs to misleading or irrelevant explanations and assess their sensitivity to explanation-domain misalignment. \\
\textbf{Random Rationales (Setting~4):} We instantiate this ablation using the \(fm_1\)-based most-faithful condition from Setting~2. We hold the selected \((x,y)\) pairs fixed and replace only their corresponding rationales with randomly sampled rationales from the same dataset (e.g., pairing \((x_2,y_2)\) with rationale \(r_5\) originally associated with \((x_5,y_5)\)). Thus, Faithful-\(fm_1\) serves as the matched reference condition, allowing performance differences to be attributed to rationale mismatch rather than changes in the selected examples.\\
\textbf{Out-of-distribution Rationales (Setting 5):}
Extending Setup 4, we implement an out-of-distribution (OOD) (can also be described as cross-domain) scenario by selecting rationales from completely different datasets, e.g., for pair \((x_5,y_5)\) from \textsc{ecqa} it will be then \((x_5,r^*_8,y_5)\) where we include $r^*_8$, which belongs to \((x^*_8,y^*_8)\) from e-\textsc{snli}. We provide further details on the OOD explanations selection for each dataset in Appendix \ref{app:adversarial-selection}. In Setup 5, we implement it only for the LLM-NLEs setting, where the rationales considered are the LLM-generated ones.  
Setup 5 can be related to the work of \citet{noisy-cot-2024}, which investigates the impact of noisy rationales on CoT prompting. However, our approach differs in key ways: while their noise is synthetic and in-domain, we introduce OOD rationales to evaluate LLM robustness under extreme explanation-domain mismatch. In contrast, their focus is on assessing sensitivity to subtle variations in rationale quality within the same task.

\begin{table}[ht]
\caption{Standard deviations across five runs for the average performance of models per dataset under Random explanations (Setting 4) and OOD explanations (Setting 5).}
\centering
\scriptsize
\begin{tabular}{lccccc}
\toprule
\multirow{2}{*}{Dataset} 
  & \multicolumn{4}{c}{Faithful - Random} 
  & \multirow{2}{*}{OOD} \\
\cmidrule(lr){2-5}
  & Self-exp. & Human & \texttt{4o-mini}& \texttt{o3-mini} & \\
\midrule
\textsc{ecqa}          & 0.011 & 0.010 & 0.013 & 0.009 & 0.043  \\
e-\textsc{snli}        & 0.023 & 0.037 & 0.022 & 0.031 & 0.027  \\
\textsc{snarks}        & 0.019 & -- & 0.015 & 0.027 & 0.015  \\
\textsc{bool}         & 0.065 & -- & 0.044 & 0.008 & 0.008  \\
\textsc{cj}            & 0.022 & -- & 0.020 & 0.015 & 0.018  \\
\textsc{gsm8k}         & 0.011 & -- & 0.011 & 0.011 & 0.008 \\
\midrule
All datasets  & 0.025 & 0.023 & 0.021 & 0.017 & 0.020  \\
\bottomrule
\end{tabular}
\label{tab:random-adversarial-std}
\end{table}

\section{Prompt Templates} \label{app:prompts_templates}

\subsection{Pre-Prompts and Final Prompts Templates} \label{app:preprompts-final-prompts-examples}

\paragraph{\textbf{CoT Prompt}}
\begin{quote}\small\ttfamily
Choose the right answer by thinking step by step. Output your thought process and the single right answer choice (e.g., (a)) at the end.  
Strictly follow this format: <reasoning>, Answer: (a).
\end{quote}

\paragraph{\textbf{Ph-CoT Preprompt for Self-Generated Explanations}}
\begin{quote}\small\ttfamily
Choose the single right answer and generate a concise n-step explanation, with only one sentence per step.  
Strictly follow this format: The answer is (a), n-step explanation: step1, step2, ..., stepn.
\end{quote}

\paragraph{\textbf{Final ICL (n-samples) Prompt Using Ph-CoT Rationales}}
\begin{quote}\small\ttfamily
Choose the single right answer and generate a concise 3-step explanation, with only one sentence per step.  
Strictly follow this format: Answer: (a).  
3-step explanation: step1, step2, step3. \\
⟨question$_1$⟩
\end{quote}

\noindent\textbf{assistant:}
\begin{quote}\small\ttfamily
⟨rationale$_1$⟩ \\
⟨answer$_1$⟩ \\
\(\dots\) 
\end{quote}

\noindent\textbf{user:}
\begin{quote}\small\ttfamily
⟨question$_n$⟩
\end{quote}

\noindent\textbf{assistant:}
\begin{quote}\small\ttfamily
⟨rationale$_n$⟩ \\
⟨answer$_n$⟩
\end{quote}

\noindent\textbf{user:}
\begin{quote}\small\ttfamily
⟨question$_{n+1}$⟩
\end{quote}

\subsection{fm1 - Faithfulness Evaluation Prompts} \label{app:faithfulness-evaluation-prompts}
This section provides the prompts employed to evaluate explanation faithfulness across several metrics.

\subsubsection{1. Question Answer Generation}

\paragraph{Question Generation Prompt}
{\small
\begin{quote}
\texttt{Generate at least 5 questions about commonsense reasoning that can be answered using the following explanation. Focus on questions that test understanding of the concepts, relationships, and reasoning presented in the explanation. Split all the questions with a newline character. Don't add anything else to your response. \\
Explanation: ⟨explanation⟩}
\end{quote}}

\paragraph{Question Evaluation Prompt}
{\small
\begin{quote}
\texttt{Can the following question be answered using the information provided in this explanation?\\ Explanation: ⟨explanation⟩ \\
Question: ⟨question⟩ \\
Answer with 'yes' if the explanation provides sufficient information to answer the question, or 'no' if the explanation lacks the necessary information. Just give me yes or no. Don't add anything else to your answer.}
\end{quote}}

\subsubsection{2. Counterfactual Stability}

\paragraph{Explanation Rewriting Prompt}
{\small
\begin{quote}
\texttt{This was the question: ⟨question⟩\\
For this question, a language model originally chose option ⟨original\_choice⟩ and gave this \\ Explanation: ⟨explanation⟩\\
Now, rewrite and modify the explanation to support option ⟨alternative\_choice⟩ instead.\\
Make the explanation logically consistent with the new choice while maintaining the same style and structure.\\
Just give me the new explanation, don't add anything else to your answer.}
\end{quote}}

\paragraph{Rephrased Explanation Answer Test Prompt}
{\small
\begin{quote}
\texttt{Given this explanation: ⟨rephrased\_explanation⟩\\
Answer the following question: ⟨question⟩\\
Important: Choose the SINGLE letter (a, b, c, d, or e) that best answers the question based on the explanation.\\
Respond with just the letter (e.g., a or b). Don't add anything else to your answer.}
\end{quote}}

\paragraph{Choice Letter Extraction Prompt}
{\small
\begin{quote}
\texttt{Extract the choice letter from this response: ⟨new\_choice⟩\\
The response should contain a single letter (a, b, c, d, or e)\\
Just give me the letter. Don't add anything else to your answer.}
\end{quote}}

\subsubsection{3. Contextual Faithfulness}

\paragraph{Important Words Extraction}
{\small
\begin{quote}
\texttt{Question: ⟨ground\_question⟩\\
Explanation: ⟨explanation⟩\\
For the above question and explanation, you predicted choice {predicted\_choice}.\\
Give me the 5 most important words from the explanation that led to this answer choice.\\
These should be words that without them, you would not be able to make the same prediction.\\
Give me only these words separated by commas, don't add anything else to your answer.}
\end{quote}}

\paragraph{Redacted Explanation Prediction Prompt}
{\small
\begin{quote}
\texttt{Question: ⟨ground\_question⟩\\
Explanation: ⟨redacted\_explanation⟩\\
Based on this explanation, which choice (a, b, c, d, or e) would you select?\\
If the explanation doesn't provide enough information to make a confident choice, respond with 'insufficient'.\\
Give me only the letter or 'insufficient'. Don't add anything else to your answer.}
\end{quote}}

\paragraph{Response Classification Prompt}
{\small
\begin{quote}
\texttt{Question: ⟨ground\_question⟩\\
I asked a model to choose between options (a, b, c, d, e) or say 'insufficient' for this question and it responded: ⟨redacted\_pred⟩ \\
Classify this response as either 'choice' (if it selected a specific option) or 'insufficient' (if it indicated lack of information). \\
Just give me the classification. Don't add anything else to your answer.}
\end{quote}}

\subsection{fm2 - Faithfulness Evaluation Prompts} \label{app:faithfulness-evaluation-prompts-fm2}
This section provides the prompts employed to evaluate counterfactual explanation faithfulness.

\subsubsection{1. Prediction Prompt}
{\small
\begin{quote}
\texttt{Consider the following paragraph, and answer the question: ⟨ground\_question⟩ The paragraph can contain redacted words marked with [REDACTED]. Answer either ⟨answer\_choices⟩ or "unknown" if the question can not be answered. Provide a single, concise answer without repetition or multiple attempts. Do not explain the answer. \\
Paragraph:  ⟨explanation⟩}
\end{quote}}

\subsubsection{2. Counterfactual Explanation}
{\small
\begin{quote}
\texttt{Edit the following paragraph such that the answer to the question ⟨ground\_question⟩ is ⟨alternative\_choice⟩. Make as few edits as possible. Your response should contain only the edited paragraph. Start with "Step1:" and maintain the step-by-step structure. Do not explain the answer or include any commentary. \\
Paragraph: ⟨explanation⟩}
\end{quote}}

\subsubsection{3. Prediction Prompt}
{\small
\begin{quote}
\texttt{Consider the following paragraph, and answer the question: ⟨ground\_question⟩ The paragraph can contain redacted words marked with [REDACTED]. Answer either ⟨answer\_choices⟩ or "unknown" if the question can not be answered. Provide a single, concise answer without repetition or multiple attempts. Do not explain the answer. \\
Paragraph:  ⟨counterfactual\_explanation⟩}
\end{quote}}

\section{Faithfulness Metrics} \label{app:metric-details}

\subsection{$fm_1$ LExT Faithfulness Metric} \label{app:lext-metric-details}

Faithfulness evaluates the extent to which a model’s generated explanations remain reliable and firmly rooted in the underlying context. 
For operations in sub-metrics Question Answer Generation (generating questions) and counterfactual (rewriting the NLE to support alternative labels), we use the same model that generates the self-NLEs also for these operations, instead of using another larger LLM (as done in \citep{lext-2025-facct}). This substantially reduces evaluation costs, as QAG and Counterfactual Stability require multiple additional model calls per explanation. It also keeps the generated questions and counterfactual rewrites aligned with the linguistic style and task representation of the model whose NLE is being evaluated. All evaluation operations are performed in separate sessions from the call that generated the original NLE.

\subsubsection{Question Answer Generation (QAG) Sub-Metric}

This sub-metric assesses whether an explanation contains sufficient information to answer five questions derived from the explanation itself. Several questions are generated (by the same model but in a separate session) based on the predicted explanation. Each question is then presented to the target model together with the explanation, and the target model returns a binary label: 1 if the question can be answered using the explanation and 0 otherwise. The QAG score is calculated as $\mathrm{QAG} = \frac{\text{Number of answerable questions}}{\text{Total number of generated questions}}$. The resulting score lies in $[0,1]$, with a higher score indicating that the explanation more consistently contains the information needed to answer questions derived from it.

\subsubsection{Counterfactual Stability Sub-Metric}

This sub-metric evaluates whether the model adapts its prediction when presented with an explanation that supports an alternative label.
The same model, in a separate session, rephrases the original explanation so that it implies an alternative prediction, for example, by changing an explanation that supports ``Yes'' into one that supports ``No.'' The counterfactual explanation and the original question are then presented to the target model. The response is assigned a raw score as follows:
\begin{itemize}
    \item $1$: The model changes its original label.
    \item $0$: The model retains its original label or does not provide a meaningful or valid response.
\end{itemize}

This results in a binary Counterfactual Stability score in $\{0,1\}$. A score of $1$ indicates that the prediction changes under the counterfactual explanation, whereas a score of $0$ indicates either that the prediction remains unchanged or that no valid response is produced.

\subsubsection{Contextual Faithfulness Sub-Metric}

This sub-metric assesses the extent to which the model's prediction depends on specific elements of the input context. The model is first asked to identify five keywords from the context (including the input and explanation) that were important for producing its prediction. These keywords are then redacted in two phases:

\begin{itemize}
    \item \textbf{Complete Redaction:} All identified keywords are removed simultaneously, and the model is prompted again. A faithful response should recognize that the remaining context is insufficient to make a reliable prediction.
    
    \item \textbf{Sequential Redaction:} The identified keywords are removed one at a time, with the model re-prompted after each removal, to assess the effect of each keyword's absence on the prediction.
\end{itemize}

If the model fails the complete-redaction phase, meaning that it does not recognize the lack of sufficient information, the Contextual Faithfulness score is set to 0. If the model passes the complete-redaction phase, its score is calculated in the sequential phase as $\mathrm{Contextual\ Faithfulness} = \frac{\text{Number of ``insufficient'' responses}}{\text{Total number of sequential prompts}}$. The resulting score lies in $[0,1]$, with a higher score indicating greater dependence on the contextual elements identified as important.
The \textbf{final faithfulness score} is calculated as the average of the three sub-metric scores, resulting in a value in $[0,1]$. To obtain a binary classification (faithful/unfaithful), similar to $fm_2$, we apply a relative threshold rather than a fixed global cutoff. For each dataset separately, we compute the 75th percentile of $fm_1$ scores over all NLEs (per model) in that dataset. An NLE with a score below this threshold is classified as unfaithful, whereas an NLE with a score greater than or equal to the threshold is classified as faithful. We choose this threshold because $fm_1$ is a continuous composite metric without a natural absolute decision boundary, and its score distributions vary across datasets. A dataset-specific threshold therefore allows ``faithful'' to denote comparatively high-scoring NLEs within each dataset. We use the upper quartile to obtain a selective set of high-scoring explanations rather than treating median-scoring explanations as faithful. The resulting labels thus represent relative within-dataset faithfulness rather than an absolute cross-dataset criterion.

\subsection{$fm_2$ Faithfulness Metric} \label{app:fm2-details}

\subsubsection{Counterfactual Explanation Metric}
This metric tests whether an explanation (self-NLE) is behaviorally consistent with the model's prediction by leveraging self-consistency checks.
We use this method from \citet{madsen-2024-self-explanations-faithful} and adapt it to test the faithfulness of self-NLEs. The procedure involves three steps:
\begin{itemize}
    \item Initial Classification: The model predicts the label for a given task question while using the self-NLE as contextual reasoning. The question and answer choices are provided in the instruction, while the self-NLE serves as the supporting paragraph.
    \item Counterfactual Generation: The model is prompted to edit the original self-NLE such that it would support an alternative label. The prompt explicitly instructs the model to make minimal edits to the self-NLE that would justify the alternative answer, without providing additional commentary. Importantly, the task input and question remain fixed, only the self-NLE is modified.
    \item Self-Consistency Check: The model re-classifies using the edited (counterfactual) self-NLE in a new inference session. If the prediction flips to the intended alternative label, the original self-NLE is considered faithful; otherwise, it is unfaithful.
\end{itemize}
The faithfulness score for each example is binary (true if the prediction flips as intended, false otherwise), reflecting whether the self-NLE aligns with the model's behavior.

\subsection{Comparison of Faithfulness Metrics} \label{app:faithfulness-metrics-comparison}
We use two different faithfulness metrics that reflect complementary approaches to evaluating model explanations. The \(fm1\) faithfulness metric assesses how well explanations reflect the model’s reasoning by combining several components—such as whether explanations answer relevant questions, their dependence on key contextual information, and their stability under counterfactual changes—into a continuous score. This metric requires generating multiple explanation-related outputs and integrates diverse signals to provide a nuanced measurement of faithfulness.

On the other hand, \(fm2\) faithfulness metric focuses specifically on causal consistency by generating minimal input edits that flip the model’s prediction. It then verifies, via a self-consistency check, whether the model’s output changes accordingly in a separate inference call. This results in a binary faithful/unfaithful label per example, offering a straightforward causal test of explanation alignment with model behavior.

The two metrics differ both in the level of detail of their outputs (continuous versus binary) and in their methodological approaches—multi-component comprehensive assessment versus a focused intervention-based test—addressing different aspects of explanation faithfulness and requiring different degrees of interaction with the model.

\FloatBarrier

\section{LLM-NLEs Generation}
We generated NLEs using two LLMs: GPT-\texttt{4o-mini} and \texttt{o3-mini}. The prompt template used to generate explanations for the datasets is shown below. For GPT-\texttt{4o-mini}, we set the temperature to 0.7 during explanation generation; We chose this balanced value so we have some aspect of controlled randomness, i.e., to enable the explainer to explore slightly varied phrasings or reasoning paths. For \texttt{o3-mini}, which is a reasoning-optimized model, temperature control is not available. We prompted the LLMs as follows to generate NLEs for each sample in the respective dataset:\\

For \textsc{snarks}:
\begin{tcolorbox}[colback=cyan!10,colframe=black!50, sharp corners]
\begin{quote}
Consider the following question and correct answer from a sarcasm detection task. Please only write a concise explanation for the provided correct answer.\\
Question: \{\textit{question}\} Answer: \{\textit{answer}\}
\end{quote}
\end{tcolorbox}

For \textsc{boolean}: 
\begin{tcolorbox}[colback=green!10,colframe=black!50, sharp corners]
\begin{quote}
Consider the following \textsc{boolean} expression and correct answer. Please only write a concise natural language explanation for the correct answer:\\
Expression: \{\textit{expression}\} Answer: \{\textit{answer}\}
\end{quote}
\end{tcolorbox}

For \textsc{ecqa}: 
\begin{tcolorbox}[colback=red!10,colframe=black!50, sharp corners]
\begin{quote}
Consider the following multiple choice question, choices and the correct answer. Only write a concise explanation for the provided correct answer:\\
Question: \{\textit{question}\}\\Choices: \{\textit{option1}\}, \{\textit{option2}\}, \{\textit{option3}\}, \{\textit{option4}\}, \{\textit{option5}\}, Answer: \{\textit{answer}\}
\end{quote}
\end{tcolorbox}

For e-\textsc{snli}:
\begin{tcolorbox}[
    colback=violet!10!white,
    colframe=violet!55!black,
    sharp corners
]
\begin{quote}
Given the following:\\
Premise: \{\textit{premise}\}\\
Hypothesis: \{\textit{hypothesis}\}\\
Label: \{\textit{label\_num}\} (where entailment = 0, neutral = 1, contradiction = 2)\\
Provide exactly one sentence that directly connects the premise to the
hypothesis. Do not include any prefixes such as ``Explanation:'' or
``Here is the explanation.'' Start directly with the explanation sentence.
The explanation should not explicitly hint at the label or contain the label
itself in any form. Focus solely on reasoning that connects the premise to
the hypothesis without revealing the classification.
\end{quote}
\end{tcolorbox}

For \textsc{causal judgment}:
\begin{tcolorbox}[colback=yellow!10,colframe=black!50, sharp corners]
\begin{quote}
Consider the following questions about causation with the correct answers. Please only write a concise explanation for the provided correct answer:\\
Question: \{\textit{question}\} Answer: \{\textit{answer}\}
\end{quote}
\end{tcolorbox}

For \textsc{gsm8k}: 
\begin{tcolorbox}[colback=gray!10,colframe=black!50, sharp corners]
\begin{quote}
Consider the following math problem and its correct answer. Please only write a concise natural language explanation for the correct answer:\\
Problem: \{\textit{problem}\} Answer: \{\textit{answer}\}
\end{quote}
\end{tcolorbox}

\FloatBarrier

\clearpage
\section{Effect of Gold-Label Conditioning on Ph-CoT Self-Explanations}
\label{app:self-exp-step}

\begin{table*}[!h]
\caption{Comparison of two self-explanation variants in the Ph-CoT setup. Step A generates post-hoc self-explanations after prediction, while Step B regenerates explanations for Step A errors conditioned on the gold label. The reported values show the performance delta (Step B $-$ Step A) across datasets and models, with positive values indicating improvement.}
\centering
\scriptsize
\setlength\tabcolsep{5pt}
\begin{tabular}{l|l|ccccc}
\toprule
\multirow{3}{*}{Dataset}
  & \multirow{3}{*}{Models}
  & \multicolumn{5}{c}{Step B - Step A} \\
\cmidrule(lr){3-7}
  &
  & Random 
  & Faithful-fm1
  & Faithful-fm2
  & Lowest-Faithful-fm1
  & Lowest-Faithful-fm2 \\
\cmidrule(lr){3-7}
  &
  & Self-exp. & Self-exp. & Self-exp. & Self-exp. & Self-exp. \\
\midrule

\multirow{5}{*}{\textsc{ecqa}} 
 & \texttt{4o-mini}      & +0.011 & +0.012 & +0.001 & 0.000 & 0.000 \\
 & \texttt{llama-70B}     & -0.098 & -0.059 & 0.000 & -0.156 & -0.002 \\
 & \texttt{llama-8B}      & +0.003 & -0.038 & -0.014 & -0.542 & 0.000 \\
 & \texttt{mistral-7B}    & -0.054 & -0.047 & -0.032 & -0.037 & -0.004 \\
 & Avg           & -0.035 & -0.033 & -0.011 & -0.183 & -0.002 \\
\midrule

\multirow{5}{*}{e-\textsc{snli}} 
 & \texttt{4o-mini}      & +0.053 & +0.012 & +0.003 & +0.105 & +0.051 \\
 & \texttt{llama-70B}     & +0.100 & -0.006 & +0.039 & +0.117 & -0.008 \\
 & \texttt{llama-8B}      & +0.079 & -0.076 & +0.053 & +0.103 & +0.089 \\
 & \texttt{mistral-7B}    & +0.039 & +0.043 & +0.025 & +0.133 & +0.124 \\
 & Avg           & +0.067 & -0.007 & +0.030 & +0.115 & +0.064 \\
\midrule

\multirow{5}{*}{\textsc{snarks}} 
 & \texttt{4o-mini}      & -0.109 & -0.132 & +0.057 & +0.134 & +0.015 \\
 & \texttt{llama-70B}     & +0.008 & +0.040 & +0.024 & +0.157 & +0.001 \\
 & \texttt{llama-8B}      & -0.014 & -0.109 & +0.086 & -0.020 & +0.181 \\
 & \texttt{mistral-7B}    & +0.109 & +0.051 & +0.074 & -0.005 & +0.177 \\
 & Avg           & -0.001 & -0.038 & +0.061 & +0.067 & +0.094 \\
\midrule

\multirow{5}{*}{\textsc{boolean}} 
 & \texttt{4o-mini}      & +0.054 & +0.007 & -0.002 & +0.005 & +0.006 \\
 & \texttt{llama-70B}     & +0.122 & -0.050 & +0.274 & +0.431 & -0.007 \\
 & \texttt{llama-8B}      & +0.260 & +0.448 & +0.081 & +0.381 & +0.203 \\
 & \texttt{mistral-7B}    & +0.271 & +0.049 & +0.214 & +0.333 & +0.362 \\
 & Avg           & +0.177 & +0.114 & +0.142 & +0.288 & +0.141 \\
\midrule

\multirow{5}{*}{\textsc{cj}} 
 & \texttt{4o-mini}      & -0.052 & -0.019 & +0.083 & +0.070 & +0.016 \\
 & \texttt{llama-70B}     & -0.007 & -0.007 & +0.027 & +0.014 & +0.112 \\
 & \texttt{llama-8B}      & +0.080 & -0.008 & +0.049 & +0.037 & -0.035 \\
 & \texttt{mistral-7B}    & +0.095 & -0.021 & +0.090 & +0.170 & +0.072 \\
 & Avg           & +0.029 & -0.013 & +0.062 & +0.072 & +0.041 \\
\midrule

All datasets & Avg & +0.047 & +0.004 & +0.056 & +0.072 & +0.068 \\
\bottomrule
\end{tabular}
\label{tab:selfexp-stepB-minus-stepA}
\end{table*}

\FloatBarrier

\section{NLE Length Descriptive Statistics} \label{app:nle-length}
Table~\ref{tab:nle-length-compressed}
presents the average length of different NLE types across datasets and models. Although longer NLEs may be expected to provide more reasoning signal, our results do not show a clear length--utility relationship. Self-NLEs are often longer, partly due to their step-by-step generation format, but they generally do not outperform LLM- or human-generated NLEs. Conversely, shorter NLEs, such as \texttt{o3-mini} explanations and human-NLEs on e-\textsc{snli}, are often (but not always) competitive or strongest. This suggests that NLE utility is driven less by length itself, but further controlled-level experiments are needed to better draw conclusions. 

\begin{table*}[!h]
\caption{Average length of NLEs across different datasets, models, and NLE types. Values report mean word counts with standard deviations shown as subscripts.}
\centering
\scriptsize
\setlength{\tabcolsep}{5pt}
\begin{tabular}{l cccc c cc}
\toprule
& \multicolumn{4}{c}{Self-NLEs} & \multicolumn{1}{c}{Human-NLE} & \multicolumn{2}{c}{LLM-NLEs} \\
\cmidrule(lr){2-5} \cmidrule(lr){6-6} \cmidrule(lr){7-8}
Dataset & \texttt{4o-mini}& llama-70b & llama-8b & mistral-7b & Human & \texttt{4o-mini}& \texttt{o3-mini} \\
\midrule
\textsc{ecqa}    & $42.18_{7.09}$ & $70.28_{8.81}$  & $54.37_{9.48}$  & $35.17_{10.56}$ & $44.14_{13.53}$ & $39.06_{9.19}$  & $29.95_{7.45}$ \\
e-\textsc{snli}  & $50.76_{9.56}$ & $76.53_{11.51}$ & $62.33_{8.76}$  & $48.83_{10.62}$ & $10.98_{4.73}$  & $23.17_{4.56}$  & $26.55_{5.38}$ \\
\textsc{snarks}  & $50.78_{9.74}$ & $81.67_{10.33}$ & $61.10_{9.81}$  & $49.64_{13.31}$ & ---              & $42.03_{8.74}$  & $38.91_{7.90}$ \\
\textsc{boolean} & $43.53_{8.50}$ & $77.87_{11.22}$ & $67.16_{11.41}$ & $50.95_{19.72}$ & ---              & $44.14_{10.98}$ & $34.02_{8.05}$ \\
\textsc{cj}      & $53.59_{9.12}$ & $96.85_{14.85}$ & $73.97_{13.90}$ & $48.94_{14.70}$ & ---              & $60.74_{13.65}$ & $43.96_{9.76}$ \\
\textsc{gsm8k}   & $62.75_{15.10}$ & $116.07_{43.25}$ & $89.92_{31.37}$ & $83.89_{29.94}$ & ---              & $72.36_{22.64}$ & $48.39_{13.51}$ \\
\bottomrule
\end{tabular}
\label{tab:nle-length-compressed}
\end{table*}

\FloatBarrier

\clearpage

\section{Lower-Cost LLMs Can Provide Competitive NLE Utility} \label{app:llm-nle-cost}
Focusing on externally generated LLM-NLEs, the random-selection results in Table~\ref{tab:rq1-baseline-random-comp-five-runs} show that NLEs generated by GPT-\texttt{4o-mini} provide predictive utility comparable to those generated by \texttt{o3-mini}. At the dataset level, each explainer yields the higher mean accuracy on three of the six benchmarks, indicating broadly comparable performance overall. Thus, in this comparison, the higher-cost explainer (\texttt{o3-mini}) does not consistently yield NLEs with greater predictive utility. These findings suggest that lower-cost explanation generators may offer a favorable cost--utility trade-off for explanation-augmented ICL. However, broader conclusions require systematic evaluation across a wider range of lower- and higher-cost models.

\section{Model Size Impact on Explanation Sensitivity} \label{app:size-sensitivity}

\begin{figure*}[htbp]
\centering
     \centering
     \includegraphics[width=0.65\linewidth]{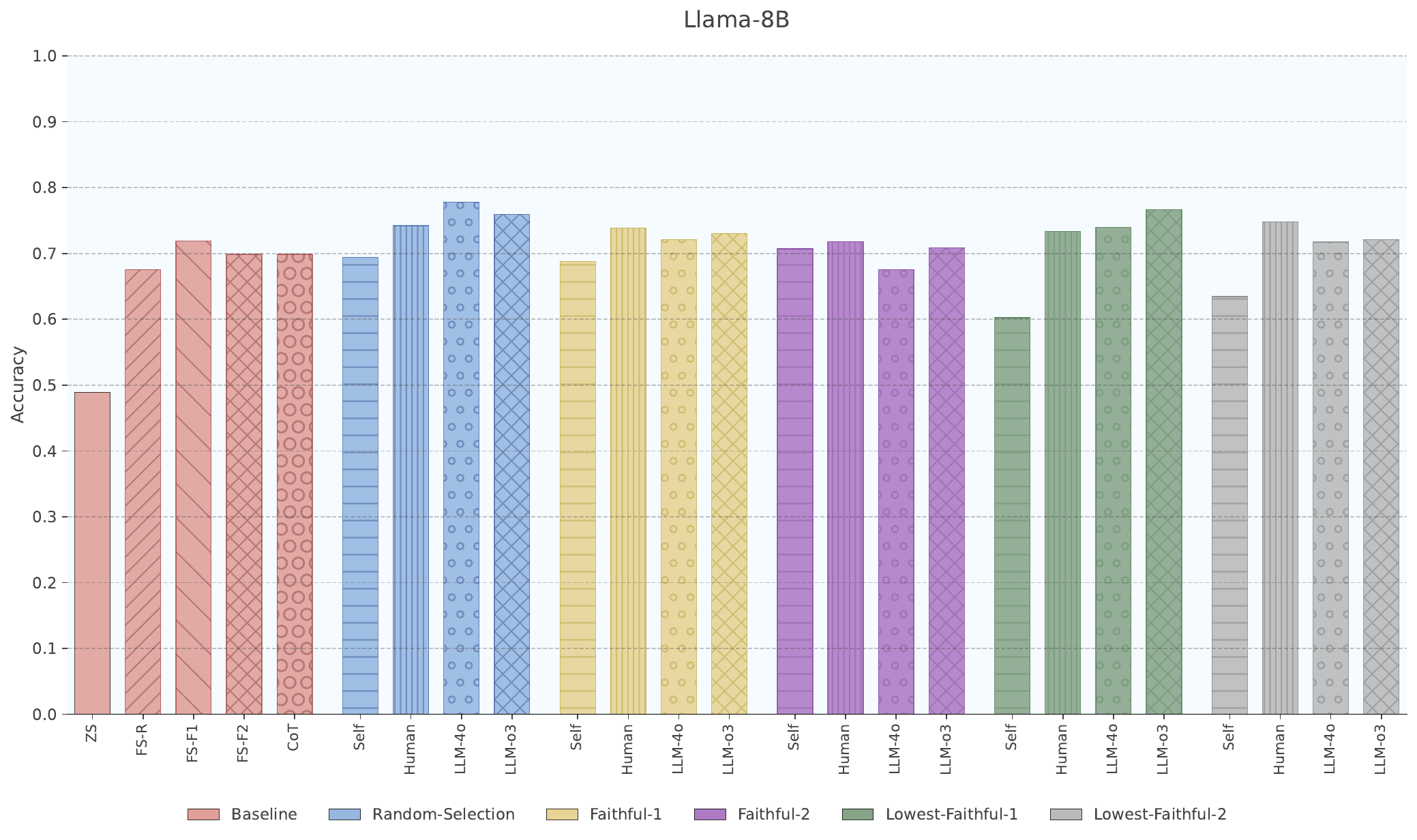}

     \centering
     \includegraphics[width=0.65\linewidth]{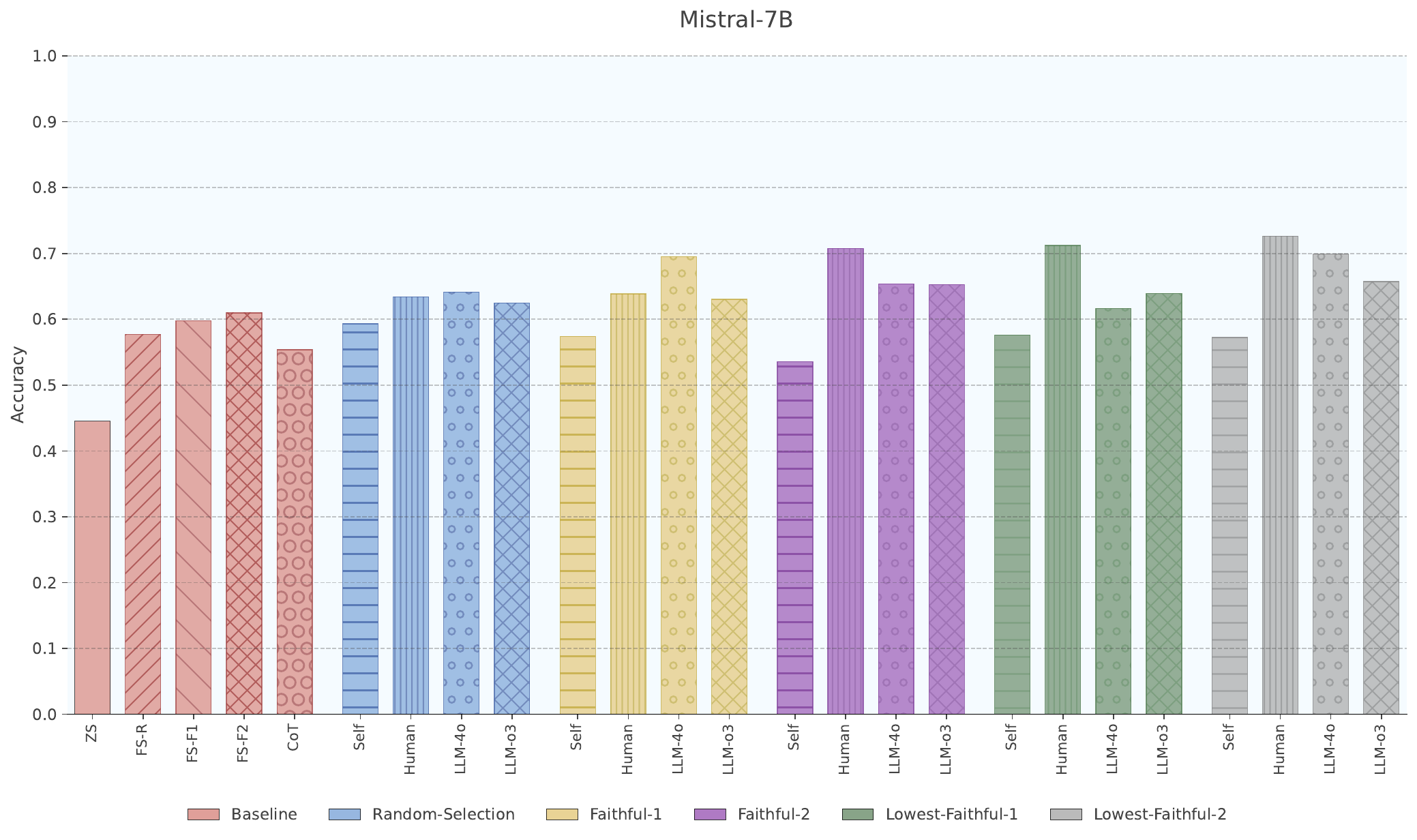}

     \centering
     \includegraphics[width=0.65\linewidth]{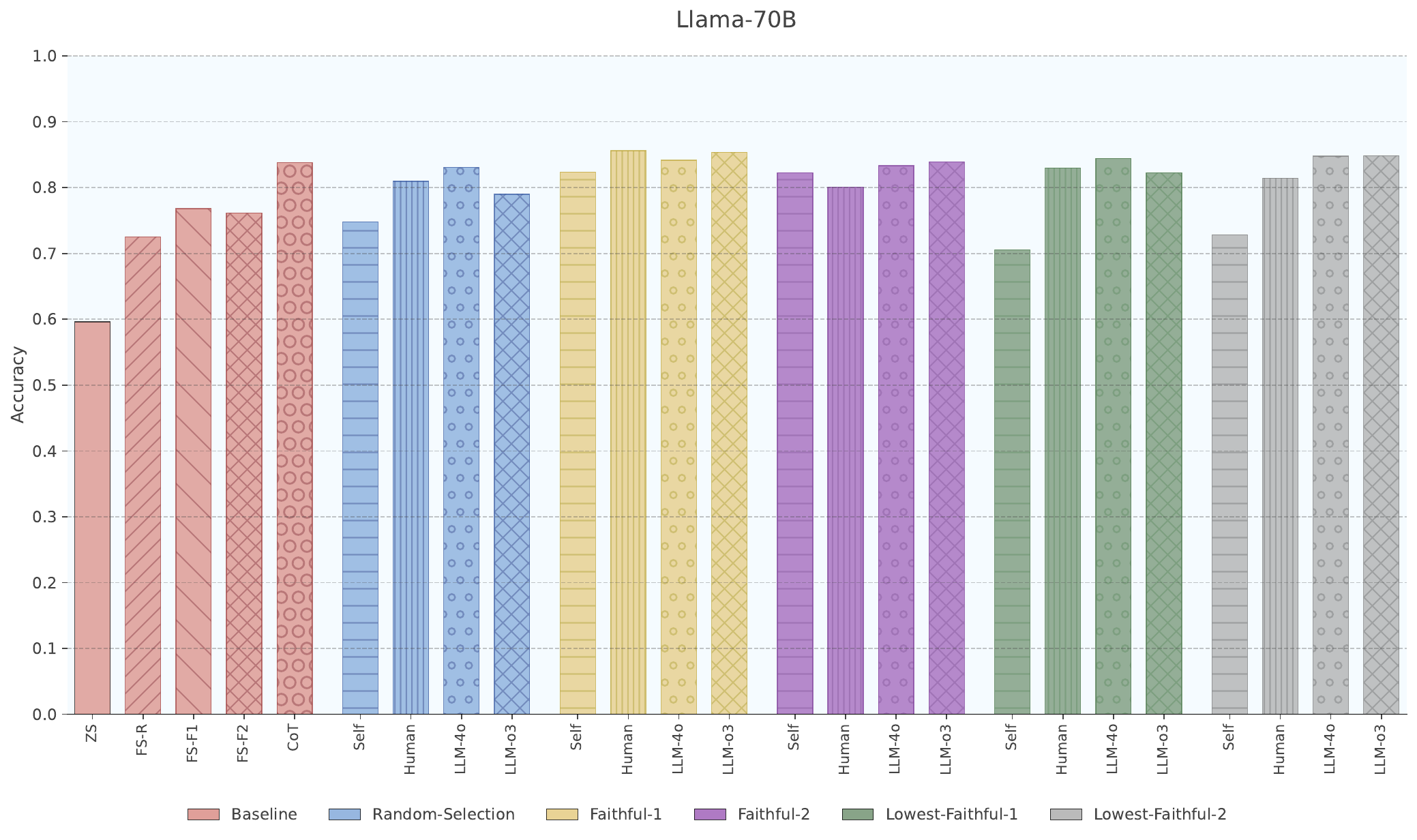}
    
    \caption{Performance comparison of explanation selection strategies using small and large LMs from the same model family (Llama- 8B and 70B) and from the different family in Mistral 7B. Results are averaged across all datasets to highlight the impact of model scale on strategy effectiveness.}
    \label{fig:llama_comparison}
\end{figure*}

\FloatBarrier
\section{Disagreement Between NLE Faithfulness Metrics} \label{app:heatmaps-disagreement-ecqa}

\begin{table*}[!h]
\caption{Disagreements between Faithfulness Metrics \(fm1\) and \(fm2\). The table details the total count, rate, and direction of disagreements across all datasets and models.}
\centering
\scriptsize
\setlength\tabcolsep{3.5pt}
\begin{tabular}{l|l|c|c|c|c}
\toprule
\multirow{2}{*}{Dataset} & \multirow{2}{*}{Model} & \multirow{2}{*}{Disagreement} & \multirow{2}{*}{Rate} & \multicolumn{2}{c}{Dis. Dir.} \\
\cmidrule(lr){5-6}
& & & & fm1~\cmark{} fm2~\xmark{} & fm1~\xmark{} fm2~\cmark{} \\
\midrule
\multirow{4}{*}{\textsc{ecqa}} & \texttt{4o-mini}& 158 & 0.510 & 95 & 63 \\
 & \texttt{llama-70B} & 168 & 0.542 & 131 & 37 \\
 & \texttt{llama-8B} & 162 & 0.523 & 117 & 45 \\
 & \texttt{mistral-7B} & 111 & 0.358 & 74 & 37 \\
 & Avg & 150 & 0.483 & 104 & 46 \\
\midrule
\multirow{4}{*}{e-\textsc{snli}} & \texttt{4o-mini}& 81 & 0.482 & 42 & 39 \\
 & \texttt{llama-70B} & 77 & 0.459 & 53 & 24 \\
 & \texttt{llama-8B} & 87 & 0.518 & 60 & 27 \\
 & \texttt{mistral-7B} & 63 & 0.375 & 36 & 27 \\
 & Avg & 77 & 0.458 & 48 & 29 \\
\midrule
\multirow{4}{*}{\textsc{snarks}} & \texttt{4o-mini}& 91 & 0.503 & 28 & 63 \\
 & \texttt{llama-70B} & 89 & 0.492 & 42 & 47 \\
 & \texttt{llama-8B} & 99 & 0.547 & 16 & 83 \\
 & \texttt{mistral-7B} & 88 & 0.486 & 83 & 5 \\
 & Avg & 92 & 0.507 & 42 & 50 \\
\midrule
\multirow{4}{*}{\textsc{boolean}} & \texttt{4o-mini}& 118 & 0.472 & 28 & 90 \\
 & \texttt{llama-70B} & 149 & 0.596 & 130 & 19 \\
 & \texttt{llama-8B} & 137 & 0.548 & 46 & 91 \\
 & \texttt{mistral-7B} & 124 & 0.496 & 55 & 69 \\
 & Avg & 132 & 0.528 & 65 & 67 \\
\midrule
\multirow{4}{*}{\textsc{cj}} & \texttt{4o-mini}& 112 & 0.589 & 23 & 89 \\
 & \texttt{llama-70B} & 93 & 0.490 & 31 & 62 \\
 & \texttt{llama-8B} & 117 & 0.616 & 27 & 90 \\
 & \texttt{mistral-7B} & 84 & 0.442 & 46 & 38 \\
 & Avg & 102 & 0.534 & 32 & 70 \\
\midrule
\multirow{4}{*}{\textsc{gsm8k}} & \texttt{4o-mini}& 127 & 0.508 & 107 & 20 \\
 & \texttt{llama-70B} & 114 & 0.456 & 109 & 5 \\
 & \texttt{llama-8B} & 108 & 0.432 & 103 & 5 \\
 & \texttt{mistral-7B} & 168 & 0.672 & 158 & 10 \\
 & Avg & 129 & 0.517 & 119 & 10 \\
\midrule
All datasets & Avg & 114 & 0.504 & 68 & 45 \\
\bottomrule
\end{tabular}
\label{tab:disagreement_analysis_faithfulness}
\end{table*}

\begin{figure*}[htbp]
    \centering
    \includegraphics[width=0.90\textwidth]{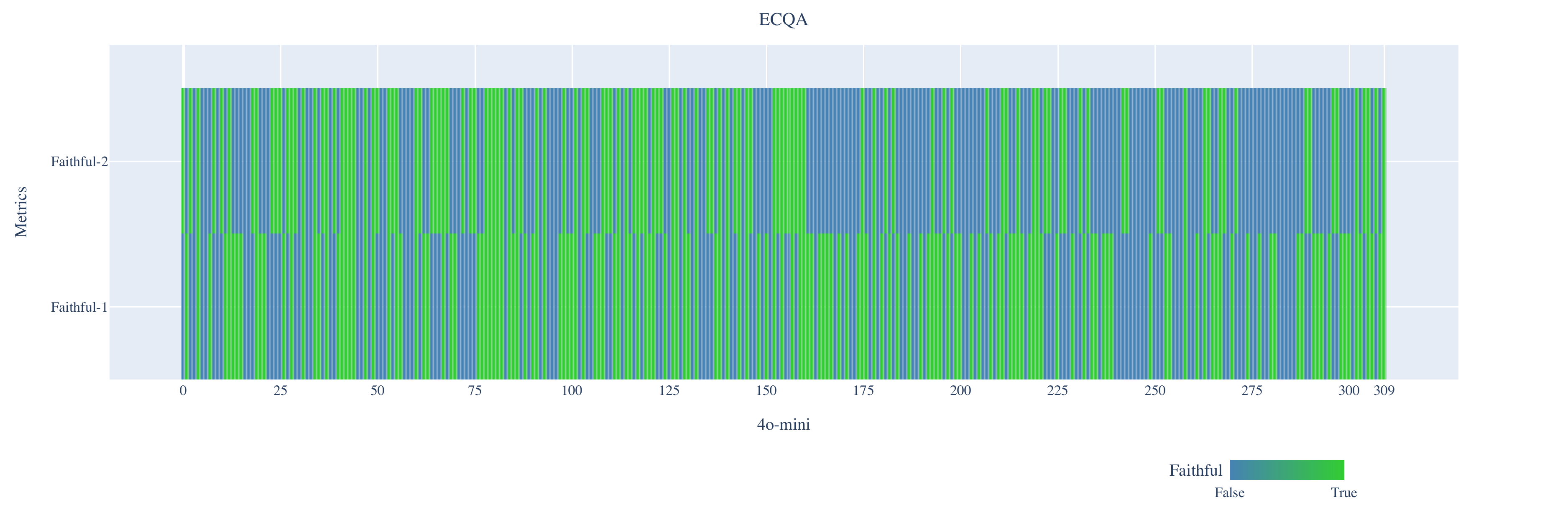}
    \caption{Binary heatmap visualization of \textsc{ecqa} test samples comparing two faithfulness metrics using the \texttt{4o-mini}model. Each column represents a test sample, while rows correspond to the two faithfulness metrics. Cells are color-coded to indicate binary outcomes: green = True, blue = False.}
    \label{fig:binary-heatmap-ecqa-4o-mini}
\end{figure*}

\begin{figure*}[htbp]
    \centering
    \includegraphics[width=0.90\textwidth]{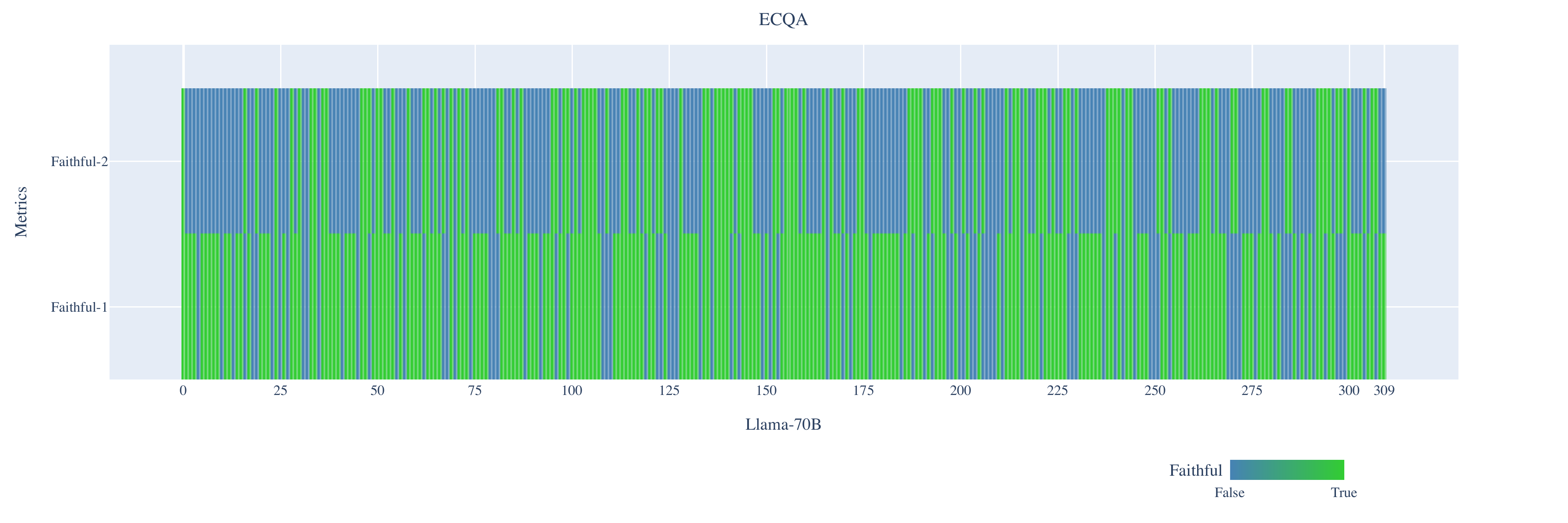}
    \caption{Binary heatmap visualization of \textsc{ecqa} test samples comparing two faithfulness metrics using the \texttt{Llama-70B} model. Each column represents a test sample, while rows correspond to the two faithfulness metrics. Cells are color-coded to indicate binary outcomes: green = True, blue = False.}
    \label{fig:binary-heatmap-ecqa-llama-70b}
\end{figure*}

\begin{figure*}[htbp]
    \centering
    \includegraphics[width=0.90\textwidth]{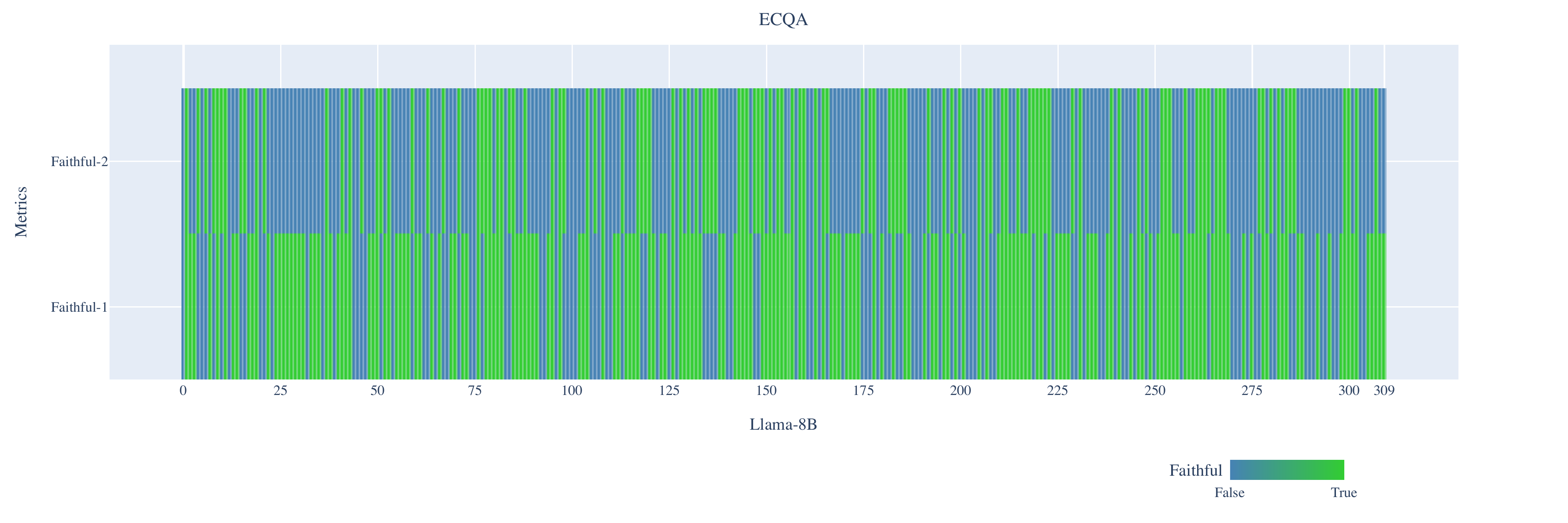}
    \caption{Binary heatmap visualization of \textsc{ecqa} test samples comparing two faithfulness metrics using the \texttt{Llama-8B} model. Each column represents a test sample, while rows correspond to the two faithfulness metrics. Cells are color-coded to indicate binary outcomes: green = True, blue = False.}
    \label{fig:binary-heatmap-ecqa-llama-8b}
\end{figure*}

\begin{figure*}[htbp]
    \centering
    \includegraphics[width=0.90\textwidth]{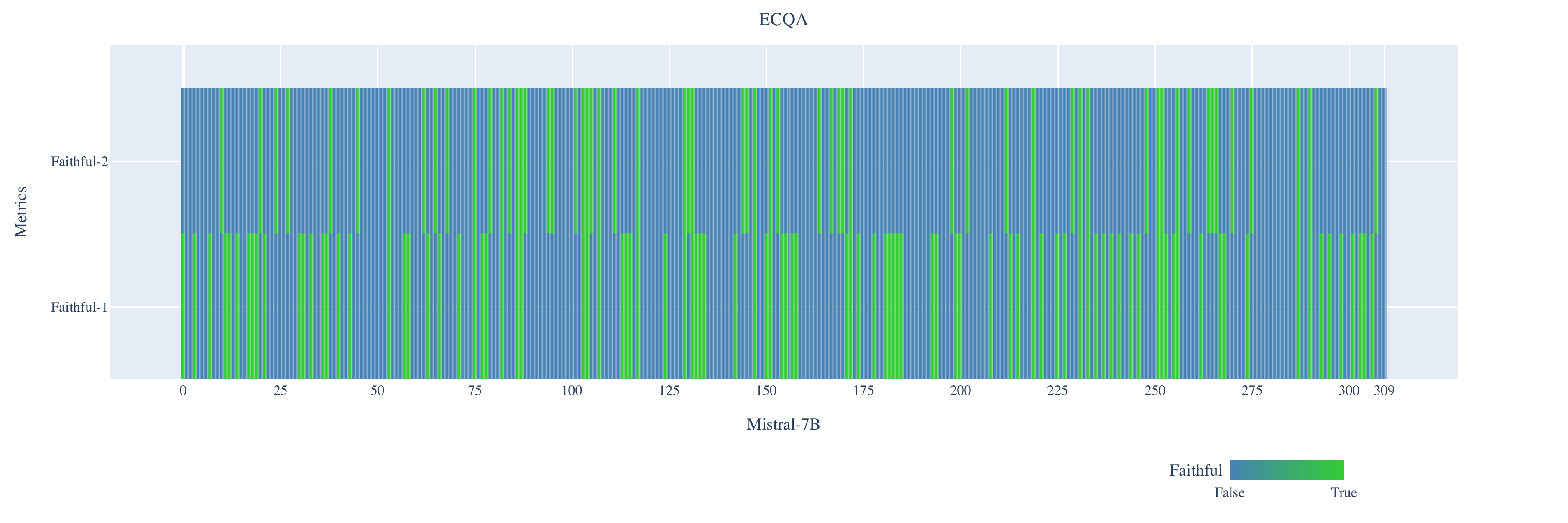}
    \caption{Binary heatmap visualization of \textsc{ecqa} test samples comparing two faithfulness metrics using the \texttt{Mistral-7B} model. Each column represents a test sample, while rows correspond to the two faithfulness metrics. Cells are color-coded to indicate binary outcomes: green = True, blue = False.}
    \label{fig:binary-heatmap-ecqa-mistral}
\end{figure*}

\FloatBarrier

\FloatBarrier
\section{Further Results} \label{app:further_results}
\subsection{Random Selection Results}

\begin{figure*}[h!]
    \centering
    \includegraphics[width=0.99\linewidth]{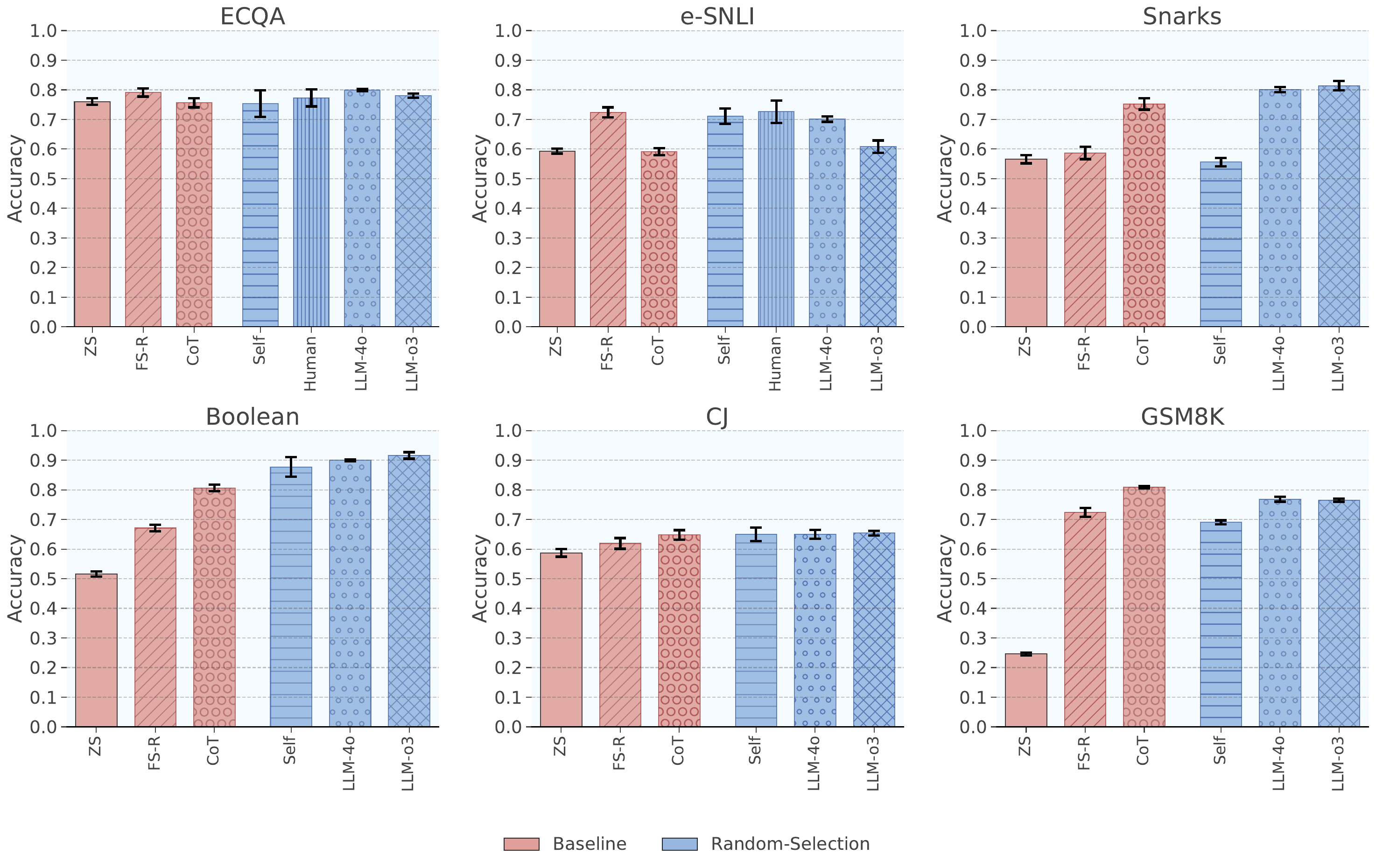}
    \caption{NLE type comparison under random selection (RQ1). Each panel reports results for one dataset. Bars present mean accuracy across five runs, averaged over the four evaluation models. Error bars indicate the standard deviation across runs. Baselines include ZS, FS-R, and CoT, while random-selection NLE conditions include Self-NLEs, Human-NLEs where available, LLM-4o, and LLM-o3. Bar fill patterns distinguish the prompting conditions: solid fill (ZS), diagonal hatching (FS-R), circular hatching (CoT), horizontal hatching (Self-NLE), vertical hatching (Human-NLE), dotted hatching (LLM-4o), and cross-hatching (LLM-o3).}
    \label{fig:random_comparison_plot}
\end{figure*}

\begin{table*}[!h]
\caption{
Model Specific results for the \textbf{\textit{Error}} sample selection setup, showing Random Selection performance. We report self-explanation (\textit{Self-exp.}) and LLM explanation two variants (\texttt{4o-mini} and \texttt{o3-mini}). Dashes (--) indicate unavailable human explanations for non-explainable datasets. Highest values in each row are highlighted in \textbf{bold}. 
}
\centering
\scriptsize
\setlength\tabcolsep{5pt}
\begin{tabular}{l|l|c|c|c|cccc}
\toprule
\multirow{3}{*}{Dataset} & \multirow{3}{*}{Models} & \multirow{3}{*}{Zero-shot}
& \multicolumn{1}{c|}{Few-shot}
& \multirow{3}{*}{CoT}
& \multicolumn{4}{c}{Random} \\
\cmidrule(lr){4-4}
\cmidrule(lr){6-9}
& & & Rand. & 
& Self-exp. & Human & \texttt{4o-mini}& \texttt{o3-mini} \\
\midrule

\multirow{5}{*}{\textsc{ecqa}}
& \texttt{4o-mini}   & 0.778 & 0.826 & 0.784 & 0.828 & 0.814 & \textbf{0.832} & 0.831 \\
& \texttt{llama-70B}  & 0.826 & 0.832 & 0.816 & 0.745 & 0.812 & \textbf{0.859} & 0.856 \\
& \texttt{llama-8B}   & 0.751 & 0.792 & 0.759 & 0.780 & 0.806 & \textbf{0.831} & 0.794 \\
& \texttt{mistral-7B} & 0.684 & \textbf{0.715} & 0.665 & 0.660 & 0.659 & 0.676 & 0.638 \\
& Avg        & 0.760 & 0.791 & 0.756 & 0.753 & 0.773 & \textbf{0.799} & 0.780 \\
\midrule

\multirow{5}{*}{e-\textsc{snli}}
& \texttt{4o-mini}   & 0.744 & 0.823 & 0.722 & 0.826 & 0.811 & 0.806 & \textbf{0.852} \\
& \texttt{llama-70B}  & 0.594 & 0.735 & 0.646 & 0.789 & \textbf{0.807} & 0.722 & 0.435 \\
& \texttt{llama-8B}   & 0.439 & 0.660 & 0.488 & 0.639 & 0.678 & \textbf{0.742} & 0.675 \\
& \texttt{mistral-7B} & 0.596 & \textbf{0.679} & 0.510 & 0.591 & 0.609 & 0.535 & 0.472 \\
& Avg        & 0.593 & 0.724 & 0.591 & 0.711 & \textbf{0.726} & 0.701 & 0.608 \\
\midrule

\multirow{5}{*}{\textsc{snarks}}
& \texttt{4o-mini}   & 0.652 & 0.609 & 0.818 & 0.512 & -- & 0.805 & \textbf{0.856} \\
& \texttt{llama-70B}  & 0.579 & 0.781 & \textbf{0.897} & 0.816 & -- & 0.863 & 0.886 \\
& \texttt{llama-8B}   & 0.559 & 0.494 & 0.709 & 0.418 & -- & \textbf{0.783} & 0.766 \\
& \texttt{mistral-7B} & 0.473 & 0.464 & 0.585 & 0.477 & -- & \textbf{0.751} & 0.746 \\
& Avg        & 0.566 & 0.587 & 0.752 & 0.556 & -- & 0.801 & \textbf{0.814} \\
\midrule

\multirow{5}{*}{\textsc{boolean}}
& \texttt{4o-mini}   & 0.677 & 0.748 & 0.962 & 0.970 & -- & 0.964 & \textbf{0.974} \\
& \texttt{llama-70B}  & 0.522 & 0.619 & \textbf{0.989} & 0.935 & -- & 0.981 & 0.985 \\
& \texttt{llama-8B}   & 0.501 & 0.729 & 0.826 & 0.872 & -- & \textbf{0.926} & 0.905 \\
& \texttt{mistral-7B} & 0.366 & 0.590 & 0.449 & 0.729 & -- & 0.730 & \textbf{0.800} \\
& Avg        & 0.516 & 0.671 & 0.806 & 0.877 & -- & 0.900 & \textbf{0.916} \\
\midrule

\multirow{5}{*}{\textsc{cj}}
& \texttt{4o-mini}   & 0.675 & \textbf{0.702} & 0.696 & 0.695 & -- & 0.677 & 0.693 \\
& \texttt{llama-70B}  & 0.667 & 0.665 & \textbf{0.726} & 0.650 & -- & 0.688 & 0.688 \\
& \texttt{llama-8B}   & 0.525 & 0.551 & 0.588 & \textbf{0.640} & -- & 0.588 & \textbf{0.640} \\
& \texttt{mistral-7B} & 0.481 & 0.559 & 0.581 & 0.614 & -- & \textbf{0.645} & 0.597 \\
& Avg        & 0.587 & 0.619 & 0.648 & 0.650 & -- & 0.650 & \textbf{0.654} \\
\midrule

\multirow{5}{*}{\textsc{gsm8k}}
& \texttt{4o-mini}   & 0.365 & 0.899 & \textbf{0.927} & 0.903 & -- & 0.890 & 0.902 \\
& \texttt{llama-70B}  & 0.388 & 0.718 & \textbf{0.954} & 0.552 & -- & 0.871 & 0.889 \\
& \texttt{llama-8B}   & 0.160 & \textbf{0.825} & 0.821 & 0.816 & -- & 0.798 & 0.773 \\
& \texttt{mistral-7B} & 0.071 & 0.455 & \textbf{0.532} & 0.487 & -- & 0.511 & 0.494 \\
& Avg        & 0.246 & 0.724 & \textbf{0.809} & 0.690 & -- & 0.768 & 0.765 \\
\midrule

All datasets & Avg
& 0.545 & 0.686 & 0.727 & 0.706 & 0.750 & \textbf{0.770} & 0.756 \\
\bottomrule
\end{tabular}
\label{tab:random-nle-comp}
\end{table*}

\FloatBarrier
\clearpage
\subsection{Self-NLEs Results Across Setups} \label{app:self-nle-across-setups}

\begin{table*}[!ht]
\caption{
Results for the \textit{Error} sample-selection setup. We report mean performance over five runs for each model (4o-mini, \texttt{Llama-70B}, \texttt{Llama-8B}, and \texttt{Mistral-7B}) on five datasets across the following prompting and explanation-selection conditions: Few-shot Rand., Few-shot Faith.-1, Few-shot Faith.-2, Random, Faithful-$fm_1$, Faithful-$fm_2$, Lowest-Faithful-$fm_1$, and Lowest-Faithful-$fm_2$. The highest value in each row is shown in \textbf{bold}. Statistical significance is assessed using the Stuart--Maxwell test, where $^{*}p < 0.10$, $^{**}p < 0.05$, and $^{***}p < 0.01$. The Few-shot Random baseline (\textit{FS-R}) serves as the reference for statistical tests involving self-NLEs (\textit{Self-exp.}). FS-R follows Setup~1, using randomly selected few-shot input-output pairs without rationales. Few-shot Faith.-1 (\textit{FS-$fm_1$}) and Few-shot Faith.-2 (\textit{FS-$fm_2$}) are matched no-rationale baselines: they use the same selected input-output pairs $(x,y)$ as the corresponding $fm_1$- and $fm_2$-based conditions, respectively, while omitting the rationales.
}
\centering
\scriptsize
\setlength\tabcolsep{3pt} 
\resizebox{\textwidth}{!}{
\begin{tabular}{l|l|ccc|c|c|c|c|c}
\toprule
\multirow{3}{*}{Dataset} & \multirow{3}{*}{Models} & \multicolumn{3}{c|}{Few-shot} & Random & Faithful-fm1 & Faithful-fm2 & Lowest-Faithful-fm1 & Lowest-Faithful-fm2 \\
\cmidrule(lr){3-5}
\cmidrule(lr){6-6}
\cmidrule(lr){7-7}
\cmidrule(lr){8-8}
\cmidrule(lr){9-9}
\cmidrule(lr){10-10}
& & Rand. (Ref) & Faith.-1 & Faith.-2 & Self-NLE & Self-NLE & Self-NLE & Self-NLE & Self-NLE \\
\midrule
\multirow{5}{*}{\textsc{ecqa}}
& \texttt{4o-mini}   & 0.826  & 0.820 & 0.823 & 0.828\textsuperscript{***} & $0.824^{***}$ & $0.820^{***}$ & $0.824^{***}$ & \textbf{0.829} \\
& \texttt{llama-70B}  & 0.832  & 0.847 & 0.838 & $0.745^{***}$ & $0.782^{***}$ & $0.813^{***}$ & $0.659^{***}$ & \textbf{0.864\textsuperscript{***}} \\
& \texttt{llama-8B}   & 0.792  & \textbf{0.799} & 0.794 & 0.780 & $0.759^{***}$ & $0.780^{**}$ & $0.230^{***}$ & 0.782 \\
& \texttt{mistral-7B} & 0.715  & \textbf{0.749} & 0.730 & $0.660^{**}$ & $0.696^{**}$ & $0.686^{**}$ & 0.678 & $0.724^{***}$ \\
& Avg        & 0.791  & \textbf{0.804} & 0.796 & 0.753 & 0.765 & 0.775 & 0.598 & 0.799 \\
\midrule
\multirow{5}{*}{e-\textsc{snli}}
& \texttt{4o-mini}   & 0.823  & 0.827 & 0.830 & $0.826^{***}$ & $0.790^{***}$ & $0.818^{**}$ & \textbf{0.846\textsuperscript{***}} & 0.832 \\
& \texttt{llama-70B}  & 0.735  & 0.806 & 0.741 & $0.789^{***}$ & \textbf{0.816\textsuperscript{***}} & $0.720^{***}$ & $0.774^{**}$ & $0.704^{***}$ \\
& \texttt{llama-8B}   & 0.660  & \textbf{0.730} & 0.633 & $0.639^{***}$ & $0.604^{***}$ & $0.633^{***}$ & $0.712^{***}$ & $0.572^{***}$ \\
& \texttt{mistral-7B} & 0.679  & 0.584 & \textbf{0.794} & $0.591^{***}$ & $0.642^{***}$ & $0.691^{***}$ & $0.607^{***}$ & $0.591^{***}$ \\
& Avg        & 0.724  & 0.737 & \textbf{0.749} & 0.711 & 0.713 & 0.716 & 0.735 & 0.675 \\
\midrule
\multirow{5}{*}{\textsc{snarks}}
& \texttt{4o-mini}   & 0.609  & 0.506 & 0.470 & $0.512^{***}$ & $0.514^{***}$ & $0.563^{***}$ & 0.654 & \textbf{0.775\textsuperscript{***}} \\
& \texttt{llama-70B}  & 0.781  & 0.757 & 0.799 & $0.816^{*}$ & 0.794 & \textbf{0.822\textsuperscript{***}} & $0.820^{***}$ & 0.814 \\
& \texttt{llama-8B}   & 0.494  & \textbf{0.618} & 0.497 & $0.418^{*}$ & $0.469^{***}$ & $0.553^{*}$ & 0.386 & $0.589^{***}$ \\
& \texttt{mistral-7B} & 0.464  & \textbf{0.575} & 0.440 & 0.477 & 0.431 & 0.441 & 0.394 & 0.480\\
& Avg        & 0.587  & 0.614 & 0.551 & 0.556 & 0.552 & 0.595 & 0.564 & \textbf{0.665} \\
\midrule
\multirow{5}{*}{\textsc{boolean}}
& \texttt{4o-mini}   & 0.748  & 0.847 & 0.900 & \textbf{0.970\textsuperscript{***}} & $0.957^{**}$ & $0.943^{***}$ & 0.930 & $0.959^{***}$ \\
& \texttt{llama-70B}  & 0.619  & 0.623 & 0.696 & $0.935^{*}$ & $0.942^{*}$ & \textbf{0.976\textsuperscript{***}} & $0.809^{***}$ & $0.972^{***}$ \\
& \texttt{llama-8B}   & 0.729  & 0.818 & 0.819 & $0.872^{***}$ & \textbf{0.893\textsuperscript{***}} & $0.852^{***}$ & $0.873^{***}$ & $0.873^{***}$ \\
& \texttt{mistral-7B} & 0.590  & 0.640 & 0.730 & $0.729^{***}$ & 0.683 & $0.745^{***}$ & $0.766^{*}$ & \textbf{0.795} \\
& Avg        & 0.671  & 0.732 & 0.786 & 0.877 & 0.869 & 0.879 & 0.845 & \textbf{0.900} \\
\midrule
\multirow{5}{*}{\textsc{cj}}
& \texttt{4o-mini}   & 0.702  & 0.695 & 0.688 & 0.695 & 0.685 & \textbf{0.703} & 0.672 & 0.683 \\
& \texttt{llama-70B}  & 0.665  & 0.691 & 0.670 & 0.650 & $0.705^{**}$ & 0.678 & 0.682 & \textbf{0.745\textsuperscript{***}} \\
& \texttt{llama-8B}   & 0.551  & 0.518 & 0.606 & \textbf{0.640\textsuperscript{***}} & 0.578 & 0.599 & $0.630^{***}$ & 0.555 \\
& \texttt{mistral-7B} & 0.559  & 0.599 & 0.516 & \textbf{0.614\textsuperscript{**}} & $0.562^{***}$ & 0.568 & $0.589^{***}$ & $0.602^{***}$ \\
& Avg        & 0.619  & 0.626 & 0.619 & \textbf{0.650} & 0.633 & 0.637 & 0.643 & 0.646 \\
\midrule
All datasets & Avg & 0.679  & 0.702 & 0.700 & 0.709 & 0.706 & 0.720 & 0.677 & \textbf{0.737} \\
\bottomrule
\end{tabular}
}
\label{tab:stat-test-step-b-only}
\end{table*}

\FloatBarrier

\clearpage

\clearpage
\subsection{Full Results Across Setups} \label{app:full-results-across-setups}

\begin{table*}[!h]
\caption{
Results for the \textbf{\textit{Error}} sample selection setup, showing the performance of average of five runs of each model (4o-mini, \texttt{Llama-70B}, \texttt{Llama-8B}, and \texttt{Mistral-7B}) on each dataset across explanation selection setups (Random, Faithful-1, Lowest-Faithful-1, and Lowest-Faithful-2). For each selection setup, we report self-explanations (\textit{Self-exp.}), along with human and LLM-generated explanations (\texttt{4o-mini} and \texttt{o3-mini}). Dashes (--) denote unavailable human explanations for non-explainable datasets. Highest values in each row are highlighted in \textbf{bold}.
}
\centering
\scriptsize
\setlength\tabcolsep{2.5pt}
\resizebox{\textwidth}{!}{
\begin{tabular}{l|l|c|ccc|c|cccc|cccc|cccc|cccc|cccc}
\toprule
\multirow{3}{*}{Dataset} & \multirow{3}{*}{Models} & \multirow{3}{*}{Zero-shot}
& \multicolumn{3}{c|}{Few-shot}
& \multirow{3}{*}{CoT}
& \multicolumn{4}{c|}{Random}
& \multicolumn{4}{c|}{Faithful-fm1}
& \multicolumn{4}{c|}{Faithful-fm2}
& \multicolumn{4}{c|}{Lowest-Faithful-fm1}
& \multicolumn{4}{c}{Lowest-Faithful-fm2} \\
\cmidrule(lr){4-6}
\cmidrule(lr){8-11}
\cmidrule(lr){12-15}
\cmidrule(lr){16-19}
\cmidrule(lr){20-23}
\cmidrule(lr){24-27}
& & & Rand. & Faith.-1 & Faith.-2 & 
& Self-exp. & Human & \texttt{4o-mini}& \texttt{o3-mini}
& Self-exp. & Human & \texttt{4o-mini}& \texttt{o3-mini}
& Self-exp. & Human & \texttt{4o-mini}& \texttt{o3-mini}
& Self-exp. & Human & \texttt{4o-mini}& \texttt{o3-mini}
& Self-exp. & Human & \texttt{4o-mini}& \texttt{o3-mini} \\
\midrule

\multirow{5}{*}{\textsc{ecqa}}
& \texttt{4o-mini}   & 0.778 & 0.826 & 0.820 & 0.823 & 0.784
& 0.828 & 0.814 & 0.832 & 0.831
& 0.824 & 0.823 & 0.822 & 0.823
& 0.820 & 0.810 & 0.814 & 0.812
& 0.824 & 0.816 & 0.816 & 0.829
& 0.829 & 0.829 & \textbf{0.838} & 0.834 \\
& \texttt{llama-70B}  & 0.826 & 0.832 & 0.847 & 0.838 & 0.816
& 0.745 & 0.812 & 0.859 & 0.856
& 0.782 & 0.861 & 0.847 & 0.858
& 0.813 & 0.860 & 0.859 & \textbf{0.864}
& 0.659 & 0.854 & 0.854 & 0.859
& 0.862 & 0.848 & 0.848 & 0.855 \\
& \texttt{llama-8B}   & 0.751 & 0.792 & 0.799 & 0.794 & 0.759
& 0.780 & 0.806 & \textbf{0.831} & 0.794
& 0.759 & 0.775 & 0.791 & 0.791
& 0.780 & 0.745 & 0.583 & 0.774
& 0.230 & 0.786 & 0.786 & 0.785
& 0.782 & 0.781 & 0.792 & 0.801 \\
& \texttt{mistral-7B} & 0.684 & 0.715 & \textbf{0.749} & 0.730 & 0.665
& 0.660 & 0.659 & 0.676 & 0.638
& 0.696 & 0.635 & 0.621 & 0.575
& 0.686 & 0.717 & 0.695 & 0.673
& 0.678 & 0.744 & 0.665 & 0.731
& 0.724 & 0.719 & 0.692 & 0.735 \\
& Avg        & 0.760 & 0.791 & 0.804 & 0.796 & 0.756
& 0.753 & 0.773 & 0.799 & 0.780
& 0.765 & 0.774 & 0.770 & 0.762
& 0.775 & 0.783 & 0.738 & 0.781
& 0.598 & 0.800 & 0.780 & 0.801
& 0.799 & 0.794 & 0.793 & \textbf{0.806} \\
\midrule

\multirow{5}{*}{e-\textsc{snli}}
& \texttt{4o-mini}   & 0.744 & 0.823 & 0.827 & 0.830 & 0.722
& 0.826 & 0.811 & 0.806 & 0.852
& 0.790 & 0.852 & 0.827 & 0.815
& 0.818 & 0.832 & 0.802 & 0.773
& 0.846 & \textbf{0.857} & \textbf{0.857} & 0.833
& 0.832 & 0.825 & 0.810 & 0.806 \\
& \texttt{llama-70B}  & 0.594 & 0.735 & 0.806 & 0.741 & 0.646
& 0.789 & 0.807 & 0.722 & 0.435
& 0.816 & \textbf{0.851} & 0.783 & 0.774
& 0.720 & 0.742 & 0.750 & 0.742
& 0.774 & 0.805 & 0.805 & 0.683
& 0.704 & 0.780 & 0.749 & 0.719 \\
& \texttt{llama-8B}   & 0.439 & 0.660 & 0.730 & 0.633 & 0.488
& 0.639 & 0.678 & \textbf{0.742} & 0.675
& 0.604 & 0.702 & 0.720 & 0.620
& 0.633 & 0.691 & 0.617 & 0.559
& 0.712 & 0.681 & 0.681 & 0.726
& 0.572 & 0.714 & 0.591 & 0.562 \\
& \texttt{mistral-7B} & 0.596 & 0.679 & 0.584 & \textbf{0.794} & 0.510
& 0.591 & 0.609 & 0.535 & 0.472
& 0.642 & 0.642 & 0.641 & 0.674
& 0.691 & 0.698 & 0.699 & 0.730
& 0.607 & 0.680 & 0.580 & 0.572
& 0.591 & 0.733 & 0.764 & 0.698 \\
& Avg        & 0.593 & 0.724 & 0.737 & 0.749 & 0.591
& 0.711 & 0.726 & 0.701 & 0.608
& 0.713 & 0.762 & 0.743 & 0.721
& 0.716 & 0.741 & 0.717 & 0.701
& 0.735 & 0.756 & 0.731 & 0.704
& 0.675 & \textbf{0.763} & 0.729 & 0.696 \\
\midrule

\multirow{5}{*}{\textsc{snarks}}
& \texttt{4o-mini}   & 0.652 & 0.609 & 0.506 & 0.470 & 0.818
& 0.512 & -- & 0.805 & \textbf{0.856}
& 0.514 & -- & 0.833 & 0.810
& 0.563 & -- & 0.818 & 0.822
& 0.654 & -- & 0.823 & 0.822
& 0.775 & -- & 0.802 & 0.817 \\
& \texttt{llama-70B}  & 0.579 & 0.781 & 0.757 & 0.799 & 0.897
& 0.816 & -- & 0.863 & 0.886
& 0.794 & -- & 0.886 & 0.914
& 0.822 & -- & 0.867 & 0.907
& 0.820 & -- & 0.874 & 0.878
& 0.814 & -- & 0.888 & \textbf{0.934} \\
& \texttt{llama-8B}   & 0.559 & 0.494 & 0.618 & 0.497 & 0.709
& 0.418 & -- & 0.783 & 0.766
& 0.469 & -- & 0.709 & 0.749
& 0.553 & -- & 0.773 & \textbf{0.790}
& 0.386 & -- & 0.701 & 0.771
& 0.589 & -- & 0.739 & 0.745 \\
& \texttt{mistral-7B} & 0.473 & 0.464 & 0.575 & 0.440 & 0.585
& 0.477 & -- & 0.751 & 0.746
& 0.431 & -- & 0.730 & \textbf{0.791}
& 0.441 & -- & 0.762 & 0.754
& 0.394 & -- & 0.701 & 0.721
& 0.480 & -- & 0.721 & 0.680 \\
& Avg        & 0.566 & 0.587 & 0.614 & 0.551 & 0.752
& 0.556 & -- & 0.801 & 0.814
& 0.552 & -- & 0.792 & 0.816
& 0.595 & -- & 0.805 & \textbf{0.818}
& 0.564 & -- & 0.775 & 0.798
& 0.665 & -- & 0.788 & 0.794 \\
\midrule

\multirow{5}{*}{\textsc{boolean}}
& \texttt{4o-mini}   & 0.677 & 0.748 & 0.847 & 0.900 & 0.962
& 0.970 & -- & 0.964 & 0.974
& 0.957 & -- & 0.978 & 0.970
& 0.943 & -- & 0.977 & 0.972
& 0.930 & -- & 0.961 & 0.965
& 0.959 & -- & 0.964 & \textbf{0.983} \\
& \texttt{llama-70B}  & 0.522 & 0.619 & 0.623 & 0.696 & 0.989
& 0.935 & -- & 0.981 & 0.985
& 0.942 & -- & 0.985 & 0.979
& 0.976 & -- & 0.990 & 0.991
& 0.809 & -- & 0.983 & 0.975
& 0.972 & -- & 0.987 & 0.984 \\
& \texttt{llama-8B}   & 0.501 & 0.729 & 0.818 & 0.819 & 0.826
& 0.872 & -- & 0.926 & 0.905
& 0.893 & -- & 0.913 & \textbf{0.930}
& 0.852 & -- & 0.902 & 0.924
& 0.873 & -- & 0.922 & 0.912
& 0.873 & -- & 0.872 & 0.889 \\
& \texttt{mistral-7B} & 0.366 & 0.590 & 0.640 & 0.730 & 0.449
& 0.729 & -- & 0.730 & 0.800
& 0.683 & -- & 0.827 & 0.763
& 0.745 & -- & 0.777 & 0.752
& 0.766 & -- & 0.750 & 0.793
& 0.795 & -- & \textbf{0.854} & 0.720 \\
& Avg        & 0.516 & 0.671 & 0.732 & 0.786 & 0.806
& 0.877 & -- & 0.900 & 0.916
& 0.869 & -- & \textbf{0.926} & 0.910
& 0.879 & -- & 0.886 & 0.910
& 0.845 & -- & 0.904 & 0.911
& 0.900 & -- & 0.919 & 0.894 \\
\midrule

\multirow{5}{*}{\textsc{cj}}
& \texttt{4o-mini}   & 0.675 & 0.702 & 0.695 & 0.688 & 0.696
& 0.695 & -- & 0.677 & 0.693
& 0.685 & -- & 0.650 & 0.616
& 0.703 & -- & 0.648 & 0.668
& 0.672 & -- & 0.660 & 0.667
& 0.683 & -- & 0.636 & 0.658 \\
& \texttt{llama-70B}  & 0.667 & 0.665 & 0.691 & 0.670 & 0.726
& 0.650 & -- & 0.688 & 0.688
& 0.705 & -- & 0.705 & 0.727
& 0.678 & -- & 0.721 & 0.711
& 0.682 & -- & 0.697 & 0.676
& 0.745 & -- & \textbf{0.751} & 0.748 \\
& \texttt{llama-8B}   & 0.525 & 0.551 & 0.518 & 0.606 & 0.588
& 0.640 & -- & 0.588 & 0.640
& 0.578 & -- & 0.503 & 0.560
& 0.599 & -- & 0.573 & \textbf{0.656}
& 0.630 & -- & 0.553 & 0.605
& 0.555 & -- & 0.512 & 0.544 \\
& \texttt{mistral-7B} & 0.481 & 0.559 & 0.599 & 0.516 & 0.581
& 0.614 & -- & 0.645 & 0.597
& 0.562 & -- & 0.647 & 0.633
& 0.568 & -- & 0.596 & 0.579
& 0.589 & -- & 0.609 & 0.639
& 0.602 & -- & \textbf{0.654} & 0.644 \\
& Avg        & 0.587 & 0.619 & 0.626 & 0.619 & 0.648
& 0.650 & -- & 0.650 & \textbf{0.654}
& 0.633 & -- & 0.626 & 0.634
& 0.637 & -- & 0.634 & \textbf{0.654}
& 0.643 & -- & 0.630 & 0.647
& 0.646 & -- & 0.638 & 0.638 \\
\midrule

\multirow{5}{*}{\textsc{gsm8k}}
& \texttt{4o-mini}   & 0.365    & 0.899    & 0.908    & \textbf{0.928}    & 0.927    
& 0.903    & --  & 0.890    & 0.902    
& 0.893    & --  & 0.881    & 0.877    
& 0.916    & --  & 0.884    & 0.884    
& 0.876    & --  & 0.887    & 0.867  
& 0.377    & --  & 0.892    & 0.893    \\
& \texttt{llama-70B}  & 0.388    & 0.718    & 0.884   & 0.825    & \textbf{0.954}    
& 0.552    & --  & 0.871    & 0.889    
& 0.902    & --  & 0.843    & 0.868    
& 0.923    & --  & 0.812    & 0.817    
& 0.488    & --  & 0.850    & 0.864    
& 0.272    & --  & 0.863    & 0.849    \\
& \texttt{llama-8B}   & 0.160    & 0.825    & 0.829    & \textbf{0.842}    & 0.821    
& 0.816    & --  & 0.798    & 0.773    
& 0.825    & --  & 0.691    & 0.732    
& 0.825    & --  & 0.605    & 0.550    
& 0.783    & --  & 0.795    & 0.798  
& 0.436    & --  & 0.797    & 0.783    \\
& \texttt{mistral-7B} & 0.071    & 0.455    & 0.437    & 0.448    & \textbf{0.532}    
& 0.487    & --  & 0.511    & 0.494    
& 0.430    & --  & 0.402    & 0.345    
& 0.082    & --  & 0.392    & 0.427    
& 0.421    & --  & 0.394    & 0.379    
& 0.141    & --  & 0.512    & 0.466    \\
& Avg
& 0.246 & 0.724 & 0.765 & 0.761 & \textbf{0.809}
& 0.690 & --  & 0.768 & 0.765
& 0.763 & --  & 0.704 & 0.706
& 0.686 & --  & 0.673 & 0.670
& 0.642 & --  & 0.732 & 0.727
& 0.307 & --  & 0.766 & 0.748 \\
\midrule

All datasets & Avg
& 0.545 & 0.686 & 0.713 & 0.711 & 0.727
& 0.706 & 0.750 & 0.770 & 0.756
& 0.716 & 0.768 & 0.760 & 0.758
& 0.715 & 0.762 & 0.747 & 0.755
& 0.671 & \textbf{0.778} & 0.759 & 0.765
& 0.665 & \textbf{0.778} & 0.772 & 0.764 \\
\bottomrule
\end{tabular}
}
\label{tab:big-table-error-five-runs}
\end{table*}

\begin{table*}[!h]
\caption{
Average results for the \textbf{\textit{Error}} sample selection setup. Values are averaged over the four evaluation models and five runs. We report baseline prompting, random NLE selection, faithful selection using \(fm_1\) and \(fm_2\), and lowest-faithful selection using \(fm_1\) and \(fm_2\). Dashes (--) denote unavailable human explanations for datasets without human rationales.
}
\centering
\small
\setlength\tabcolsep{2.8pt}
\resizebox{\textwidth}{!}{
\begin{tabular}{l|c|ccc|c|cccc|cccc|cccc|cccc|cccc}
\toprule
\multirow{3}{*}{Dataset} 
& \multirow{3}{*}{Zero-shot}
& \multicolumn{3}{c|}{Few-shot}
& \multirow{3}{*}{CoT}
& \multicolumn{4}{c|}{Random}
& \multicolumn{4}{c|}{Faithful-\(fm_1\)}
& \multicolumn{4}{c|}{Faithful-\(fm_2\)}
& \multicolumn{4}{c|}{Lowest-Faithful-\(fm_1\)}
& \multicolumn{4}{c}{Lowest-Faithful-\(fm_2\)} \\
\cmidrule(lr){3-5}
\cmidrule(lr){7-10}
\cmidrule(lr){11-14}
\cmidrule(lr){15-18}
\cmidrule(lr){19-22}
\cmidrule(lr){23-26}
& & Rand. & Faith.-1 & Faith.-2
& 
& Self-exp. & Human & \texttt{4o-mini}& \texttt{o3-mini}
& Self-exp. & Human & \texttt{4o-mini}& \texttt{o3-mini}
& Self-exp. & Human & \texttt{4o-mini}& \texttt{o3-mini}
& Self-exp. & Human & \texttt{4o-mini}& \texttt{o3-mini}
& Self-exp. & Human & \texttt{4o-mini}& \texttt{o3-mini} \\
\midrule

\textsc{ecqa}
& 0.760 & 0.791 & 0.804 & 0.796 & 0.756
& 0.753 & 0.773 & 0.799 & 0.780
& 0.765 & 0.774 & 0.770 & 0.762
& 0.775 & 0.783 & 0.738 & 0.781
& 0.598 & 0.800 & 0.780 & 0.801
& 0.799 & 0.794 & 0.793 & \textbf{0.806} \\

e-\textsc{snli}
& 0.593 & 0.724 & 0.737 & 0.749 & 0.591
& 0.711 & 0.726 & 0.701 & 0.608
& 0.713 & 0.762 & 0.743 & 0.721
& 0.716 & 0.741 & 0.717 & 0.701
& 0.735 & 0.756 & 0.731 & 0.704
& 0.675 & \textbf{0.763} & 0.729 & 0.696 \\

\textsc{snarks}
& 0.566 & 0.587 & 0.614 & 0.551 & 0.752
& 0.556 & -- & 0.801 & 0.814
& 0.552 & -- & 0.792 & 0.816
& 0.595 & -- & 0.805 & \textbf{0.818}
& 0.564 & -- & 0.775 & 0.798
& 0.665 & -- & 0.788 & 0.794 \\

\textsc{boolean}
& 0.516 & 0.671 & 0.732 & 0.786 & 0.806
& 0.877 & -- & 0.900 & 0.916
& 0.869 & -- & \textbf{0.926} & 0.910
& 0.879 & -- & 0.886 & 0.910
& 0.845 & -- & 0.904 & 0.911
& 0.900 & -- & 0.919 & 0.894 \\

\textsc{cj}
& 0.587 & 0.619 & 0.626 & 0.619 & 0.648
& 0.650 & -- & 0.650 & \textbf{0.654}
& 0.633 & -- & 0.626 & 0.634
& 0.637 & -- & 0.634 & \textbf{0.654}
& 0.643 & -- & 0.630 & 0.647
& 0.646 & -- & 0.638 & 0.638 \\

\textsc{gsm8k}
& 0.246 & 0.724 & 0.765 & 0.761 & \textbf{0.809}
& 0.690 & -- & 0.768 & 0.765
& 0.763 & -- & 0.704 & 0.706
& 0.686 & -- & 0.673 & 0.670
& 0.642 & -- & 0.732 & 0.727
& 0.307 & -- & 0.766 & 0.748 \\

\midrule
All datasets
& 0.545 & 0.686 & 0.713 & 0.711 & 0.727
& 0.706 & 0.750 & 0.770 & 0.756
& 0.716 & 0.768 & 0.760 & 0.758
& 0.715 & 0.762 & 0.747 & 0.755
& 0.671 & \textbf{0.778} & 0.759 & 0.765
& 0.665 & \textbf{0.778} & 0.772 & 0.764 \\
\bottomrule
\end{tabular}
}
\label{tab:avg-results-error-five-runs}
\end{table*}

\begin{table*}[!ht]
\centering
\scriptsize
\setlength{\tabcolsep}{3.5pt}
\caption{Dataset-level change in accuracy relative to random selection by NLE source. Setting 1 corresponds to random selection, Setting 2 to most-faithful selection using $fm_1$ or $fm_2$, and Setting 3 to lowest-faithful selection using $fm_1$ or $fm_2$. The Random column reports mean accuracy under random selection, averaged over the four evaluation models and five runs. The remaining columns report $\Delta$ accuracy relative to Random, with the absolute accuracy under that setup shown in parentheses. Positive $\Delta$ values indicate improvement over random selection. \textbf{Bold} values indicate the highest absolute accuracy within each dataset and selection \textbf{column}. Human-NLE rows are shown only for \textsc{ecqa} and e-\textsc{snli}, where human rationales are available.}
\label{tab:nle-source-selection-delta-by-dataset}
\begin{tabular}{llccccc}
\toprule
& & \textbf{Setting 1} 
& \multicolumn{2}{c}{\textbf{Setting 2}} 
& \multicolumn{2}{c}{\textbf{Setting 3}} \\
\cmidrule(lr){3-3} \cmidrule(lr){4-5} \cmidrule(lr){6-7}
\textbf{Dataset} & \textbf{NLE source}
& \textbf{Random}
& \shortstack{\textbf{$\Delta$ Faithful-}\\\textbf{$fm_1$}}
& \shortstack{\textbf{$\Delta$ Faithful-}\\\textbf{$fm_2$}}
& \shortstack{\textbf{$\Delta$ Lowest-}\\\textbf{$fm_1$}}
& \shortstack{\textbf{$\Delta$ Lowest-}\\\textbf{$fm_2$}} \\
\midrule

\multirow{4}{*}{\textsc{ecqa}}
& Self-NLE  & 0.753 & $+0.012$ (0.765) & $+0.022$ (0.775) & $-0.155$ (0.598) & $+0.046$ (0.799) \\
& Human-NLE & 0.773 & $+0.001$ (\textbf{0.774}) & $+0.010$ (\textbf{0.783}) & $+0.027$ (0.800) & $+0.021$ (0.794) \\
& LLM-4o    & \textbf{0.799} & $-0.029$ (0.770) & $-0.061$ (0.738) & $-0.019$ (0.780) & $-0.006$ (0.793) \\
& LLM-o3    & 0.780 & $-0.018$ (0.762) & $+0.001$ (0.781) & $+0.021$ (\textbf{0.801}) & $+0.026$ (\textbf{0.806}) \\
\midrule

\multirow{4}{*}{e-\textsc{snli}}
& Self-NLE  & 0.711 & $+0.002$ (0.713) & $+0.005$ (0.716) & $+0.024$ (0.735) & $-0.036$ (0.675) \\
& Human-NLE & \textbf{0.726} & $+0.036$ (\textbf{0.762}) & $+0.015$ (\textbf{0.741}) & $+0.030$ (\textbf{0.756}) & $+0.037$ (\textbf{0.763}) \\
& LLM-4o    & 0.701 & $+0.042$ (0.743) & $+0.016$ (0.717) & $+0.030$ (0.731) & $+0.028$ (0.729) \\
& LLM-o3    & 0.608 & $+0.113$ (0.721) & $+0.093$ (0.701) & $+0.096$ (0.704) & $+0.088$ (0.696) \\
\midrule

\multirow{3}{*}{\textsc{snarks}}
& Self-NLE  & 0.556 & $-0.004$ (0.552) & $+0.039$ (0.595) & $+0.008$ (0.564) & $+0.109$ (0.665) \\
& LLM-4o    & 0.801 & $-0.009$ (0.792) & $+0.004$ (0.805) & $-0.026$ (0.775) & $-0.013$ (0.788) \\
& LLM-o3    & \textbf{0.814} & $+0.002$ (\textbf{0.816}) & $+0.004$ (\textbf{0.818}) & $-0.016$ (\textbf{0.798}) & $-0.020$ (\textbf{0.794}) \\
\midrule

\multirow{3}{*}{\textsc{boolean}}
& Self-NLE  & 0.877 & $-0.008$ (0.869) & $+0.002$ (0.879) & $-0.032$ (0.845) & $+0.023$ (0.900) \\
& LLM-4o    & 0.900 & $+0.026$ (\textbf{0.926}) & $-0.014$ (0.886) & $+0.004$ (0.904) & $+0.019$ (\textbf{0.919}) \\
& LLM-o3    & \textbf{0.916} & $-0.006$ (0.910) & $-0.006$ (\textbf{0.910}) & $-0.005$ (\textbf{0.911}) & $-0.022$ (0.894) \\
\midrule

\multirow{3}{*}{\textsc{cj}}
& Self-NLE  & 0.650 & $-0.017$ (0.633) & $-0.013$ (0.637) & $-0.007$ (0.643) & $-0.004$ (\textbf{0.646}) \\
& LLM-4o    & 0.650 & $-0.024$ (0.626) & $-0.016$ (0.634) & $-0.020$ (0.630) & $-0.012$ (0.638) \\
& LLM-o3    & \textbf{0.654} & $-0.020$ (\textbf{0.634}) & $+0.000$ (\textbf{0.654}) & $-0.007$ (\textbf{0.647}) & $-0.016$ (0.638) \\
\midrule

\multirow{3}{*}{\textsc{gsm8k}}
& Self-NLE  & 0.690 & $+0.073$ (\textbf{0.763}) & $-0.004$ (\textbf{0.686}) & $-0.048$ (0.642) & $-0.383$ (0.307) \\
& LLM-4o    & \textbf{0.768} & $-0.064$ (0.704) & $-0.095$ (0.673) & $-0.036$ (\textbf{0.732}) & $-0.002$ (\textbf{0.766}) \\
& LLM-o3    & 0.765 & $-0.059$ (0.706) & $-0.095$ (0.670) & $-0.038$ (0.727) & $-0.017$ (0.748) \\

\bottomrule
\end{tabular}
\end{table*}

\FloatBarrier

\begin{table*}[!h]
\caption{
Results for the \textbf{\textit{Error}} sample selection setup, reporting the \textbf{standard deviations} across five runs of each model (4o-mini, \texttt{Llama-70B}, \texttt{Llama-8B}, and \texttt{Mistral-7B}) on all datasets under different explanation selection strategies (Random, Faithful-1, Faithful-2, Lowest-Faithful-1, and Lowest-Faithful-2). For each selection setup, we report self-explanations (\textit{Self-exp.}), along with human and LLM-generated explanations (\texttt{4o-mini} and \texttt{o3-mini}). Dashes (--) indicate unavailable human explanations for non-explainable datasets. The highest standard deviation in each row is highlighted in \textbf{bold}.
}
\centering
\scriptsize
\setlength\tabcolsep{2pt}
\resizebox{\textwidth}{!}{
\begin{tabular}{l|l|c|ccc|c|cccc|cccc|cccc|cccc|cccc}
\toprule
\multirow{3}{*}{Dataset} & \multirow{3}{*}{Models} & \multirow{3}{*}{Zero-shot}
& \multicolumn{3}{c|}{Few-shot}
& \multirow{3}{*}{CoT}
& \multicolumn{4}{c|}{Random}
& \multicolumn{4}{c|}{Faithful-fm1}
& \multicolumn{4}{c|}{Faithful-fm2}
& \multicolumn{4}{c|}{Lowest-Faithful-fm1}
& \multicolumn{4}{c}{Lowest-Faithful-fm2} \\

\cmidrule(lr){4-6}
\cmidrule(lr){8-11}
\cmidrule(lr){12-15}
\cmidrule(lr){16-19}
\cmidrule(lr){20-23}
\cmidrule(lr){24-27}

& & & Rand. & Faith.-1 & Faith.-2 &
& Self-exp. & Human & \texttt{4o-mini}& \texttt{o3-mini}
& Self-exp. & Human & \texttt{4o-mini}& \texttt{o3-mini} 
& Self-exp. & Human & \texttt{4o-mini}& \texttt{o3-mini} 
& Self-exp. & Human & \texttt{4o-mini}& \texttt{o3-mini} 
& Self-exp. & Human & \texttt{4o-mini}& \texttt{o3-mini}  \\
\midrule

\textsc{ecqa} & 4o-mini
& 0.007 & 0.006 & 0.009 & 0.013 & 0.005
& 0.010 & 0.004 & 0.004 & 0.012
& 0.008 & 0.005 & 0.005 & 0.008
& 0.006 & 0.008 & 0.006 & 0.008
& 0.005 & 0.004 & 0.006 & 0.006
& 0.008 & 0.005 & 0.008 & 0.008 \\

& \texttt{llama-70B}
& 0.007 & 0.008 & 0.002 & 0.004 & 0.006
& 0.172  & 0.122 & 0.006 & 0.006
& 0.126 & 0.005 & 0.003 & 0.005
& 0.009 & 0.006 & 0.008 & 0.005
& \textbf{0.236} & 0.022 & 0.207 & 0.007
& 0.004 & 0.005 & 0.006 & 0.005 \\

& \texttt{llama-8B}
& 0.015 & 0.003 & 0.005 & 0.004 & 0.006
& 0.009 & 0.006 & 0.006 & 0.021
& 0.032 & 0.018 & 0.021 & 0.007
& 0.009 & 0.139 & \textbf{0.152} & 0.004
& 0.059 & 0.012 & 0.007 & 0.011
& 0.004 & 0.007 & 0.011 & 0.011 \\

& \texttt{mistral-7B}
& 0.003 & 0.008 & 0.002 & 0.000 & 0.010
& 0.040 & 0.008 & 0.005 & 0.005
& 0.029 & 0.207 & 0.174 & \textbf{0.244}
& 0.012 & 0.008 & 0.012 & 0.006
& 0.018 & 0.001 & 0.010 & 0.006
& 0.007 & 0.006 & 0.007 & 0.003 \\

& Avg
& 0.008 & 0.006 & 0.005 & 0.005 & 0.007
& 0.058 & 0.035 & 0.005 & 0.011
& 0.049 & 0.059 & 0.051 & 0.066
& 0.009 & 0.040 & 0.045 & 0.006
& \textbf{0.080} & 0.010 & 0.058 & 0.008
& 0.006 & 0.006 & 0.008 & 0.007 \\
\midrule

e-\textsc{snli} & 4o-mini
& 0.013 & 0.011 & 0.010 & 0.004 & 0.001
& 0.022 & 0.014 & 0.016 & 0.027
& 0.015 & 0.014 & 0.018 & 0.013
& 0.014 & 0.011 & 0.019 & 0.008
& 0.021 & 0.012 & \textbf{0.038} & 0.012
& 0.018 & 0.008 & 0.011 & 0.009 \\

& \texttt{llama-70B}
& 0.014 & 0.012 & 0.013 & 0.009 & 0.015
& 0.027 & 0.018 & 0.026 & \textbf{0.130}
& 0.010 & 0.014 & 0.012 & 0.003
& 0.010 & 0.010 & 0.007 & 0.013
& 0.007 & 0.003 & 0.007 & 0.012
& 0.004 & 0.010 & 0.010 & 0.007 \\

& \texttt{llama-8B}
& 0.017 & \textbf{0.046} & 0.010 & 0.011 & 0.015
& 0.029 & 0.025 & 0.011 & 0.041
& 0.043 & 0.014 & 0.026 & 0.029
& 0.021 & 0.009 & 0.011 & 0.012
& 0.021 & 0.010 & 0.017 & 0.014
& 0.007 & 0.020 & 0.022 & 0.022 \\

& \texttt{mistral-7B}
& 0.008 & 0.017 & 0.003 & 0.003 & 0.003
& 0.070 & \textbf{0.126} & 0.009 & 0.010
& 0.033 & 0.000 & 0.007 & 0.010
& 0.004 & 0.006 & 0.005 & 0.008
& 0.011 & 0.005 & 0.006 & 0.007
& 0.005 & 0.005 & 0.007 & 0.014 \\

& Avg
& 0.013 & 0.022 & 0.009 & 0.007 & 0.009
& 0.037 & 0.046 & 0.016 & \textbf{0.052}
& 0.025 & 0.011 & 0.016 & 0.014
& 0.012 & 0.009 & 0.011 & 0.010
& 0.015 & 0.008 & 0.017 & 0.011
& 0.009 & 0.011 & 0.013 & 0.013 \\
\midrule

\textsc{snarks} & 4o-mini
& 0.029 & 0.046 & 0.010 & 0.017 & 0.020
& 0.033 & -- & 0.017 & 0.021
& 0.060 & -- & 0.015 & 0.008
& \textbf{0.120} & -- & 0.012 & 0.021
& 0.064 & -- & 0.029 & 0.014
& 0.025 & -- & 0.011 & 0.019 \\

& \texttt{llama-70B}
& 0.025 & 0.027 & 0.017 & 0.013 & 0.010
& 0.075 & -- & 0.015 & 0.020
& \textbf{0.078} & -- & 0.008 & 0.007
& 0.022 & -- & 0.003 & 0.005
& 0.012 & -- & 0.009 & 0.005
& 0.009 & -- & 0.003 & 0.005 \\

& \texttt{llama-8B}
& 0.017 & 0.014 & 0.025 & 0.012 & 0.024
& 0.014 & -- & 0.017 & 0.041
& \textbf{0.120} & -- & 0.018 & 0.009
& 0.064 & -- & 0.007 & 0.006
& 0.021 & -- & 0.013 & 0.012
& 0.010 & -- & 0.009 & 0.010 \\

& \texttt{mistral-7B}
& 0.005 & 0.016 & 0.007 & 0.000 & 0.018
& 0.022 & -- & 0.008 & 0.003
& 0.013 & -- & 0.005 & 0.005
& 0.005 & -- & 0.003 & 0.005
& 0.007 & -- & 0.003 & 0.005
& 0.000 & -- & 0.011 & 0.007 \\

& Avg
& 0.019 & 0.026 & 0.015 & 0.011 & 0.018
& 0.036 & -- & 0.014 & 0.021
& \textbf{0.068} & -- & 0.012 & 0.007
& 0.053 & -- & 0.006 & 0.009
& 0.026 & -- & 0.014 & 0.009
& 0.011 & -- & 0.009 & 0.010 \\
\midrule

\textsc{boolean} & 4o-mini
& 0.020 & 0.017 & \textbf{0.019} & 0.009 & 0.004
& 0.012 & -- & 0.007 & 0.013
& 0.017 & -- & 0.007 & 0.008
& 0.015 & -- & 0.005 & 0.008
& 0.017 & -- & 0.005 & 0.009
& 0.003 & -- & 0.007 & 0.007 \\

& \texttt{llama-70B}
& 0.023 & 0.006 & 0.012 & 0.011 & 0.005
& 0.090 & -- & 0.012 & 0.005
& 0.081 & -- & 0.002 & 0.007
& 0.016 & -- & 0.004 & 0.003
& \textbf{0.168} & -- & 0.004 & 0.003
& 0.003 & -- & 0.002 & 0.004 \\

& \texttt{llama-8B}
& 0.022 & 0.041 & 0.005 & 0.004 & 0.016
& \textbf{0.051} & -- & 0.012 & 0.031
& 0.017 & -- & 0.019 & 0.005
& 0.021 & -- & 0.003 & 0.007
& 0.006 & -- & 0.008 & 0.024
& 0.015 & -- & 0.007 & 0.011 \\

& \texttt{mistral-7B}
& 0.004 & 0.025 & 0.002 & 0.000 & 0.021
& \textbf{0.057} & -- & 0.005 & 0.004
& \textbf{0.057} & -- & 0.005 & 0.002
& 0.006 & -- & 0.004 & 0.006
& 0.008 & -- & 0.004 & 0.007
& 0.005 & -- & 0.005 & 0.019 \\

& Avg
& 0.017 & 0.022 & 0.010 & 0.006 & 0.012
& \textbf{0.053} & -- & 0.009 & 0.013
& 0.044 & -- & 0.008 & 0.006
& 0.015 & -- & 0.004 & 0.006
& 0.050 & -- & 0.005 & 0.011
& 0.007 & -- & 0.005 & 0.010 \\
\midrule

\textsc{cj} & 4o-mini
& 0.015 & 0.025 & 0.015 & 0.014 & 0.012
& 0.014 & -- & 0.026 & 0.026
& 0.019 & -- & 0.016 & 0.026
& 0.013 & -- & 0.009 & 0.014
& 0.029 & -- & 0.025 & 0.047
& 0.017 & -- & 0.016 & 0.029 \\

& \texttt{llama-70B}
& 0.004 & 0.028 & 0.007 & 0.007 & 0.016
& 0.090 & -- & 0.048 & 0.021
& 0.065 & -- & 0.002 & 0.007
& 0.025 & -- & 0.019 & 0.006
& 0.058 & -- & 0.009 & 0.010
& 0.007 & -- & 0.007 & 0.008 \\

& \texttt{llama-8B}
& 0.009 & 0.017 & 0.022 & 0.010 & 0.034
& 0.018 & -- & 0.017 & 0.034
& 0.022 & -- & 0.013 & 0.016
& 0.013 & -- & 0.006 & 0.009
& 0.022 & -- & 0.013 & \textbf{0.050}
& 0.024 & -- & 0.012 & 0.008 \\

& \texttt{mistral-7B}
& 0.005 & 0.030 & 0.002 & 0.000 & 0.007
& 0.010 & -- & 0.003 & 0.002
& 0.019 & -- & 0.004 & 0.003
& 0.007 & -- & 0.003 & 0.008
& 0.016 & -- & 0.000 & 0.005
& 0.005 & -- & 0.007 & 0.005 \\

& Avg
& 0.008 & 0.025 & 0.012 & 0.008 & 0.017
& 0.033 & -- & 0.024 & 0.021
& 0.031 & -- & 0.009 & 0.013
& 0.015 & -- & 0.009 & 0.009
& 0.031 & -- & 0.012 & 0.028
& 0.013 & -- & 0.011 & 0.013 \\
\midrule

\textsc{gsm8k} & 4o-mini
& 0.009 & 0.014 & 0.012 & 0.014 & 0.011
& 0.007 & -- & \textbf{0.022} & 0.012
& 0.009 & -- & 0.015 & 0.013
& 0.010 & -- & 0.011 & 0.012
& 0.021 & -- & 0.013 & 0.013
& 0.000 & -- & 0.015 & 0.006  \\

& \texttt{llama-70B}
& 0.010 & 0.020 & 0.012 & 0.015 & 0.004
& \textbf{0.083} & -- & 0.005 & 0.005
& 0.056 & -- & 0.012 & 0.009
& 0.017 & -- & 0.018 & 0.015
& 0.027 & -- & 0.009 & 0.007 
& 0.029 & -- & 0.009 & 0.009  \\

& \texttt{llama-8B}
& 0.004 & 0.014 & 0.009 & 0.007 & 0.015
& 0.023 & -- & 0.016 & 0.013
& \textbf{0.140} & -- & 0.007 & 0.014
& 0.012 & -- & 0.009 & 0.012
& 0.017 & -- & 0.007 & 0.015
& 0.017 & -- & 0.010 & 0.013  \\

& \texttt{mistral-7B}
& 0.003 & \textbf{0.027} & 0.006 & 0.005 & 0.013
& 0.013 & -- & 0.007 & 0.008
& \textbf{0.027} & -- & 0.020 & 0.016
& 0.004 & -- & 0.011 & 0.010
& 0.009 & -- & 0.008 & 0.005
& 0.012 & -- & 0.020 & 0.004   \\

& Avg
& 0.007 & 0.019 & 0.010 & 0.010 & 0.011
& 0.032 & -- & 0.013 & 0.010
& \textbf{0.061} & -- & 0.014 & 0.013
& 0.011 & -- & 0.012 & 0.012
& 0.019 & -- & 0.009 & 0.010
& 0.015 & -- & 0.014 & 0.008 \\
\midrule

All datasets & Avg
& 0.012 & 0.020 & 0.010 & 0.008 & 0.012
& 0.042 & 0.041 & 0.014 & 0.021
& \textbf{0.046} & 0.035 & 0.018 & 0.020
& 0.019 & 0.025 & 0.015 & 0.009
& 0.037 & 0.009 & 0.019 & 0.013
& 0.010 & 0.009 & 0.010 & 0.010 \\

\bottomrule
\end{tabular}
}
\label{tab:big-table-error-five-runs-std-dev}
\end{table*}

\begin{figure*}[htbp]
    \centering
     \includegraphics[width=\linewidth]{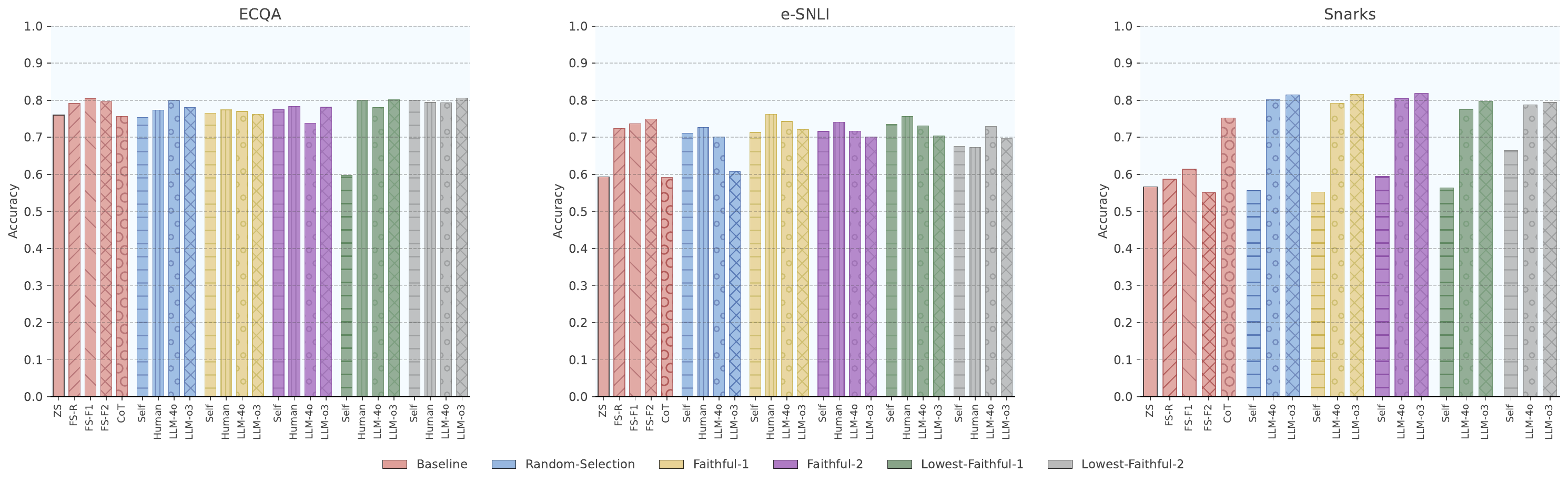}
     \includegraphics[width=\linewidth]{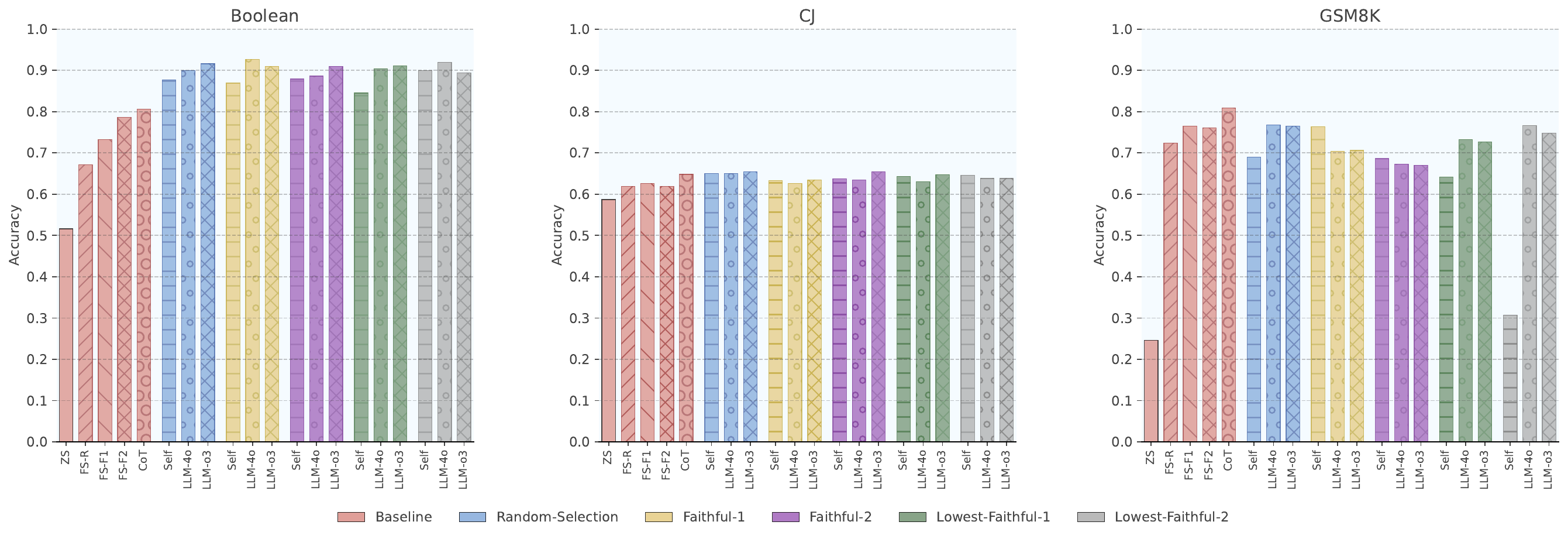}
    \caption{Accuracy values across explanation selection strategies: \textcolor{red}{Baseline}, \textcolor{blue}{Random Selection}, \textcolor{yellow!80!black}{ Faithful Selection-\(fm1\)}, \textcolor{purple}{Faithful Selection-\(fm2\)}, \textcolor{green!60!black}{Lowest-Faithful-\(fm1\) Selection}, and \textcolor{gray!80!black}{Lowest-Faithful-\(fm2\) Selection}; \textbf{averaged over the four considered models} and evaluated on six datasets: \textsc{ecqa}, e-\textsc{snli}, \textsc{snarks}, \textsc{boolean}, Causal Judgment, and \textsc{gsm8k}. Baselines include Zero-shot, Few-shot with random or faithful inputs and outputs only (no explanations), and CoT. Other strategies involve Self-NLEs, Human-provided, or LLM-NLEs (from \texttt{4o-mini} and \texttt{o3-mini}).}
    \label{fig:all_setups_comparison_plots}
\end{figure*}

\begin{figure*}[htbp]
\centering

\begin{subfigure}{\linewidth}
\includegraphics[width=0.95\linewidth]{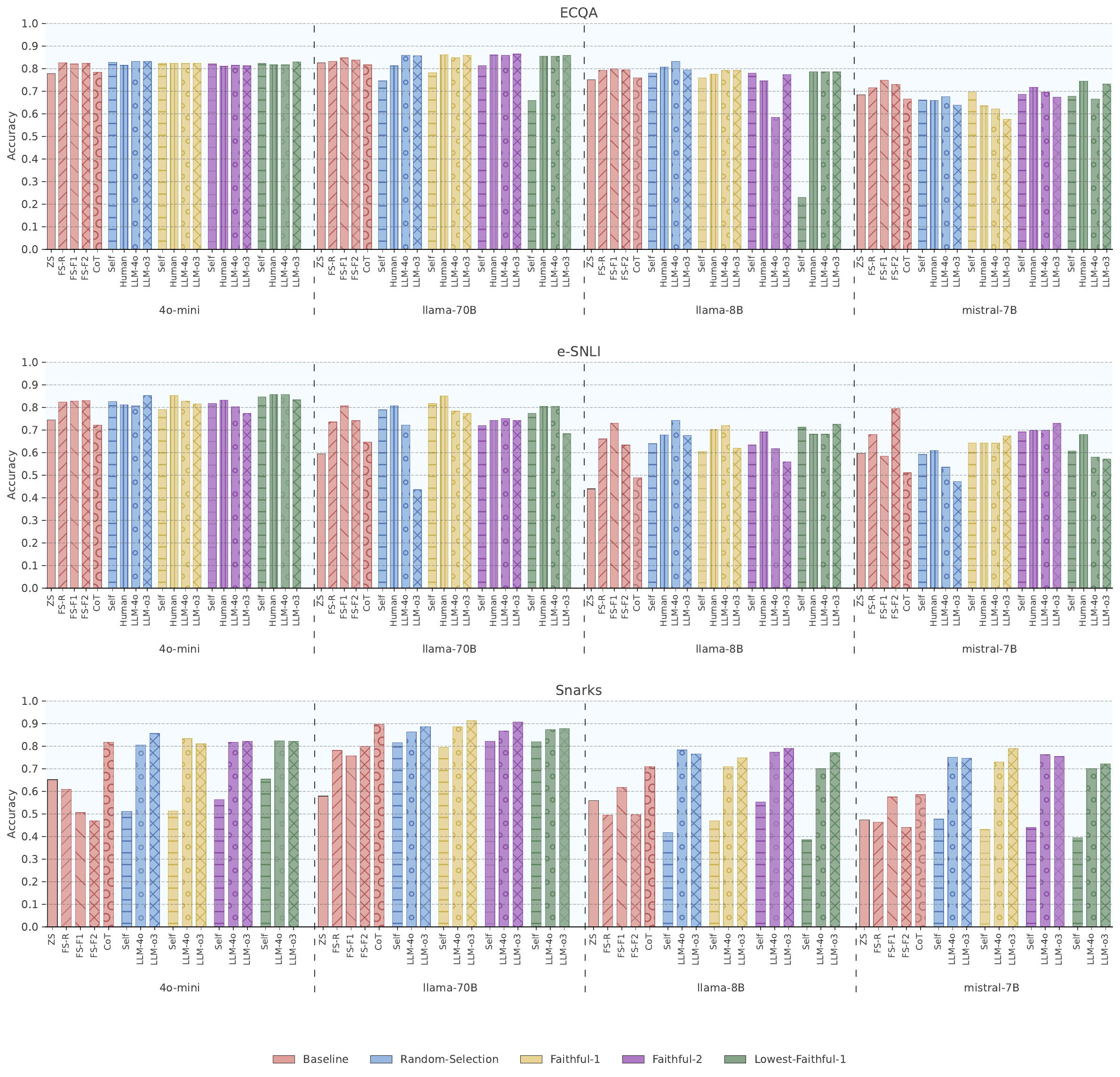}
\caption{\textsc{ecqa}, e-\textsc{snli} \& \textsc{snarks}}
\label{fig:all-values-plots}
\end{subfigure}

\end{figure*}

\begin{figure*}[htbp]
\ContinuedFloat
\centering
\begin{subfigure}{\linewidth}
\includegraphics[width=0.95\linewidth]{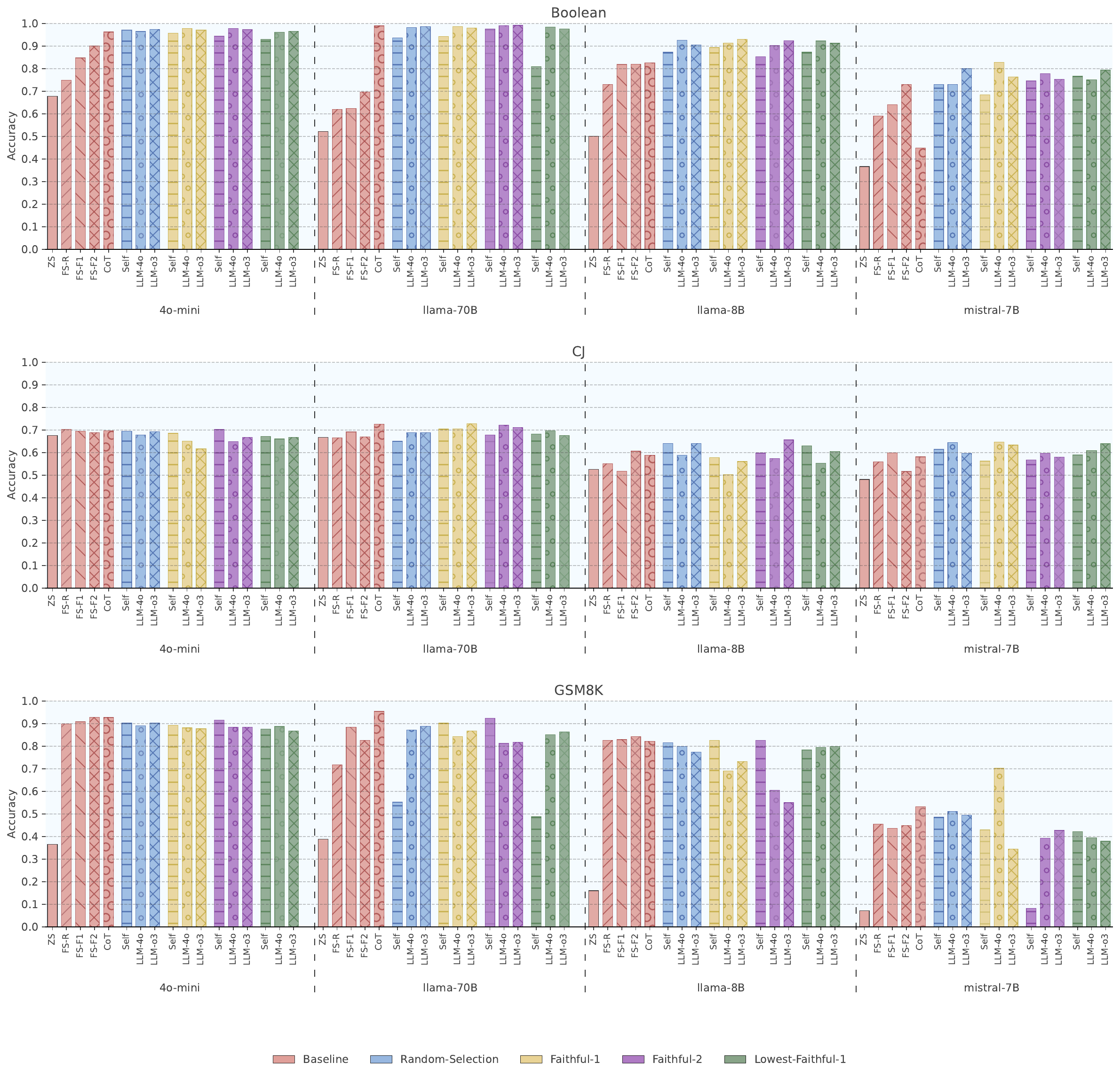}
\caption{\textsc{boolean}, Causal Judgment \& \textsc{gsm8k}}
\label{fig:all-values-plots2}
\end{subfigure}

\caption{Model-specific accuracy results across explanation selection strategies, Baseline (red), Random Selection (blue), Faithful Selection 1 (yellow), Faithful Selection 2 (purple), and Lowest-Faithful Selection (green) evaluated using \texttt{4o-mini}(left), \texttt{Llama-70B} (center-left), \texttt{Llama-8B} (center-right), and \texttt{Mistral-7B} (right). Baselines include Zero-shot, Few-shot with random or faithful inputs and outputs only (no explanations), and CoT. Other strategies involve Self-generated, Human-provided, or LLM-generated explanations (from \texttt{4o-mini} and \texttt{o3-mini}).}
\end{figure*}

\FloatBarrier

\section{Examples of Faithful-Random and OOD Model Outputs} \label{app:random-adv-outputs}

\textit{(Figures~\ref{app:prompt_structure_input_output_ll-explanation_4omini-bool}--\ref{app:fig-adv-setting-example-snarks-bool-correct-llama70} below)}

\definecolor{verylightgreen}{RGB}{180, 240, 180}

\begin{figure*}[ht]
  \centering
  {\small
  \begin{tcolorbox}[
    colback=white,
    colframe=verylightgreen!65!black,
    title=\textbf{LLM Prompt Structure, Input and Output Example},
    boxrule=0.5pt,
    arc=2pt,
    left=4pt,
    right=4pt,
    top=4pt,
    bottom=4pt,
    width=\textwidth,
    enhanced,
    breakable
  ]
    \textbf{Instruction to LLM}\smallskip
    \begin{lstlisting}[
   basicstyle=\scriptsize\ttfamily,
   breaklines=true,
   numbers=none
    ]
(user) Evaluate the following boolean expression as (1) if True and (0) if False. 
Generate a concise explanation and the right single answer. 
Strictly follow this template: Explanation: <explanation>. Answer: 0
    \end{lstlisting}

    \textbf{Few‐Shot Examples}\smallskip
    \begin{lstlisting}[
   basicstyle=\scriptsize\ttfamily,
   breaklines=true,
   numbers=none
    ]
(user) True or False and not True and False is
(assistant) Explanation: The expression evaluates to True because "True and not False" simplifies to "True and True," which is True.
Then, "not not True" simplifies to True as well. Finally, combining these results, we have "True and True," which is True.. The answer is (1).

(user) not not ( False and not False ) is
(assistant) Explanation: The expression evaluates to True because the "or" operator returns True if at least one of its operands is True. In 
this case, the expression contains the values "False," "True," and the result of "False and not False." The "not False" evaluates to True, 
making the entire expression True due to the presence of at least one True value in the "or" operations.. The answer is (0).

(user) not not not ( True and False ) is
(assistant) Explanation: The expression "not ( ( False and True ) )" evaluates as follows: "False and True" results in False, since both conditions must be true for the AND operation to yield True. Then, applying the NOT operator to False gives us True. Therefore, the final result of the expression is True.. The answer is (1).

(user) True and not not False and not False is
(assistant) Explanation: The expression simplifies as follows: "not not False" evaluates to "False," and "not False" evaluates to "True." Therefore, the expression becomes "False or True or True." Since at least one of the values is "True," the final result is "True.". The answer is (0)

(user) True or False and not ( True ) is
(assistant) Explanation: The expression evaluates to False because the first part, "not True," becomes False. When you combine False with any other value using "and," the result is always False. Therefore, the entire expression is False.. The answer is (1).

(user) not not ( not True ) or False is
(assistant) Explanation: The expression evaluates as follows:
1. The "not False" part becomes True.
2. Then the expression simplifies to False or (True and False).
3. Since "True and False" is False, we have False or False.
4. Finally, False or False results in False.
Therefore, the overall expression evaluates to False.. The answer is (0).

(user) not ( True ) and ( True ) is
    \end{lstlisting}

    \textbf{Model Output and Explanation}\smallskip
    \begin{lstlisting}[
   basicstyle=\scriptsize\ttfamily,
   breaklines=true,
   numbers=none
    ] 
Explanation: The expression "not (True)" evaluates to False. Therefore, the expression simplifies to "False and True." In a logical AND operation, if one of the operands is False, the result is False. Thus, the overall expression evaluates to False. The answer is (0).
    \end{lstlisting}
  \end{tcolorbox}
  }
  \caption{Illustration of an in‐context learning prompt using \textbf{Random Rationales (Setting 4)} on the \textbf{Boolean dataset} with \textbf{4o-mini} as the evaluation model. The prompt includes an instruction followed by six few‐shot examples (top), with explanations generated by 4o-mini. The final input (bottom) is the test question. The model’s prediction is \textbf{correct}. This example reflects an error case under the \textbf{LLM‐explanation strategy.}}
  \label{app:prompt_structure_input_output_ll-explanation_4omini-bool}
\end{figure*}

\begin{figure*}[ht]
  \centering
  {\small
  \begin{tcolorbox}[
    colback=white,
    colframe=verylightgreen!65!black,
    title=\textbf{LLM Prompt Structure, Input and Output Example},
    boxrule=0.5pt,
    arc=2pt,
    left=4pt,
    right=4pt,
    top=4pt,
    bottom=4pt,
    width=\textwidth,
    enhanced,
    breakable
  ]
    \textbf{Instruction to LLM}\smallskip
    \begin{lstlisting}[
   basicstyle=\scriptsize\ttfamily,
   breaklines=true,
   numbers=none
    ]
(user) Evaluate the following boolean expression as (1) if True and (0) if False. 
Generate a concise explanation and the right single answer. 
Strictly follow this template: Explanation: <explanation>. Answer: 0
    \end{lstlisting}

    \textbf{Few‐Shot Examples}\smallskip
    \begin{lstlisting}[
   basicstyle=\scriptsize\ttfamily,
   breaklines=true,
   numbers=none
    ]
(user) True or False and not True and False is
(assistant) Explanation: The expression is evaluated by first applying the "not" operators: "not False" becomes True, and "not not True" simplifies to True. Then, using the "and" operator on True, True, and True results in True, which is why the final answer is True.. The answer is (1).

(user) not not ( False and not False ) is
(assistant) Explanation: First, 'not False' evaluates to True. Then, applying 'and', the subexpression 'False and True' becomes False. Finally, the 'or' operations simplify the overall expression to 'False or True or False', which evaluates to True.. The answer is (0).

(user) not not not ( True and False ) is
(assistant) Explanation: False and True evaluates to False because the 'and' operator requires both values to be True. When you apply 'not' to False, it inverts to True, which is why the final answer is True.. The answer is (1).

(user) True and not not False and not False is
(assistant) Explanation: Double negation turns not not False into False while not False becomes True. The expression then simplifies to False or True or True, and because an OR expression is true if any part is true, the overall result is True.. The answer is (0).

(user) True or False and not ( True ) is
(assistant) Explanation: The expression first negates True, turning it into False, and then the "and" operation between two False values results in False since "and" only returns True if both operands are True.. The answer is (1).

(user) not not ( not True ) or False is
(assistant) Explanation: First, the inner expression (not False) evaluates to True. Then, True and False evaluates to False because both operands must be True for an AND operation to result in True. Finally, False or False evaluates to False, which is why the overall expression is False.. The answer is (0).

(user) not ( True ) and ( True ) is
    \end{lstlisting}

    \textbf{Model Output and Explanation}\smallskip
    \begin{lstlisting}[
   basicstyle=\scriptsize\ttfamily,
   breaklines=true,
   numbers=none
    ] 
Explanation: The expression starts with 'not (True)', which evaluates to False. Then, the AND operation between this False and True results in False, since AND requires both operands to be True to return True.. The answer is (0).    \end{lstlisting}
  \end{tcolorbox}
  }
  \caption{Illustration of an in‐context learning prompt using \textbf{Random Rationales (Setting 4)} on the \textbf{Boolean dataset} with \textbf{4o-mini} as the evaluation model. The prompt includes an instruction followed by six few‐shot examples (top), with explanations generated by \textbf{o3-mini}. The final input (bottom) is the test question. The model’s prediction is \textbf{correct}. This example reflects an error case under the \textbf{LLM‐explanation strategy}.
  }
  \label{app:prompt_structure_input_output_ll-explanation_o3mini-bool}
\end{figure*}

\begin{figure*}[ht]
  \centering

    \begin{adjustbox}{max width=\textwidth, max height=0.8\textheight, center}
    \begin{minipage}{\textwidth}

  {\small
  \begin{tcolorbox}[
    colback=white,
    colframe=verylightgreen!65!black,
    title=\textbf{LLM Prompt Structure, Input and Output Example},
    boxrule=0.5pt,
    arc=2pt,
    left=4pt,
    right=4pt,
    top=4pt,
    bottom=4pt,
    width=\textwidth,
    enhanced,
  ]
    \textbf{Instruction to LLM}\smallskip
    \begin{lstlisting}[
   basicstyle=\tiny\ttfamily,
   breaklines=true,
   numbers=none
    ]
(user) Evaluate the following boolean expression and generate a concise 3-step explanation, with exactly one sentence per step. The answer is (1) if True, (0) if False. Strictly follow this answer format with the label in parentheses: The answer is (0), 3-step explanation: step1, step2, ..., stepn.
    \end{lstlisting}

    \textbf{Few‐Shot Examples}\smallskip
    \begin{lstlisting}[
   basicstyle=\scriptsize\ttfamily,
   breaklines=true,
   numbers=none
    ]
(user) False or ( not False ) and False is
(assistant) Explanation:
Step 1: The expression 'not True' evaluates to False because the 'not' operator inverts the boolean value.
Step 2: The expression 'False and True and True' evaluates to False because any 'and' operation with a False value results in False.
Step 3: The expression 'False or False' evaluates to False because the 'or' operator returns True only if at least one of the operands is True.. Answer: (0).

(user) False or ( False and not False ) is
(assistant) Explanation:
Step 1: The expression 'not ( True or True )' evaluates to 'not True', which is False.
Step 2: The expression 'not True' is False, so the 'and' operation with False results in False.
Step 3: Since the final result of the expression is False, the answer is (0).. Answer: (0).

(user) True and not not False and not False is
(assistant) Explanation:
Step 1: The expression inside the parentheses, "False and True", evaluates to False because in a boolean AND operation, both conditions must be True for the result to be True.
Step 2: The NOT operator is applied to the result of the expression inside the parentheses, which is False, so the NOT operator flips the result to True.
Step 3: The final result of the expression is True, which corresponds to the value (1).. Answer: (0).

(user) False and not ( True ) or True is
(assistant) Explanation:
Step 1: The expression starts with "True and not True", which evaluates to False because "not True" is False and "and" requires both operands to be True.
Step 2: Since the first part of the expression is False, the "or" operator will evaluate the rest of the expression, which is "False or True".
Step 3: The "or" operator requires only one operand to be True, so "False or True" evaluates to True.. Answer: (1).

(user) True and ( not True or False ) is
(assistant) Explanation:
Step 1: The expression 'not False' evaluates to True because the 'not' operator negates the value of False, making it True.
Step 2: The expression 'True and False' evaluates to False because the 'and' operator requires both operands to be True for the result to be True.
Step 3: The expression 'False or False or False' evaluates to False because the 'or' operator requires at least one operand to be True for the result to be True, but all operands are False.. Answer: (0).

(user) not not False and not not not False is
(assistant) Explanation:
Step 1: The expression 'False and not False' evaluates to False because 'not False' is True, but 'False and True' is False.
Step 2: The expression 'False or False' evaluates to False because both conditions are False.
Step 3: The final result is False because the 'or' operation between False and False is False.. Answer: (0).

(user) not ( True ) and ( True ) is
    \end{lstlisting}

    \textbf{Model Output and Explanation}\smallskip
    \begin{lstlisting}[
   basicstyle=\scriptsize\ttfamily,
   breaklines=true,
   numbers=none
    ] 
Explanation:
Step 1: The expression 'not True' evaluates to False because the 'not' operator negates the value of True.
Step 2: The expression 'True and True' evaluates to True because both operands are True.
Step 3: The final result is False because 'False and True' is False.. Answer: (0)  \end{lstlisting} \end{tcolorbox}
  }

        \end{minipage}
        \end{adjustbox}
  
  \caption{Illustration of an in‐context learning prompt using \textbf{Random Rationales (Setting 4)} on the \textbf{Boolean dataset} with \textbf{\texttt{Llama-8B}} as the evaluation model. The prompt includes an instruction followed by six few‐shot examples (top), with explanations generated by the evaluation model itself. The final input (bottom) is the test question. The model’s prediction is \textbf{correct}. This example reflects an error case under the \textbf{Self‐explanation strategy.}}
  \label{app:prompt_structure_input_output_ll-explanation_4omini-bool-llama8b}
\end{figure*}

\begin{figure*}[ht]
  \centering

  {\small
  \begin{tcolorbox}[
    colback=white,
    colframe=verylightgreen!65!black,
    title=\textbf{LLM Prompt Structure, Input and Output Example},
    boxrule=0.5pt,
    arc=2pt,
    left=4pt,
    right=4pt,
    top=4pt,
    bottom=4pt,
    width=\textwidth,
    enhanced,
  ]
    \textbf{Instruction to LLM}\smallskip
    \begin{lstlisting}[
   basicstyle=\scriptsize\ttfamily,
   breaklines=true,
   numbers=none
    ]
(user) Choose the single right answer and generate a concise 3-step explanation, with only one sentence per step. Strictly follow this format: Answer: (a). 3-step explanation: step1, step2, ..., stepn.
    \end{lstlisting}

    \textbf{Few‐Shot Examples}\smallskip
    \begin{lstlisting}[
   basicstyle=\tiny\ttfamily,
   breaklines=true,
   numbers=none
    ]
(user) What type of home is most likely to have an attic filled with historical items? (a) top of home (b) older houses (c) victorian house (d) own home (e) material
(assistant) Explanation:
step1: Eating too much dinner can put excessive pressure on the stomach, leading to discomfort and potential pain.
step2: The stomach's digestive system is overwhelmed by the large amount of food, causing it to work harder and leading to possible cramps.
step3: As a result, consuming a large dinner is likely to cause stomach ache, which is characterized by sharp pains or a dull ache in the abdominal area.. Answer: (c).

(user) He had done a great job with the scary decorations, they were completely what? (a) terrible (b) little (c) halloween (d) small (e) horrible
(assistant) Explanation:
step1: The question asks about the outcome of performing well on a test, which implies an academic setting.
step2: In an academic setting, a test is typically used to evaluate a student's knowledge and understanding of a subject.
step3: The result of this evaluation is often reflected in a school grade, which can be an A if the student performs exceptionally well.. Answer: (e).

(user) What could a balalaika be needed for? (a) artist junkyard (b) music store (c) singing (d) symphony (e) orchestra
(assistant) Explanation:
step1: The term "first violin" is a common designation in musical ensembles, indicating a specific role or position.
step2: This position is typically associated with a section of musicians who play the violin, often found in large musical groups.
step3: A symphony orchestra is a type of musical ensemble that often features a first violin section, making it the most relevant and correct context for the term "first violin".. Answer: (d).

(user) What is likely to have more than one level in a city? (a) carpenter's shop (b) video game (c) garage (d) total number (e) building
(assistant) Explanation:
step1: The presence of an inkwell suggests a piece of furniture used for writing or reading, which is often associated with a desk.
step2: Desks are commonly found in homes and offices, and they typically have a flat surface that can hold items like inkwells, papers, and writing utensils.
step3: Given the context of the ink spilling onto the rug, it is likely that the desk was placed on the rug, leading to the conclusion that the piece of furniture on her rug might have been a desk.. Answer: (c).

(user) Where do you buy tickets at a ticket booth for games? (a) train station (b) cathedral (c) metro station (d) fairgrounds (e) amusement park
(assistant) Explanation:
step1: A season of successful skiing implies regular physical activity and exercise, which can contribute to overall physical health.
step2: Engaging in skiing and other winter sports can help improve cardiovascular health, boost the immune system, and increase muscle strength.
step3: As a result, a person who has a successful skiing season is likely to experience improvements in their physical condition, leading to a healthier body.. Answer: (d).

(user) What building has a lot of glue bottles? (a) art room (b) police station (c) classroom (d) desk drawer (e) school
(assistant) Explanation:
step1: When skin is pinched in something, the immediate effect is the application of external force on the skin.
step2: This external force compresses the skin and the underlying tissues, stimulating the nerve endings that detect pressure and touch.
step3: As a result, the brain interprets the signals from these nerve endings as a sensation of pressure, making the skin feel pressed or squeezed.. Answer: (e).

(user) The sanctions against the school were a punishing blow, and they seemed to what the efforts the school had made to change? (a) ignore (b) enforce (c) authoritarian (d) yell at (e) avoid

    \end{lstlisting}

    \textbf{Model Output and Explanation}\smallskip
    \begin{lstlisting}[
   basicstyle=\scriptsize\ttfamily,
   breaklines=true,
   numbers=none
    ] 
Answer: (a)
3-step explanation:
step1: The question describes a situation where sanctions are imposed on a school, implying a negative consequence or penalty.
step2: The phrase ""punishing blow"" suggests that the sanctions have a significant and detrimental impact on the school's efforts to change or improve.
step3: The word ""ignore"" fits the context, implying that the sanctions overshadow or nullify the school's attempts to make positive changes, rendering their efforts ineffective.
\end{lstlisting} \end{tcolorbox}
  }

  \caption{Illustration of an in‐context learning prompt using \textbf{Random Rationales (Setting 4)} on the \textbf{ECQA dataset} with \textbf{\texttt{Llama-70B}} as the evaluation model. The prompt includes an instruction followed by six few‐shot examples (top), with explanations generated by the evaluation model itself. The final input (bottom) is the test question. The model’s prediction is \textbf{correct}. This example reflects an error case under the \textbf{Self‐explanation strategy.}}
  \label{app:prompt_structure_input_output_ll-explanation_4omini-bool-llama70b}
\end{figure*}

\begin{figure*}[ht]
  \centering

 \begin{adjustbox}{max width=\textwidth, max height=0.8\textheight, center}
    \begin{minipage}{\textwidth}
    
  {\small
  \begin{tcolorbox}[
    colback=white,
    colframe=verylightgreen!65!black,
    title=\textbf{LLM Prompt Structure, Input and Output Example},
    boxrule=0.5pt,
    arc=2pt,
    left=4pt,
    right=4pt,
    top=4pt,
    bottom=4pt,
    width=\textwidth,
    enhanced,
  ]
    \textbf{Instruction to LLM}\smallskip
    \begin{lstlisting}[
   basicstyle=\scriptsize\ttfamily,
   breaklines=true,
   numbers=none
    ]
(user) Evaluate the following boolean expression as (1) if True and (0) if False. Generate a concise explanation and the right single answer. Strictly follow this template: Explanation: <explanation>. Answer: 0
    \end{lstlisting}

    \textbf{Few-shot Examples}
    \\[0.5em]
    \begin{lstlisting}[
   basicstyle=\scriptsize\ttfamily,
   breaklines=true,
   numbers=none
    ]
(user) False or ( not False ) and False is
(assistant) Explanation: The sarcastic element in (a) comes from the mocking tone created by comparing the league's supposedly "consistent" punishments to the inconsistency of politics. The speaker uses irony--suggesting that the idea of consistency is as absurd as expecting politics to be fair--to undermine the seriousness of the claim. This sarcastic contrast is what makes (a) the correct answer. The answer is (0).

(user) False or ( False and not False ) is
(assistant) Explanation: Answer (b) is sarcastic because it praises using ad hominem attacks as if they would genuinely strengthen an argument, when in fact such attacks are generally seen as a weak, fallacious tactic. The irony in stating that these attacks "really help your case" signals that the speaker doesn't truly believe in their effectiveness, creating a sarcastic tone.. The answer is (0).

(user) True and not not False and not False is
(assistant) Explanation: Option (a) is sarcastic because it uses an exaggerated negative phrase ("huge waste") to imply the opposite of what is literally stated. The irony in dismissing the value of research interests as wasteful reveals a sarcastic tone.. The answer is (0).

(user) False and not ( True ) or True is
(assistant) Explanation: Option (a) is sarcastic because it intentionally contradicts itself by saying they haven't spoken in years while calling the person "my bestie," implying irony. In contrast, option (b) has no contradictory elements. The answer is (1).

(user) True and ( not True or False ) is
(assistant) Explanation: Option (a) is sarcastic because it uses exaggerated praise for a harmful attitude--instead of genuinely approving of generalizing and demeaning women, it ironically comments on such behavior to highlight its absurdity. This contrasts with (b), which states the consequence straightforwardly without irony. The answer is (0).


(user) not not False and not not not False is
(assistant) Explanation: Statement (a) is sarcastic because it ironically claims that being ten times slower in design will lead Honda to market dominance, which undermines the logic that slow progress is beneficial. The exaggeration highlights the absurdity of using sluggish performance as an advantage, making the tone sarcastic.. The answer is (0).

(user) not False or True and False and False is

    \end{lstlisting}

    \textbf{Model Output and Explanation}\smallskip
    \begin{lstlisting}[
   basicstyle=\scriptsize\ttfamily,
   breaklines=true,
   numbers=none
    ] 
Explanation: The expression is evaluated as follows:
- not False is True
- True or True is True
- True and False is False
- False and False is False

So, the final result is False. 

Answer: 0
\end{lstlisting} \end{tcolorbox}
  }

    \end{minipage}
    \end{adjustbox}
  
  \caption{Illustration of an in‐context learning prompt using \textbf{OOD Rationales (Setting 5)} on the \textbf{Boolean dataset} with \textbf{\texttt{Llama-8B}} as the evaluation model. \textbf{Snarks dataset} LLM-explanations generated by \texttt{o3-mini} are used as OOD-random explanations from a different domain. The prompt includes an instruction followed by six few‐shot examples (top). The final input (bottom) is the test question. The model’s prediction is \textbf{wrong}. This example reflects an error case under the \textbf{LLM‐explanation strategy.}}
  \label{app:prompt_structure_input_output_ll-explanation_4omini-bool-wrong}
\end{figure*}

\begin{figure*}[ht]
  \centering
  {\small
  \begin{tcolorbox}[
    colback=white,
    colframe=verylightgreen!65!black,
    title=\textbf{LLM Prompt Structure, Input and Output Example},
    boxrule=0.5pt,
    arc=2pt,
    left=4pt,
    right=4pt,
    top=4pt,
    bottom=4pt,
    width=\textwidth,
    enhanced,
  ]
    \textbf{Instruction to LLM}\smallskip
    \begin{lstlisting}[
   basicstyle=\scriptsize\ttfamily,
   breaklines=true,
   numbers=none
    ]
(user) Evaluate the following boolean expression as (1) if True and (0) if False. Generate a concise explanation and the right single answer. Strictly follow this template: Explanation: <explanation>. Answer: 0
    \end{lstlisting}

    \textbf{Few-shot Examples}
    \\[0.5em]
    \begin{lstlisting}[
   basicstyle=\scriptsize\ttfamily,
   breaklines=true,
   numbers=none
    ]
(user) False or ( not False ) and False is
(assistant) Explanation: The sarcastic element in (a) comes from the mocking tone created by comparing the league's supposedly "consistent" punishments to the inconsistency of politics. The speaker uses irony--suggesting that the idea of consistency is as absurd as expecting politics to be fair--to undermine the seriousness of the claim. This sarcastic contrast is what makes (a) the correct answer. The answer is (0).

(user) False or ( False and not False ) is
(assistant) Explanation: Answer (b) is sarcastic because it praises using ad hominem attacks as if they would genuinely strengthen an argument, when in fact such attacks are generally seen as a weak, fallacious tactic. The irony in stating that these attacks "really help your case" signals that the speaker doesn't truly believe in their effectiveness, creating a sarcastic tone.. The answer is (0).

(user) True and not not False and not False is
(assistant) Explanation: Option (a) is sarcastic because it uses an exaggerated negative phrase ("huge waste") to imply the opposite of what is literally stated. The irony in dismissing the value of research interests as wasteful reveals a sarcastic tone.. The answer is (0).

(user) False and not ( True ) or True is
(assistant) Explanation: Option (a) is sarcastic because it intentionally contradicts itself by saying they haven't spoken in years while calling the person "my bestie," implying irony. In contrast, option (b) has no contradictory elements. The answer is (1).

(user) True and ( not True or False ) is
(assistant) Explanation: Option (a) is sarcastic because it uses exaggerated praise for a harmful attitude--instead of genuinely approving of generalizing and demeaning women, it ironically comments on such behavior to highlight its absurdity. This contrasts with (b), which states the consequence straightforwardly without irony. The answer is (0).


(user) not not False and not not not False is
(assistant) Explanation: Statement (a) is sarcastic because it ironically claims that being ten times slower in design will lead Honda to market dominance, which undermines the logic that slow progress is beneficial. The exaggeration highlights the absurdity of using sluggish performance as an advantage, making the tone sarcastic.. The answer is (0).

(user) not ( True ) and ( True ) is

    \end{lstlisting}

    \textbf{Model Output and Explanation}\smallskip
    \begin{lstlisting}[
   basicstyle=\scriptsize\ttfamily,
   breaklines=true,
   numbers=none
    ] 
Explanation: The expression ""not (True) and (True)"" can be broken down as follows:
- ""not (True)"" is False, because the ""not"" operator negates the truth value of the expression inside the parentheses.
- ""and"" is a logical operator that requires both operands to be True for the result to be True.
- Since the first operand is False, the result of the ""and"" operation is False, regardless of the second operand.

Answer: 0
\end{lstlisting} \end{tcolorbox}
  }
  \caption{Illustration of an in‐context learning prompt using \textbf{OOD Rationales (Setting 5)} on the \textbf{Boolean dataset} with \textbf{\texttt{Llama-8B}} as the evaluation model. \textbf{Snarks dataset} LLM-explanations generated by \texttt{o3-mini} are used as OOD-random explanations from a different domain. The prompt includes an instruction followed by six few‐shot examples (top). The final input (bottom) is the test question. The model’s prediction is \textbf{correct}. This example reflects an error case under the \textbf{LLM‐explanation strategy.}}
  \label{app:prompt_structure_input_output_ll-explanation_4omini-bool-llama8b-snarks-correct}
\end{figure*}

\begin{figure*}[ht]
  \centering
  {\small
  \begin{tcolorbox}[
    colback=white,
    colframe=verylightgreen!65!black,
    title=\textbf{LLM Prompt Structure, Input and Output Example},
    boxrule=0.5pt,
    arc=2pt,
    left=4pt,
    right=4pt,
    top=4pt,
    bottom=4pt,
    width=\textwidth,
    enhanced,
    breakable
  ]
    \textbf{Instruction to LLM}\smallskip
    \begin{lstlisting}[
   basicstyle=\scriptsize\ttfamily,
   breaklines=true,
   numbers=none
    ]
(user) Choose the single right answer. Generate a concise explanation and the right single answer. Strictly follow this template: Explanation: <explanation>. Answer: (a)
    \end{lstlisting}

    \textbf{Few‐Shot Examples}\smallskip
    \begin{lstlisting}[
   basicstyle=\scriptsize\ttfamily,
   breaklines=true,
   numbers=none
    ]
(user) What type of home is most likely to have an attic filled with historical items? (a) top of home (b) older houses (c) victorian house (d) own home (e) material
(assistant) Explanation: The premise describes two older people sitting together on a bench, which suggests a level of companionship that could include holding hands.. Answer: (c)

(user) He had done a great job with the scary decorations, they were completely what? (a) terrible (b) little (c) halloween (d) small (e) horrible
(assistant) Explanation: Playing soccer on a wet field implies that the men are outdoors, making it impossible for them to be playing inside the house.. Answer: (e)

(user) What could a balalaika be needed for? (a) artist junkyard (b) music store (c) singing (d) symphony (e) orchestra
(assistant) Explanation: The female speaker is presenting to a group, indicating that she is indeed standing in front of people.. Answer: (d)

(user) What is likely to have more than one level in a city? (a) carpenter's shop (b) video game (c) garage (d) total number (e) building
(assistant) Explanation: The girl's smile suggests that she likely has straight teeth, as smiling often highlights one's dental alignment.. Answer: (c)

(user) Where do you buy tickets at a ticket booth for games? (a) train station (b) cathedral (c) metro station (d) fairgrounds (e) amusement park
(assistant) Explanation: The black poodle in the premise is a type of pet that is holding a dog toy, which aligns with the description in the hypothesis.. Answer: (d)

(user) What building has a ot of glue bottles? (a) art room (b) police station (c) classroom (d) desk drawer (e) school
(assistant) Explanation: The premise specifies that a male food worker is grilling meat, while the hypothesis states that a woman is grilling meat, indicating a difference in gender and potentially the identity of the person grilling.. Answer: (e)

(user) The sanctions against the school were a punishing blow, and they seemed to what the efforts the school had made to change? (a) ignore (b) enforce (c) authoritarian (d) yell at (e) avoid

    \end{lstlisting}

    \textbf{Model Output and Explanation}\smallskip
    \begin{lstlisting}[
   basicstyle=\scriptsize\ttfamily,
   breaklines=true,
   numbers=none
    ] 
Explanation: The context suggests that the woman is traveling, and the act of checking baggage is typically associated with air travel, indicating that she is likely heading to an airport.
Answer: (e)
\end{lstlisting} \end{tcolorbox}
  }
  \caption{Illustration of an in‐context learning prompt using \textbf{OOD Rationales (Setting 5)} on the \textbf{ECQA dataset} with \textbf{\texttt{Llama-70B}} as the evaluation model. \textbf{e-SNLI dataset} LLM-explanations generated by \texttt{4o-mini} are used as OOD-random explanations from a different domain. The prompt includes an instruction followed by six few‐shot examples (top). The final input (bottom) is the test question. The model’s prediction is \textbf{wrong}. This example reflects an error case under the \textbf{LLM‐explanation strategy.}}
    \label{app:fig-adv-setting-example-ecqa-esnli-wrong}
\end{figure*}

\begin{figure*}[ht]
  \centering
  {\small
  \begin{tcolorbox}[
    colback=white,
    colframe=verylightgreen!65!black,
    title=\textbf{LLM Prompt Structure, Input and Output Example},
    boxrule=0.5pt,
    arc=2pt,
    left=4pt,
    right=4pt,
    top=4pt,
    bottom=4pt,
    width=\textwidth,
    enhanced,
    breakable
  ]
    \textbf{Instruction to LLM}\smallskip
    \begin{lstlisting}[
   basicstyle=\scriptsize\ttfamily,
   breaklines=true,
   numbers=none
    ]
(user) Choose the single right answer. Generate a concise explanation and the right single answer. Strictly follow this template: Explanation: <explanation>. Answer: (a)
    \end{lstlisting}

    \textbf{Few‐Shot Examples}\smallskip
    \begin{lstlisting}[
   basicstyle=\scriptsize\ttfamily,
   breaklines=true,
   numbers=none
    ]
(user) What type of home is most likely to have an attic filled with historical items? (a) top of home (b) older houses (c) victorian house (d) own home (e) material
(assistant) Explanation: The premise describes two older people sitting together on a bench, which suggests a level of companionship that could include holding hands.. Answer: (c)

(user) He had done a great job with the scary decorations, they were completely what? (a) terrible (b) little (c) halloween (d) small (e) horrible
(assistant) Explanation: Playing soccer on a wet field implies that the men are outdoors, making it impossible for them to be playing inside the house.. Answer: (e)

(user) What could a balalaika be needed for? (a) artist junkyard (b) music store (c) singing (d) symphony (e) orchestra
(assistant) Explanation: The female speaker is presenting to a group, indicating that she is indeed standing in front of people.. Answer: (d)

(user) What is likely to have more than one level in a city? (a) carpenter's shop (b) video game (c) garage (d) total number (e) building
(assistant) Explanation: The girl's smile suggests that she likely has straight teeth, as smiling often highlights one's dental alignment.. Answer: (c)

(user) Where do you buy tickets at a ticket booth for games? (a) train station (b) cathedral (c) metro station (d) fairgrounds (e) amusement park
(assistant) Explanation: The black poodle in the premise is a type of pet that is holding a dog toy, which aligns with the description in the hypothesis.. Answer: (d)

(user) What building has a ot of glue bottles? (a) art room (b) police station (c) classroom (d) desk drawer (e) school
(assistant) Explanation: The premise specifies that a male food worker is grilling meat, while the hypothesis states that a woman is grilling meat, indicating a difference in gender and potentially the identity of the person grilling.. Answer: (e)

(user) What is it called when you slowly cook using a grill? (a) backyard (b) restaurant (c) crockpot (d) neighbor's house (e) barbeque

    \end{lstlisting}

    \textbf{Model Output and Explanation}\smallskip
    \begin{lstlisting}[
   basicstyle=\scriptsize\ttfamily,
   breaklines=true,
   numbers=none
    ] 
Explanation: The term ""barbeque"" refers to a method of cooking where food is slowly cooked over low heat, often using a grill, which matches the description given.
Answer: (e)
\end{lstlisting} \end{tcolorbox}
  }
    \caption{Illustration of an in‐context learning prompt using \textbf{OOD Rationales (Setting 5)} on the \textbf{Snarks dataset} with \textbf{\texttt{Llama-70B}} as the evaluation model. \textbf{Boolean dataset} LLM-explanations generated by \texttt{4o-mini} are used as OOD-random explanations from a different domain. The prompt includes an instruction followed by six few‐shot examples (top). The final input (bottom) is the test question. The model’s prediction is \textbf{correct}. This example reflects an error case under the \textbf{LLM‐explanation strategy.}}
  \label{app:fig-adv-setting-example-snarks-book-correct}
\end{figure*}

\begin{figure*}[h]
  \centering

  \begin{adjustbox}{max width=\textwidth, max height=0.80\textheight, center}
  \begin{minipage}{\textwidth}
  
  {\scriptsize
  \begin{tcolorbox}[
    colback=white,
    colframe=verylightgreen!65!black,
    title=\textbf{LLM Prompt Structure, Input and Output Example},
    boxrule=0.5pt,
    arc=2pt,
    left=4pt,
    right=4pt,
    top=4pt,
    bottom=4pt,
    width=\textwidth,
    enhanced,
  ]
    \textbf{Instruction to LLM}\smallskip
    \begin{lstlisting}[
   basicstyle=\scriptsize\ttfamily,
   breaklines=true,
   numbers=none
    ]
(user) Choose the single right answer. Generate a concise explanation and the right single answer. Strictly follow this template: Explanation: <explanation>. Answer: (a)
    \end{lstlisting}

    \textbf{Few‐Shot Examples}\smallskip
    \begin{lstlisting}[
   basicstyle=\scriptsize\ttfamily,
   breaklines=true,
   numbers=none
    ]
(user) Which statement is sarcastic? (a) Try having your landscape fill the entirety of the photo. That'd look good. (b) Try having your face fill the entirety of the photo. That'd look good.
(assistant) Explanation: The expression evaluates to False because the inner "not not True" simplifies to True. This leads to "True and True and not True," which further simplifies to "True and True and False." Since any logical operation with False results in False, the final result is False.. Answer: (b)

(user) Which statement is sarcastic? (a) You mean Knuth up arrow notation isn't unfamiliar for you? (b) You mean Knuth up arrow notation isn't normal for you?
(assistant) `Explanation: The expression evaluates as follows:`
1. `"not True" becomes False.`
2. `The expression simplifies to "True and False or False or True".`
3. `"True and False" results in False.`
4. `Now we have "False or False or True".`
5. `Evaluating this from left to right, "False or False" is False, and "False or True" is True.`
`Thus, the final result of the expression is True.. Answer: (b)`

(user) Which statement is sarcastic? (a) You mean your team did the most damage, then you just did the final blow. Amazing display of team skill. (b) You mean your team did the most damage, then you just did the final blow. Amazing display of individual skill.
(assistant) Explanation: The expression evaluates as follows: "not True" is False, and "False or True" is True. Therefore, the entire expression becomes "False and True," which is False.. Answer: (b)

(user) Which statement is sarcastic? (a) Armed revolutionaries have not been making a lot of change for middle class Americans lately. (b) Armed revolutionaries have been making a lot of change for middle class Americans lately.
(assistant) Explanation: The expression "not False" evaluates to True, and "False or False" evaluates to False. Therefore, the overall expression becomes "True and False," which results in False.. Answer: (b)

(user) Which statement is sarcastic? (a) Well, Alabama did beat the 1st ranked team. Surely that should be enough to get them in. (b) Well, Alabama did beat the 17th ranked team. Surely that should be enough to get them in.
(assistant) Explanation: The expression evaluates to True because the 'not True' part becomes False, and the 'False or' part results in False. However, the 'False and True' portion is also False, but the final 'or True' makes the entire expression True. Therefore, any expression that includes 'or True' will ultimately be True.. Answer: (b)

(user) Which statement is sarcastic? (a) God, to think of how good it could have been having a person who prepared for things ahead of time in office. (b) God, to think of how bad it could have been having a person who prepared for things ahead of time in office.
(assistant) Explanation: The expression evaluates as follows: "not True" becomes "False," so we have "False or False or False or False." Since all components are false, the overall result is "False.". Answer: (b)

(user) Which statement is sarcastic? (a) Protein supplements are way more expensive than $1 breakfast gruel. (b) Protein supplements are way more expensive than $1 breakfast burritos.

    \end{lstlisting}

    \textbf{Model Output and Explanation}\smallskip
    \begin{lstlisting}[
   basicstyle=\scriptsize\ttfamily,
   breaklines=true,
   numbers=none
    ] 
Explanation: Statement (a) is a genuine comparison, while statement (b) is sarcastic because it implies that protein supplements are overpriced by comparing them to a relatively expensive food item like breakfast burritos, rather than a cheap alternative like gruel. Answer: (b)
\end{lstlisting} 
\end{tcolorbox}
  }

    \end{minipage}
    \end{adjustbox}

  \caption{Illustration of an in‐context learning prompt using \textbf{OOD Rationales (Setting 5)} on the \textbf{Snarks dataset} with \textbf{\texttt{Llama-70B}} as the evaluation model. \textbf{Boolean dataset} LLM-explanations generated by \texttt{4o-mini} are used as OOD-random explanations from a different domain. The prompt includes an instruction followed by six few‐shot examples (top). The final input (bottom) is the test question. The model’s prediction is \textbf{correct}. This example reflects an error case under the \textbf{LLM‐explanation strategy.}}
  \label{app:fig-adv-setting-example-snarks-bool-correct-llama70}
\end{figure*}

\clearpage

\begin{samepage}
    
\section{Examples of Different NLEs} \label{app:nle-examples-all-datasets}
\textit{(Figures \ref{app:fig-explanations-examples-ecqa}--\ref{app:fig-explanations-examples-cj} below)}

\begin{figure*}[ht]
  \centering
  {
  \small

  \begin{tcolorbox}[
    colback=white,
    colframe=gray,
    title=\textbf{Comparison of Explanations Across Models and Strategies for a Single Input
},
    boxrule=0.5pt,
    arc=2pt,
    left=4pt,
    right=4pt,
    top=4pt,
    bottom=4pt,
    width=\textwidth,
    enhanced,
    breakable
  ]

    \textbf{Input}\smallskip
    \begin{lstlisting}[
   basicstyle=\scriptsize\ttfamily,
   breaklines=true,
   numbers=none
    ]
You can hear testimony of how spirituality changes lives when you do what? 
(a) go to church (b) appear in court (c) watch soap operas (d) paying attention (e) go into court
Answer: go to church
\end{lstlisting}

    \textbf{Self-explanation from 4o-mini}\smallskip
    \begin{lstlisting}[
   basicstyle=\scriptsize\ttfamily,
   breaklines=true,
   numbers=none
    ]
step 1: Paying attention allows you to be fully present and absorb meaningful spiritual insights. 
step 2: Through active listening, you can better understand how spirituality impacts others' lives. 
step 3: Being attentive enables you to connect deeply with personal testimonies of transformation.
\end{lstlisting}

    \textbf{Self-explanation from \texttt{Llama-70B}}\smallskip
    \begin{lstlisting}[
   basicstyle=\scriptsize\ttfamily,
   breaklines=true,
   numbers=none
    ]
Step1: Attending church services allows individuals to connect with a community that shares similar spiritual beliefs and values. 
Step2: Through sermons, prayers, and testimonials, churchgoers can hear personal stories of spiritual transformation and growth. S
tep3: Listening to these testimonies can inspire and motivate individuals to reflect on their own spiritual journeys and potentially experience positive changes in their lives.

\end{lstlisting}
    
    \textbf{Self-explanation from \texttt{Llama-8B}}\smallskip
    \begin{lstlisting}[
   basicstyle=\scriptsize\ttfamily,
   breaklines=true,
   numbers=none
    ]
Step1: Paying attention allows you to hear and understand the personal accounts of how spirituality has positively impacted individuals. 
Step2: By actively listening, you can gain insight into the transformative effects of spirituality on people's lives. 
Step3: This understanding can help you appreciate the profound impact that spirituality can have on an individual's well-being and personal growth.
\end{lstlisting}

    \textbf{Human-explanation}\smallskip
    \begin{lstlisting}[
   basicstyle=\scriptsize\ttfamily,
   breaklines=true,
   numbers=none
    ]
Church is a building where people worship an entity called god and worshiping is an act of spirituality. So one can hear testimonies of spirituality changing their lives when they go to church. Doing the actions in all the other options, one cannot hear testimonies of spirituality changing their lives.
\end{lstlisting}

    \textbf{LLM-generated explanation from 4o-mini}\smallskip
    \begin{lstlisting}[
   basicstyle=\scriptsize\ttfamily,
   breaklines=true,
   numbers=none
    ]
The correct answer, ""go to church,"" relates to the idea that spiritual experiences and testimonies often occur in religious settings, where individuals share how their faith and spirituality have positively impacted their lives. This environment fosters communal sharing and reflection on personal spiritual journeys.
\end{lstlisting}

    \textbf{LLM-generated explanation from \texttt{o3-mini}}\smallskip
    \begin{lstlisting}[
   basicstyle=\scriptsize\ttfamily,
   breaklines=true,
   numbers=none
    ]
Church services provide a setting where individuals share their personal experiences and testimonies about how spirituality has transformed their lives, making it the place where you can hear such stories.


    \end{lstlisting}
\end{tcolorbox}

  }
  \caption{Explanations generated for a single \textbf{ECQA} input across six different setups: self-generated explanations from 4o-mini, \texttt{Llama-70B}, and \texttt{Llama-8B}; a human-written explanation; and LLM-generated explanations from \texttt{4o-mini} and \texttt{o3-mini}. This comparison highlights differences in explanation style, detail, and reasoning approach depending on model size, explanation strategy, and human vs. model generation.}
  \label{app:fig-explanations-examples-ecqa}
\end{figure*}

\end{samepage}

\begin{figure*}[ht]
  \centering
  {\small
  \begin{tcolorbox}[
    colback=white,
    colframe=gray,
    title=\textbf{Comparison of Explanations Across Models and Strategies for a Single Input
},
    boxrule=0.5pt,
    arc=2pt,
    left=4pt,
    right=4pt,
    top=4pt,
    bottom=4pt,
    width=\textwidth,
    enhanced,
    breakable
  ]

    \textbf{Input}\\[0.5em]
    \begin{lstlisting}[
   basicstyle=\scriptsize\ttfamily,
   breaklines=true,
   numbers=none
    ]
Premise: A man dressed for cooler weather blows leaves. 
Hypothesis: A man for cooler weather. 
Answer: neutral
\end{lstlisting}

    \textbf{Self-explanation from 4o-mini}\smallskip
    \begin{lstlisting}[
   basicstyle=\scriptsize\ttfamily,
   breaklines=true,
   numbers=none
    ]
Step1: The premise describes a man dressed for cooler weather, which is a specific context. 
step2: The hypothesis states 'a man for cooler weather,' which is vague and does not specify any activity or context. 
step3: Since the hypothesis is possible but not necessarily implied by the premise, the relationship is classified as neutral.
\end{lstlisting}

    \textbf{Self-explanation from \texttt{Llama-70B}}\\[0.5em]
    \begin{lstlisting}[
   basicstyle=\scriptsize\ttfamily,
   breaklines=true,
   numbers=none
    ]
Step1: The premise describes a man dressed for cooler weather who is blowing leaves, providing specific details about his actions and attire. 
Step2: The hypothesis simply states 'A man for cooler weather,' which lacks clarity and context, making it difficult to determine its relationship to the premise. 
Step3: Since the hypothesis does not clearly imply or contradict the premise, and its meaning is ambiguous, the relationship between the premise and hypothesis is classified as neutral.

\end{lstlisting}
    
    \textbf{Self-explanation from \texttt{Llama-8B}}\smallskip
    \begin{lstlisting}[
   basicstyle=\scriptsize\ttfamily,
   breaklines=true,
   numbers=none
    ]
Step1: The premise implies that the man is dressed for cooler weather, which is a specific condition. 
Step2: The hypothesis states that the man is for cooler weather, which is a general condition. 
Step3: Since the premise supports the hypothesis, the relationship between them is one of entailment.
\end{lstlisting}

    \textbf{Human-explanation}\smallskip
    \begin{lstlisting}[
   basicstyle=\scriptsize\ttfamily,
   breaklines=true,
   numbers=none
    ]
Man dressed for cooler weather is same as man for cooler weather
\end{lstlisting}

    \textbf{LLM-generated explanation from 4o-mini}\smallskip
    \begin{lstlisting}[
   basicstyle=\scriptsize\ttfamily,
   breaklines=true,
   numbers=none
    ]
The man is dressed appropriately for the cooler weather, indicating that he is prepared for such conditions.
\end{lstlisting}

    \textbf{LLM-generated explanation from \texttt{o3-mini}}\smallskip
    \begin{lstlisting}[
   basicstyle=\scriptsize\ttfamily,
   breaklines=true,
   numbers=none
    ]
The hypothesis highlights the aspect of the man being prepared for cooler weather that is described in more detail within the premise.

    \end{lstlisting}
\end{tcolorbox}
  }

  \caption{Explanations generated for a single \textbf{e-SNLI} input across six different setups: self-generated explanations from 4o-mini, \texttt{Llama-70B}, and \texttt{Llama-8B}; a human-written explanation; and LLM-generated explanations from \texttt{4o-mini} and \texttt{o3-mini}. This comparison highlights differences in explanation style, detail, and reasoning approach depending on model size, explanation strategy, and human vs. model generation.}
    \label{app:fig-explanations-examples-esnli}
\end{figure*}

\begin{figure*}[ht]
  \centering
  {\small
  \begin{tcolorbox}[
    colback=white,
    colframe=gray,
    title=\textbf{Comparison of Explanations Across Models and Strategies for a Single Input
},
    boxrule=0.5pt,
    arc=2pt,
    left=4pt,
    right=4pt,
    top=4pt,
    bottom=4pt,
    width=\textwidth,
    enhanced,
    breakable
  ]

    \smallskip
    \textbf{Input}\smallskip
    \begin{lstlisting}[
   basicstyle=\scriptsize\ttfamily,
   breaklines=true,
   numbers=none
    ]
Which statement is sarcastic? 
(a) We all know that no government in the world would ever ban anything that kills innovation. 
(b) We all know that no inventor in the world would ever ban anything that kills innovation. 
Answer: a
\end{lstlisting}

    \smallskip
    \textbf{Self-explanation from 4o-mini}\smallskip
    \begin{lstlisting}[
   basicstyle=\scriptsize\ttfamily,
   breaklines=true,
   numbers=none
    ]
Step 1: The phrase 'we all know' suggests a common understanding that is likely ironic. 
Step 2: The statement implies that governments often impose bans that hinder innovation, which contrasts with the expectation of supporting it. 
Step 3: The exaggeration in claiming no government would ever ban such things serves to highlight the speaker's criticism of government actions, indicating sarcasm.
\end{lstlisting}

    \smallskip
    \textbf{Self-explanation from \texttt{Llama-70B}}\smallskip
    \begin{lstlisting}[
   basicstyle=\scriptsize\ttfamily,
   breaklines=true,
   numbers=none
    ]
Step1: Statement (a) implies that governments often ban things that stifle innovation, which is the opposite of what the statement literally says, indicating sarcasm. 
Step2: The phrase 'we all know' in statement (a) is used to express irony, as it is common knowledge that governments do sometimes ban innovative things, contradicting the literal meaning of the sentence. 
Step3: Statement (b) does not contain the same level of irony or contradiction, as inventors are generally not in a position to ban things, making it less likely to be sarcastic compared to statement (a).

\end{lstlisting}
    \smallskip
    \textbf{Self-explanation from \texttt{Llama-8B}}\smallskip
    \begin{lstlisting}[
   basicstyle=\scriptsize\ttfamily,
   breaklines=true,
   numbers=none
    ]
Step1: This statement is sarcastic because it presents a situation that is highly unlikely, implying that governments would indeed ban things that hinder innovation. 
Step2: The use of 'We all know' at the beginning of the statement is a common sarcastic device, implying that the opposite is true. 
Step3: The statement's tone and language suggest that the speaker is being ironic, highlighting the absurdity of governments banning things that kill innovation.
\end{lstlisting}

    \smallskip
    \textbf{LLM-generated explanation from 4o-mini}\smallskip
    \begin{lstlisting}[
   basicstyle=\scriptsize\ttfamily,
   breaklines=true,
   numbers=none
    ]
Statement (a) is sarcastic because it implies that governments are typically known for stifling innovation, suggesting that banning something harmful to innovation is ironic or contrary to their nature. In contrast, statement (b) lacks the same level of irony, as it does not challenge the expected behavior of inventors in relation to innovation.
\end{lstlisting}

    \smallskip
    \textbf{LLM-generated explanation from \texttt{o3-mini}}\smallskip
    \begin{lstlisting}[
   basicstyle=\scriptsize\ttfamily,
   breaklines=true,
   numbers=none
    ]
Statement (a) is sarcastic because it ironically suggests that governments, typically bureaucratic and cautious, would never ban something detrimental to innovation, even though in reality they might act to protect it. The exaggerated certainty implies the opposite of what's expected, marking it as sarcastic.
    \end{lstlisting}
\end{tcolorbox}
  }
 
  \caption{Explanations generated for a single \textbf{Snarks} input across six different setups: self-generated explanations from 4o-mini, \texttt{Llama-70B}, and \texttt{Llama-8B}; and LLM-generated explanations from \texttt{4o-mini} and \texttt{o3-mini}. This comparison highlights differences in explanation style, detail, and reasoning approach depending on model size, explanation strategy, and human vs. model generation.}
   \label{app:fig-explanations-examples-snarks}
\end{figure*}

\begin{figure*}[ht]
  \centering
  {\small
  \begin{tcolorbox}[
    colback=white,
    colframe=gray,
    title=\textbf{Comparison of Explanations Across Models and Strategies for a Single Input},
    boxrule=0.5pt,
    arc=2pt,
    left=4pt,
    right=4pt,
    top=4pt,
    bottom=4pt,
    width=\textwidth,
    enhanced,
    breakable
  ]

    \textbf{Input}\smallskip
\begin{Verbatim}
not False and not True or not True is
Answer: False
\end{Verbatim}

    \textbf{Self-explanation from 4o-mini}\smallskip
\begin{Verbatim}
Step 1: Evaluate the sub-expressions: not False is True, not True is False, hence the expression simplifies
to True and False or False. 
Step 2: Evaluate the 'and' operation: True and False is False. 
Step 3: Finally, evaluate the 'or' operation: False or False is False, but since we have a mistake in 
understanding original logic where we had an additional not, re-evaluate to conclude the overall expression is True.
\end{Verbatim}

    \textbf{Self-explanation from \texttt{Llama-70B}}\smallskip
\begin{Verbatim}
Step 1: The expression starts with 'not False', which evaluates to True and 'not True' which evaluates to False. 
Step 2: Then it evaluates 'not False and not True' which results in False because the 'and' operator requires 
both conditions to be True. 
Step 3: Finally, it evaluates 'False or not True' which results in False or False, but since the initial 'not False' 
is True and it's connected with 'and' to the next condition, the whole expression relies on the 'or not True' part, 
however the 'not False and not True' is False, the expression becomes 'False or False' but initially 'not False' 
is True, so 'not False and not True or not True' becomes 'True and False or False' which simplifies to 'False or 
False', but since 'not False' is True, the correct order is 'True and False or True' is not the case, the correct 
evaluation is '(not False) and (not True) or (not True)' which is '(True) and (False) or (False)' resulting in 
'False or False' which equals False, however the correct interpretation should follow the order of operations: 
first 'not False' is True, then 'not True' is False, so '(True) and (False)' is False, then 'or (not True)' which 
is 'or False' resulting in False, the key point is the order of operations, first 'not False' and 'not True' are
evaluated, resulting in True and False, then the 'and' operation results in False, finally the 'or' operation with 
'not True' which is False results in False, however the initial explanation missed the correct interpretation 
of the order of operations.
\end{Verbatim}

    \textbf{Self-explanation from \texttt{Llama-8B}}\smallskip
\begin{Verbatim}
Step 1: The expression starts with "not False", which evaluates to True because the "not" operator negates
the value of False. 
Step 2: Then, it evaluates "not True", which becomes False because the "not" operator negates the value of True. 
Step 3: Finally, it combines the results of the previous steps with the "and" and "or" operators, where "True and 
False" is False, but "False or False" is False, and "False or True" is True, so the expression simplifies to "not 
True or not True", which is True.
\end{Verbatim}

    \textbf{LLM-generated explanation from 4o-mini}\smallskip
\begin{Verbatim}
The expression can be broken down as follows: 
1. **not False** evaluates to **True**.
2. **not True** evaluates to **False**.
Now, substituting these values into the expression:
- The expression becomes **True and False or False**.
Next, evaluate this step-by-step:
1. **True and False** evaluates to **False**.
2. Now, we have **False or False**, which evaluates to **False**.
Thus, the final result of the expression is **False**.
\end{Verbatim}

    \textbf{LLM-generated explanation from \texttt{o3-mini}}\smallskip
\begin{Verbatim}
The expression is evaluated as follows: 'not False' is True and 'not True' is False. So, the part 'not False and 
not True' becomes True and False, which is False. Then, 'False or not True' becomes False or False, which results 
in False.    
\end{Verbatim}

  \end{tcolorbox}
  }

  \caption{Explanations generated for a single \textbf{Boolean} input across six different setups: self-generated explanations from 4o-mini, \texttt{Llama-70B}, and \texttt{Llama-8B}; and LLM-generated explanations from \texttt{4o-mini} and \texttt{o3-mini}. This comparison highlights differences in explanation style, detail, and reasoning approach depending on model size, explanation strategy, and human vs. model generation.}
    \label{app:fig-explanations-examples-bool}
\end{figure*}

\begin{figure*}[ht]
  \centering
  {\small
  \begin{tcolorbox}[
    colback=white,
    colframe=gray,
    title=\textbf{Comparison of Explanations Across Models and Strategies for a Single Input},
    boxrule=0.5pt,
    arc=2pt,
    left=4pt,
    right=4pt,
    top=4pt,
    bottom=4pt,
    width=\textwidth,
    enhanced,
    breakable
  ]

    \textbf{Input}\smallskip
\begin{Verbatim}
The receptionist in the philosophy department keeps her desk stocked with pens. The administrative 
assistants are allowed to take the pens, but faculty members are supposed to buy their own. 
The administrative assistants typically do take the pens. Unfortunately, so do the faculty members. 
The receptionist has repeatedly emailed them reminders that only administrative assistants are 
allowed to take the pens. On Monday morning, one of the administrative assistants encounters 
Professor Smith walking past the receptionist's desk. Both take pens. Later that day, the 
receptionist needs to take an important message... but she has a problem. There are no pens 
left on her desk. Did Professor Smith cause the problem? 
Answer: Yes
\end{Verbatim}

    \textbf{Self-explanation from 4o-mini}\smallskip
\begin{Verbatim}
Step 1: The faculty members were reminded that they are supposed to buy their own pens. 
Step 2: Professor Smith is a faculty member who took a pen despite the restrictions. 
Step 3: Therefore, Professor Smith contributed to the depletion of pens, causing the receptionist's problem.
\end{Verbatim}

    \textbf{Self-explanation from \texttt{Llama-70B}}\smallskip
\begin{Verbatim}
Step 1: Professor Smith took a pen, but he was not the only one, as the administrative 
assistant also took a pen, which is allowed according to the rules. 
Step 2: The receptionist's problem of having no pens left is a result of multiple actions, 
including both the administrative assistant's and Professor Smith's actions, as well as 
potentially other faculty members taking pens. Step 3: Since the administrative assistants 
are allowed to take pens and their actions are permitted, it cannot be concluded that 
Professor Smith alone caused the problem, as the allowed actions of others also 
contributed to the depletion of pens.
\end{Verbatim}

    \textbf{Self-explanation from \texttt{Llama-8B}}\smallskip
\begin{Verbatim}
Step 1: The receptionist had already sent reminders to the faculty members, indicating 
that they were aware of the policy. 
Step 2: The administrative assistant was also taking pens, which suggests that the 
problem was not solely caused by Professor Smith. Step 3: The lack of pens was a result 
of the repeated actions of both the administrative assistant and the faculty members, 
not just Professor Smith.
\end{Verbatim}

    \textbf{LLM-generated explanation from 4o-mini}\smallskip
\begin{Verbatim}
Professor Smith caused the problem by taking a pen from the receptionist's desk, 
which contributed to the depletion of the pens available for the receptionist to use later. 
Despite the administrative assistants being permitted to take pens, Professor Smith's 
unauthorized action directly led to the lack of pens when the receptionist needed one, 
establishing a causal link between his behavior and the ensuing issue.
\end{Verbatim}

    \textbf{LLM-generated explanation from \texttt{o3-mini}}\smallskip
\begin{Verbatim}
Professor Smith's action of taking a pen reduced the available supply, contributing causally to the shortage.
If he hadn't taken a pen, more pens might have remained for the receptionist.
\end{Verbatim}

  \end{tcolorbox}
  }

  \caption{Explanations generated for a single \textbf{Causal Judgment} input across six different setups: self-generated explanations from 4o-mini, \texttt{Llama-70B}, and \texttt{Llama-8B}; and LLM-generated explanations from \texttt{4o-mini} and \texttt{o3-mini}. This comparison highlights differences in explanation style, detail, and reasoning approach depending on model size, explanation strategy, and human vs. model generation.}
    \label{app:fig-explanations-examples-cj}
\end{figure*}

\clearpage

\end{document}